\documentclass[pdflatex,sn-mathphys-num]{sn-jnl}%
\usepackage[mathlines]{lineno}

\usepackage{graphicx}
\usepackage{silence}
\usepackage[compatibility=false]{caption}
\usepackage{etoolbox}
\usepackage{multirow}%
\usepackage{amsmath,amssymb,amsfonts}%
\usepackage{amsthm}%
\usepackage[title]{appendix}%
\usepackage{xcolor}%
\usepackage{textcomp}%
\usepackage{manyfoot}%
\usepackage{booktabs}%
\usepackage{algorithm}%
\usepackage{algorithmicx}%
\usepackage{algpseudocode}%
\usepackage{listings}%
\usepackage{array}
\usepackage{longtable}
\usepackage[dvipsnames,table]{xcolor}
\usepackage{booktabs}

\usepackage{amsmath,amsfonts,bm}

\def\eqref#1{equation~\ref{#1}}
\def\plaineqref#1{\ref{#1}}

\def\1{\bm{1}}

\DeclareMathAlphabet{\mathsfit}{\encodingdefault}{\sfdefault}{m}{sl}
\SetMathAlphabet{\mathsfit}{bold}{\encodingdefault}{\sfdefault}{bx}{n}

\newcommand{\revred}[1]{{\color{black}#1}}

\makeatletter
\patchcmd{\@maketitle}
  {Corresponding author(s). E-mail(s):}
  {Corresponding author. Email:}
  {}
  {\PackageError{sn-article}{Unable to update the corresponding-author label}{}}
\makeatother

\theoremstyle{thmstyleone}%
\newtheorem{theorem}{Theorem}%
\newtheorem{proposition}[theorem]{Proposition}%
\newtheorem{corollary}[theorem]{Corollary}%

\theoremstyle{thmstyletwo}%

\theoremstyle{thmstylethree}%
\newtheorem{assumption}{Assumption}

\def\0{{\mathbb 0}}

\begin{document}
\title[GeoFunFlow]{GeoFunFlow: Geometric function flow matching for joint probabilistic inference of physical fields and complex geometries}
\author[1]{\fnm{Yajie} \sur{Ji}}\equalcont{These authors contributed equally to this work.}
\author[2]{\fnm{Sifan} \sur{Wang}}\equalcont{These authors contributed equally to this work.}
\author[3]{\fnm{Zhikai} \sur{Wu}}
\author[4]{\fnm{David} \sur{van Dijk}}
\author*[1,5]{\fnm{Lu} \sur{Lu}}\email{lu.lu@yale.edu}

\affil[1]{\orgdiv{Department of Statistics and Data Science},
\orgname{Yale University},
\orgaddress{\city{New Haven}, \state{CT}, \country{USA}}}

\affil[2]{\orgdiv{Institute for Foundations of Data Science},
\orgname{Yale University},
\orgaddress{\city{New Haven}, \state{CT}, \country{USA}}}

\affil[3]{\orgdiv{Department of Computer Science},
\orgname{Yale University},
\orgaddress{\city{New Haven}, \state{CT}, \country{USA}}}

\affil[4]{\orgdiv{Department of Internal Medicine},
\orgname{Yale University},
\orgaddress{\city{New Haven}, \state{CT}, \country{USA}}}

\affil[5]{\orgdiv{Department of Chemical and Environmental Engineering},
\orgname{Yale University},
\orgaddress{\city{New Haven}, \state{CT}, \country{USA}}}

\abstract{Inverse problems governed by partial differential equations (PDEs) arise widely in science and engineering, but are often ill-posed and limited by sparse, noisy observations. In many applications, measurements reveal only part of the physical state, while the domain geometry may also be unknown even though it shapes the observed response. Joint field and geometry inference across varying computational domains and discretizations remains challenging, whereas many existing machine learning approaches are designed for known geometries and deterministic field reconstruction. Here, we introduce {\em GeoFunFlow}, a probabilistic framework that unifies field reconstruction on known domains and joint field and geometry inference on unknown domains. GeoFunFlow combines a geometric function autoencoder (GeoFAE) with flow matching in the latent space to model a joint distribution over physical fields and geometries. GeoFAE establishes a common representation across spatial discretizations that captures the relationship between physical fields and domain geometries, with unknown geometry represented by a signed distance function. The resulting representation allows observations to guide both field reconstruction and geometry recovery, while latent rectified flow enables efficient conditional sampling and spatially resolved uncertainty quantification. A calibration procedure further provides geometry uncertainty estimates with interpretable empirical coverage. Across seven benchmarks spanning porous media flow, fluid mechanics, and optical tomography, GeoFunFlow accurately recovers fields and geometries across complex, variable, and unknown domains while quantifying spatially resolved conditional uncertainty.}


\maketitle

\section{Introduction}\label{sec1}

Inverse problems governed by partial differential equations (PDEs) are central to scientific discovery and engineering design, as they aim to recover unknown physical states from indirect, partial, or noisy observations. They arise in diverse settings, including medical imaging, where tissue properties must be reconstructed from MRI or CT data~\cite{natterer2001mathematical}, geophysics, where subsurface structures are inferred from seismic measurements~\cite{tarantola2005inverse}, and fluid dynamics, where flow fields must be recovered from limited sensor observations~\cite{cotter2009bayesian}. Yet in many real-world applications, the difficulty of these problems is shaped not only by sparse and noisy data, but also by geometry: the underlying computational domains are rarely simple canonical regions, and instead often involve patient-specific anatomies, complex geological formations, industrial components with cavities or defects, and flows around moving or irregular bodies~\cite{santos2023development}. These inverse problems are typically ill-posed, and the difficulty is further intensified when geometry plays an active role. In many applications, the domain is irregular or geometrically complex, so that geometry not only determines the boundary conditions of the governing PDEs but also strongly affects the resulting physical fields. In more challenging cases, the geometry is itself only partially observed or entirely unknown and must be inferred jointly with the physical state, which substantially enlarges the space of admissible solutions and increases ambiguity. In inverse problems of this kind, uncertainty is not merely a secondary output, because sparse observations, unknown geometry, and intrinsic non-uniqueness can all give rise to multiple plausible explanations, making reliable uncertainty quantification (UQ) essential for informed downstream decision-making.

Classical approaches typically formulate PDE inverse problems as PDE-constrained optimization, sometimes within Bayesian frameworks for uncertainty quantification~\cite{tarantola2005inverse,cotter2009bayesian}. These methods can be computationally prohibitive at scale, as each iteration may require forward and adjoint solves, while posterior sampling may demand thousands of such solves. The cost becomes even higher in geometrically complex or unknown domains, where mesh generation, discretization, and repeated solves across candidate geometries introduce substantial additional overhead~\cite{starostin2025fast,chang2023recovering,zhou2023neural,yin2025tddm}. When the domain itself is unknown, uncertainty in the physical state becomes inseparable from uncertainty in the geometry on which that state is defined. This motivates the central question of this work: how to infer the joint distribution of physical fields and geometry from partial observations.
Machine-learning-based surrogates offer a scalable route to inverse inference. Much of the recent progress has been driven by \emph{operator learning}, which seeks to approximate PDE solution maps directly between function spaces. Seminal neural operators such as DeepONet~\cite{lu2021learning} and the Fourier Neural Operator (FNO)~\cite{li2021fourier}, together with a broad family of spectral~\cite{bonev2023spherical}, convolutional~\cite{wen2022u,raonic2023convolutional}, and transformer-based variants~\cite{cao2021choose,hao2023gnot}, have shown good performance across domains including weather forecasting~\cite{pathak2022fourcastnet}, micromechanics~\cite{WangLiuMicrometer}, astrophysics~\cite{mao2023ppdonet}, and geosciences~\cite{zhu2023fourier}. These methods have been especially successful for forward modeling on structured domains, where discretizations are regular and inductive biases such as convolution or spectral structure are naturally aligned with the data. More recently, geometry-aware neural operators have extended this paradigm to irregular meshes and complex domains by incorporating geometric priors, point-cloud representations, or mesh-aware attention mechanisms~\cite{li2023fourier,li2023geometry,serrano2023operator,yin2024scalable,chen2024positional,alkin2024universal,wu2024transolver,zeng2025point,jiao2026unknownmanifolds}. %
However, existing paradigms remain incomplete when geometry plays a central role, especially in the presence of geometric variability, unknown domains, and uncertainty.

A parallel development in scientific machine learning has shifted the focus from deterministic prediction to \emph{distributional modeling}. Generative models, particularly diffusion models~\cite{sohl2015deep,song2020score,ho2020denoising},  provide a natural framework for learning distributions over possible solutions rather than producing a single point estimate. This capability is especially valuable for ill-posed inverse problems, in which sparse and noisy observations may be consistent with multiple physically plausible solutions, making uncertainty quantification essential for reliable inference. Following the success of diffusion models in vision and language~\cite{ho2020denoising,song2020score}, generative approaches have increasingly begun to reshape scientific machine learning, with applications ranging from turbulence and weather to metamaterial design and high-dimensional dynamics~\cite{du2024conditional,li2024generative,bastek2023inverse,li2024learning,gao2024generative,shysheya2024conditional}. Within PDE learning, recent methods such as DiffFNO~\cite{liu2025difffno}, the Wavelet Diffusion Neural Operator~\cite{hu2025wavelet}, Denoising Diffusion Operators~\cite{lim2025score}, FunDiff~\cite{wang2026fundiff}, DiffusionPDE~\cite{huang2024diffusionpde}, RED-DiffEq~\cite{shan2026regularization}, physics-informed diffusion models~\cite{bastek2025physics}, and physics-constrained flow matching~\cite{utkarsh2025physics} have demonstrated the potential of generative models for function-space modeling, inverse inference, and the incorporation of physical constraints. A particularly promising direction is latent generative PDE modeling, in which high-dimensional states are compressed into learned latent variables and generation is performed in the reduced space, improving computational efficiency while offering greater flexibility across discretizations and conditioning modalities. Recent examples, including AROMA~\cite{serrano2024aroma}, Text2PDE~\cite{zhou2025text2pde}, and latent generative PDE solvers based on flow matching~\cite{li2025generative,chen2026latent}, further highlight the potential of latent-space modeling for irregular or heterogeneous discretizations.

The distributional perspective has further motivated methods that explicitly address UQ in generative and operator-learning frameworks. These efforts include approximate Bayesian UQ for surrogate-based inverse PDE solvers~\cite{wang2024total, winovich2026active}, diffusion-based neural operators with calibrated uncertainty estimates~\cite{matveev2026light}, and more recent flow-matching frameworks that propagate uncertainty through generative dynamics and distinguish aleatoric from epistemic uncertainty under distribution shift~\cite{han2026flow,wu2026uncertainty}. Despite these advances, uncertainty-aware inverse inference remains less explored for problems involving complex, sample-dependent, or unknown geometries, particularly when physical fields and domain geometry must be recovered jointly with their coupled uncertainty.

Against this backdrop, we introduce GeoFunFlow, a geometric function-space framework for probabilistic inverse inference across complex, variable, and unknown domains. GeoFunFlow combines continuous function representation with conditional rectified flow to reconstruct physical fields on prescribed geometries and jointly infer fields and geometry from partial observations. Our main contributions are as follows:
\begin{itemize}
    \item We develop a geometric function autoencoder (GeoFAE) that learns shared latent representations of physical fields and geometry from inputs of varying sizes and structures. GeoFAE maps flexible combinations of coordinates, physical and geometric variables, and available observations into a common latent space, and decodes the target fields at arbitrary query coordinates.
    
    \item GeoFunFlow models the joint conditional distribution of physical fields and geometry from sparse and noisy observations using conditional rectified flow in the learned latent space. Sampling from this distribution yields multiple plausible reconstructions and spatially resolved uncertainty estimates. For inferred geometries, an additional calibration procedure improves the empirical coverage of uncertainty regions derived from the predicted signed distance function (SDF).
    
    \item We establish approximation bounds for the learned conditional distribution, relating its error after decoding to the approximation errors introduced by latent flow matching and autoencoding.
    
    \item We validate GeoFunFlow on diverse inverse PDE problems spanning structured and unstructured discretizations as well as variable and unknown geometries, demonstrating accurate reconstruction and spatially resolved uncertainty quantification for both physical fields and inferred geometries.
\end{itemize}

\section{Results}\label{sec2}

We evaluate GeoFunFlow from both theoretical and empirical perspectives. After introducing the inverse problem setting and outlining the workflow, we present a decomposition of posterior approximation error. Experiments then assess field reconstruction on known domains, joint field and geometry inference on unknown domains, and the accuracy and calibration of posterior uncertainty estimates. Formal notation, architectural details, training and sampling procedures, and uncertainty quantification details are presented in Sec.~\ref{sec4}.

\subsection{Problem setup and GeoFunFlow overview}\label{sec:problem_overview}

\begin{figure}[htbp]
    \centering
    \includegraphics[width=\textwidth]{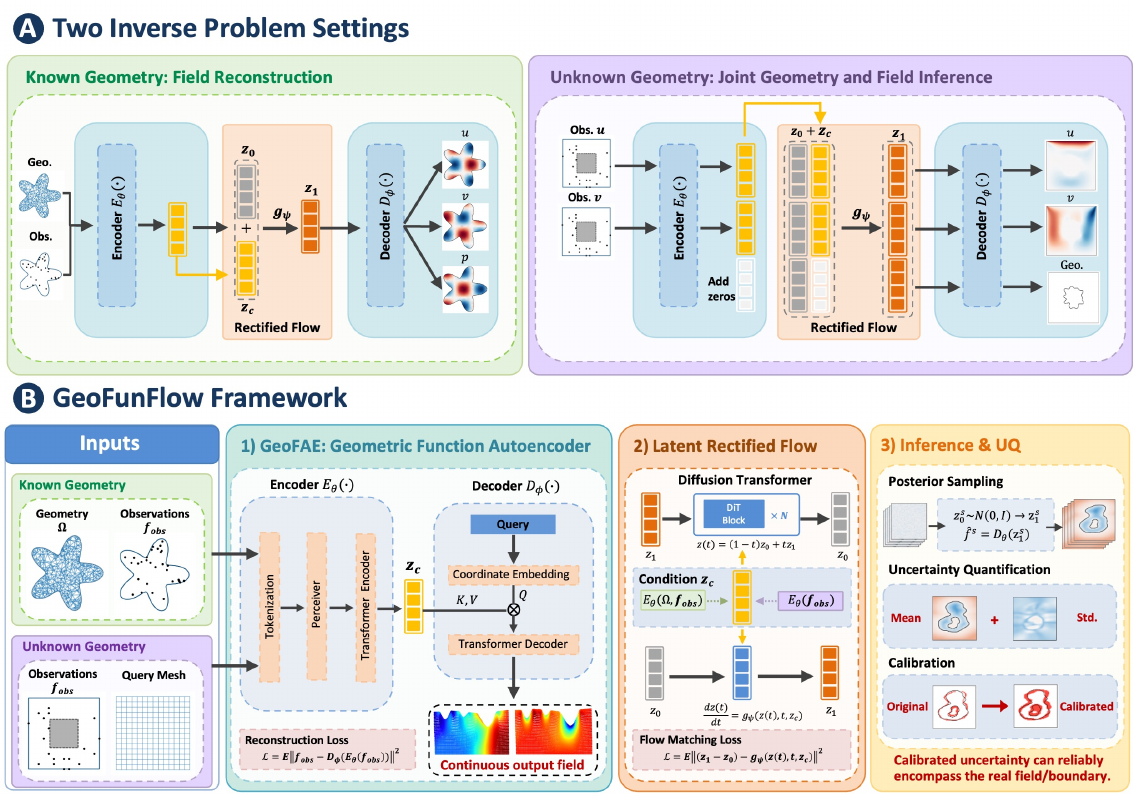}
    \caption{\textbf{Overview of the two inverse problem settings and GeoFunFlow.} (\textbf{A}) GeoFunFlow addresses two representative inverse settings: field reconstruction from sparse and noisy observations when the geometry is known, and joint geometry-field inference when the geometry is unknown. (\textbf{B}) The GeoFunFlow framework consists of three coupled components: (1) GeoFAE, a geometric function autoencoder that learns latent representations on known or unknown geometries; (2) a latent rectified-flow model that performs conditional posterior sampling in the learned latent space; and (3) an inference and uncertainty quantification module that estimates spatially resolved field uncertainty and calibrates SDF-derived geometry uncertainty regions.}
    \label{fig:framework}
\end{figure}

\begingroup
In inverse problems, we aim to recover full physical fields from observations that may be sparse, noisy, or restricted to a subset of the relevant variables. Geometry is central to this task because it defines where the governing physics applies and how information propagates through the domain. Input representations may also differ substantially across samples, ranging from regular grids to irregular meshes and point clouds. A broadly applicable framework must therefore reconstruct continuous physical fields, recover the domain geometry when it is unknown, and operate without fixed assumptions about discretization, sensor configuration, or the set of observed variables. These settings are unified within a common formulation for inference from partial observations, presented formally in Sec.~\ref{sec:problem_formulation}.

We consider two settings in this work expressed within a common formulation (Fig.~\ref{fig:framework}A): (1) given a prescribed domain, the model reconstructs unobserved regions or variables of the physical state from partial measurements. (2) when the domain is unknown, the model jointly infers the physical fields and geometry represented by an SDF. In either setting, the input consists of spatial coordinates, available measurements, and observation masks specifying which quantities are known. The model then predicts the complete target functions. For a fixed input, GeoFunFlow models a learned conditional distribution over these functions, which we hereafter refer to as the posterior. The two settings thus differ only in which information is observed and which must be inferred, eliminating the need for formulations tailored to individual tasks.
\endgroup

\begingroup
GeoFunFlow implements this view through a shared latent representation (Fig.~\ref{fig:framework}B). GeoFAE maps functions sampled on regular or irregular discretizations into a fixed-dimension latent space and decodes latent states continuously at arbitrary query coordinates. A conditional rectified-flow model then generates latent states by numerically solving the underlying ordinary differential equation (ODE) systems. This design separates the variable physical discretization from the fixed representation used for generative modeling and allows field reconstruction, missing-variable inference, and joint geometry--field recovery to share one architecture. These components are detailed in Sec.~\ref{sec:architecture}.

Each generated latent state produces one plausible solution of the partially observed system. Repeating the generation from independent Gaussian noise yields an ensemble that represents the learned conditional uncertainty in both physical fields and, when inferred, geometry. GeoFunFlow summarizes this ensemble by its predictive mean and spatially resolved variability. For inferred geometries, we apply a held-out calibration procedure described in Sec.~\ref{sec:uq-method} to rescale the SDF uncertainty without changing the generated solutions or the spatial ordering of their variability. 
\endgroup

\subsection{Posterior approximation analysis}\label{sec:theory-results}

\begingroup
GeoFunFlow combines condition encoding, latent transport, numerical ODE
integration, and continuous decoding. Let $P^\star(\cdot\mid\mathbf c)$ be the
target conditional law for a partial condition $\mathbf c$, let
$\widehat P_h(\cdot\mid\mathbf c)$ be the learned law generated using maximum
ODE step size $h$, and let $\mathbf z_{\mathbf c}=E_\theta(\mathbf c)$, where $E_{\theta}$ is the encoder with parameters $\theta$. The analysis in
Appendix~\ref{app:theory} gives the error decomposition
\[
W_2^\Omega\!\left(
\widehat P_h(\cdot\mid\mathbf c),
P^\star(\cdot\mid\mathbf c)
\right)
\le
L_D\!\left[
\epsilon_{\rm cond}(\mathbf c)
+\epsilon_{\rm flow}(\mathbf z_{\mathbf c})
+\epsilon_{\rm ODE}(\mathbf z_{\mathbf c};h)
\right]
+\epsilon_{\rm rec}(\mathbf c).
\]
Here, $\epsilon_{\rm cond}$ measures information lost in condition encoding, $\epsilon_{\rm flow}$ the continuous latent-transport error, $\epsilon_{\rm ODE}$ the numerical integration error, and $\epsilon_{\rm rec}$ the reconstruction error. $L_D$ is the Lipschitz constant of the decoder. The result separates the contributions of representation, generative modeling, numerical solver, and decoding. For joint field--geometry inference, the same bound applies to the coupled field--SDF distribution. Appendix~\ref{app:theory} states the complete theorem and assumptions, relates the flow term to the population rectified--flow excess loss, and gives the additional conditions required to control the induced boundary distribution.

With the method and theory established, we evaluate the performance on GeoFunFlow. We first test whether it reconstructs physical fields from sparse and noisy measurements on supplied domains. We then test whether the same framework recovers an unobserved domain from the physical response measured around it. Finally, we examine whether variation across the generated solutions identifies where the inverse problem remains uncertain and whether SDF calibration improves the coverage of the inferred geometry.
\endgroup

\subsection{Field reconstruction on known geometries}

We first evaluate GeoFunFlow on four datasets covering distinct geometries, spatial dimensions, discretizations, and physical regimes. The same network architecture is used throughout, with a separate model trained and evaluated for each dataset. The datasets comprise Darcy flow in porous media~\cite{lu2022comprehensive}, incompressible flow past a cylinder~\cite{pfaff2020learning}, compressible flow over an airfoil~\cite{pfaff2020learning}, and turbulent three-dimensional flow around the Ahmed body~\cite{li2023geometry}.
Together, these problems cover fixed and varying geometries, steady fields and snapshot reconstruction for unsteady flows, and irregular meshes ranging from a few thousand nodes to large-scale computational fluid dynamics (CFD) discretizations. 
This benchmark suite therefore tests whether GeoFunFlow remains effective as the geometry, dimensionality, and physical regime vary, rather than only in a single favorable setting. 

We evaluate predictive accuracy using relative $L^2$ error and assess uncertainty quality through the Pearson correlation between pointwise prediction error and predictive standard deviation. Definitions of these evaluation and UQ metrics are provided in Appendix~\ref{app:metrics}. Unless otherwise stated, a $1\%$ field-noise level means that the standard deviation of the additive observation noise is $1\%$ of the mean $L^2$ norm of the corresponding field over the dataset. Details of the governing equations, discretizations, data splits, and observation protocols are provided in Appendix~\ref{app:experiments}.

\begin{figure}[!htb]
    \centering
    \includegraphics[width=\linewidth]{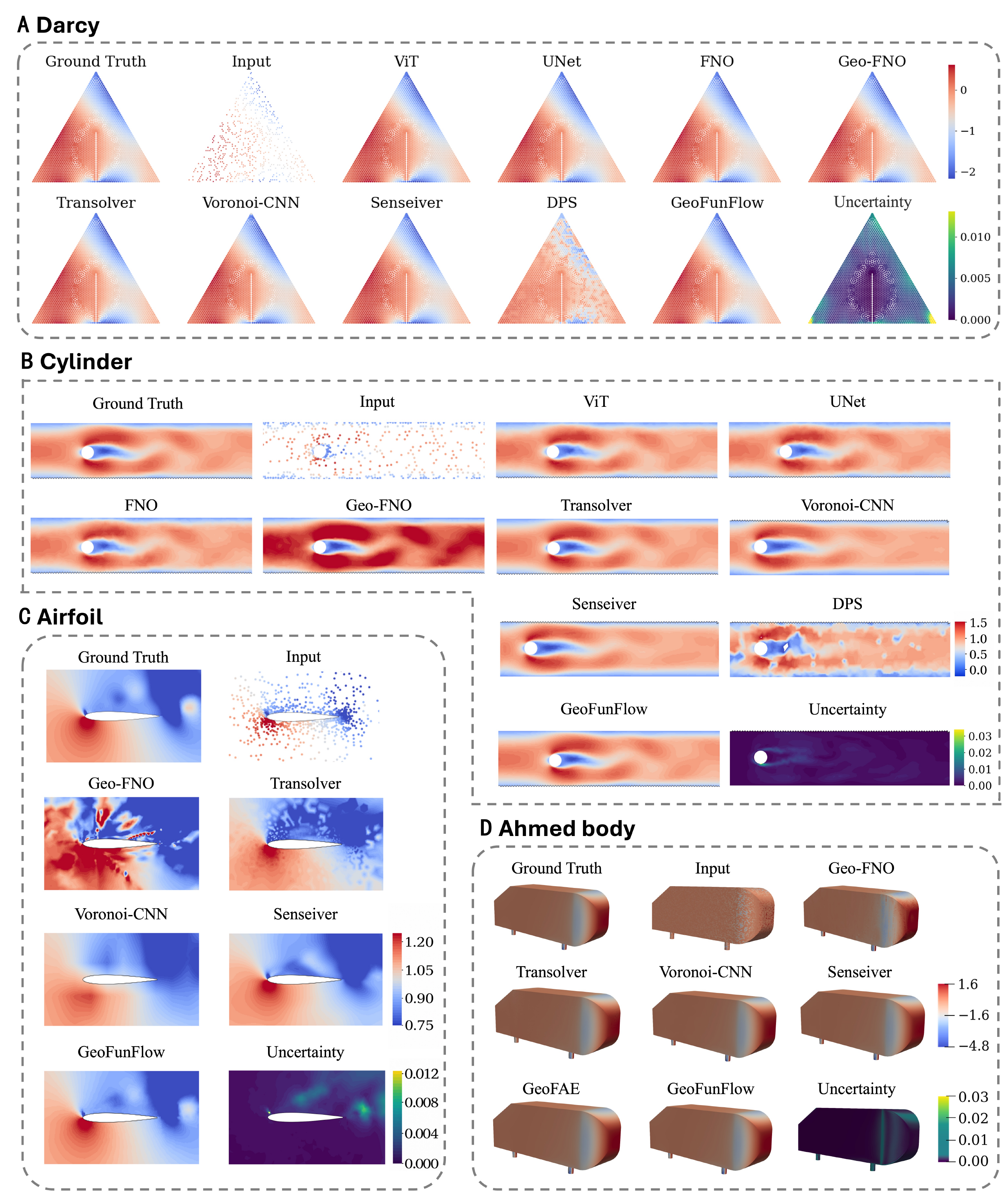}
    \caption{\textbf{Reconstruction results.} Reconstruction of the target field from noisy (1\%) and sparse observations (25\%) on the (\textbf{A}) Darcy, (\textbf{B}) Cylinder, (\textbf{C}) Airfoil, and (\textbf{D}) Ahmed body datasets. This setting matches the quantitative comparison in Table~\ref{tab:error_uq_main_compact}. For each dataset, we show one representative sample comparing our approach against several competitive baselines.}
    \label{fig:darcy_and_ahmed}
\end{figure}

Fig.~\ref{fig:darcy_and_ahmed} and Table~\ref{tab:error_uq_main_compact} report the standard sparse-observation setting with $1\%$ noise and a $25\%$ observation fraction. The central message is not only that GeoFunFlow is competitive with specialized deterministic baselines, but that it does so while solving the harder uncertainty-aware inverse problem. Across the four datasets, GeoFunFlow preserves the main spatial structures of the target fields and maintains low relative $L^2$ error after replacing a point predictor with posterior sampling in the learned latent space. This shows that the generative component provides a practical posterior representation, rather than merely adding uncertainty as an auxiliary output, without substantially sacrificing reconstruction fidelity. For a more difficult setting with $5\%$ noise and a $5\%$ observation fraction, Appendix~\ref{app:results} shows that GeoFunFlow maintains strong reconstruction accuracy across all four benchmarks.

As a representative generative approach capable of posterior inference from incomplete observations, diffusion posterior sampling (DPS)~\cite{chung2022diffusion} uses a pretrained diffusion prior together with observation-likelihood guidance to generate posterior samples conditioned on the measurements. DPS provides a useful computational comparison: it requires $1000$ denoising steps per posterior sample and produces substantially larger reconstruction errors on Darcy and Cylinder, whereas GeoFunFlow uses only $10$ rectified-flow steps. Although DPS reports a slightly higher error--uncertainty correlation on Darcy, the comparison in Appendix~\ref{app:results} (Fig.~\ref{fig:darcy_dps_geofunflow_uq}) shows that this value is accompanied by a much larger and more spatially diffuse reconstruction error. Because Pearson $r$ measures only spatial covariation, this apparent advantage should not be interpreted as superior reconstruction accuracy or calibrated UQ. This lower sampling cost makes posterior-based UQ more practical on irregular meshes and large CFD discretizations. The uncertainty maps in Fig.~\ref{fig:darcy_and_ahmed} further show that predictive spread concentrates in regions weakly constrained by the sparse observations, such as wake structures and geometrically complex surface regions.

\begin{table*}[!ht]
\centering
\begin{threeparttable}
\caption{
\textbf{Test performance on four benchmarks under $1\%$ observation noise and a $25\%$ observation fraction.} $N$ denotes the number of observation points used for reconstruction in each case.
Entries report the relative $L^2$ error ($\downarrow$). 
Values in parentheses report the Pearson correlation coefficient $r$ ($\uparrow$) between the pointwise absolute error and a method-specific uncertainty estimate. For Kriging, uncertainty is given by the Gaussian-process posterior standard deviation. For GeoFunFlow and DPS, it is estimated as the sample standard deviation over $10$ posterior samples generated for each observation. For posterior sampling, DPS requires 1,000 generative inference steps per sample, compared with only 10 for GeoFunFlow, while deterministic baselines require a single forward pass.
Best and second-best relative $L^2$ errors are highlighted in \textbf{bold} and \underline{underline}, respectively. 
A dash (--) indicates that the method is not applicable to the corresponding benchmark.
}
\label{tab:error_uq_main_compact}
\setlength{\tabcolsep}{7pt}
\begin{tabular}{lllll}
\toprule
& Darcy & Cylinder & Airfoil & Ahmed body \\
& \small{($N$=573)} & \small{($N$=250)} & \small{($N$=1250)} & \small{($N$=10000)} \\
\midrule
Interpolant      & 0.0550 & 0.2841  & 0.0382      & 0.2390      \\
Kriging          & 0.0477 (0.2737) & 0.5165 (0.1623) & 0.3770 (0.0376)    & 0.1521 (0.4900)    \\
ViT    & 0.0290 & 0.1578 & --      & --      \\
UNet   & 0.0256 & 0.1542 & --      & --      \\
FNO    & 0.0177 & 0.1628 & --      & --      \\
Geo-FNO          & \textbf{0.0101} & 0.3843 & 0.2671 & 0.2795 \\
Transolver       & 0.0289 & 0.1358 & 0.2692 & 0.1262 \\
Voronoi-CNN      & 0.0440 & 0.0814 & 0.3021 & \underline{0.0860} \\
Senseiver        & 0.0293 & 0.1055 & 0.0217 & 0.1070 \\
DPS    & 0.4007 (0.5640) & 0.2971 (0.5701) & -- & -- \\
\midrule
GeoFAE (ours)     & \underline{0.0103} & \textbf{0.0642} & \textbf{0.0096} & 0.0819 \\
GeoFunFlow (ours) & 0.0131 (0.5526) & \underline{0.0691} (0.6252) & \underline{0.0100} (0.4951) & \textbf{0.0814} (0.6048) \\
\bottomrule
\end{tabular}
\end{threeparttable}
\end{table*}

\subsection{Joint field--geometry inference from partial field information}
\label{sec:joint_results}

In many practical inverse problems, the geometry of the physical domain is inaccessible to direct measurement and must be inferred indirectly from the observed system response. Examples include identifying internal structures in medical imaging~\cite{natterer2001mathematical} and geophysical exploration~\cite{tarantola2005inverse}, detecting immersed obstacles from flow measurements~\cite{santos2023development}, and reconstructing object boundaries from scattered or tomographic data~\cite{wang2025learning}. We therefore turn to the more challenging setting in which geometry is not given explicitly but must be recovered from its imprint on partially observed physical fields. This setting is harder than field reconstruction on known domains because nonuniqueness arises at two levels simultaneously: multiple geometries may be consistent with the same sparse observations, while small geometric perturbations can induce nonlocal changes in the resulting physical state. We consider three benchmarks spanning internal obstacles, unsteady external flows, and strongly ill-posed tomography. Together, they assess whether GeoFunFlow can accurately recover geometry, remain robust to increasing observation noise, and localize geometric uncertainty in a physically meaningful manner.

\subsubsection{Lid-driven cavity flow}
\label{sec:joint_results_LDC}
\begin{figure}[!htb]
    \centering
    \includegraphics[width=\linewidth]{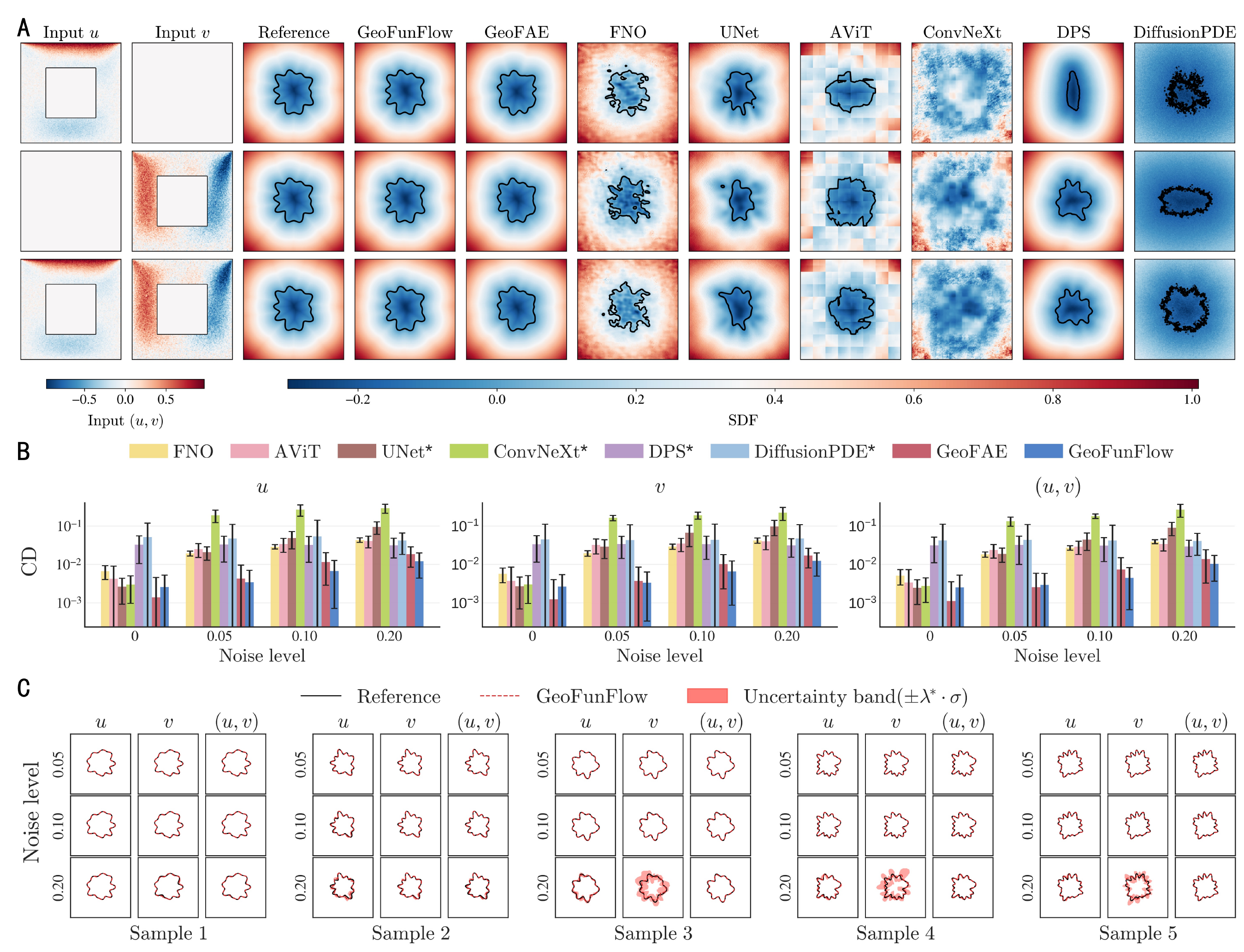}
    \caption{\textbf{Geometry inference results on the lid-driven cavity flow (LDC) benchmark with complex internal obstacles.} (\textbf{A}) Visual comparisons of the recovered geometry under three observation settings and different noise levels. The first, second, and third rows correspond to conditioning on only the horizontal velocity $u$, only the vertical velocity $v$, and both channels $(u,v)$, respectively. The black contour indicates the inferred obstacle boundary, and GeoFunFlow remains closest to the reference geometry across all settings. (\textbf{B}) Chamfer distance (CD) for geometry recovery under increasing observation noise, showing that GeoFunFlow consistently remains among the lowest-error methods. An asterisk ($*$) indicates that at least one test case for the corresponding method and noise level did not yield a converged geometry. (\textbf{C}) Representative uncertainty bands for five test samples, where the red contour visualizes the predicted geometric uncertainty around the reference boundary. The uncertainty is typically smallest when both $u$ and $v$ are observed, which is also the setting that yields the most accurate predictions.}
    \label{fig:ldc-results}
\end{figure}

We first evaluate GeoFunFlow on the lid-driven cavity flow (LDC) benchmark~\cite{tali2025flowbench}, where the task is to infer complex internal obstacle geometry from partial velocity observations in a confined flow. Fig.~\ref{fig:ldc-results} shows that GeoFunFlow achieves the most accurate and stable geometry reconstruction on the LDC benchmark. In Fig.~\ref{fig:ldc-results}A, the first, second, and third rows correspond to the three inverse settings in which only $u$, only $v$, or both $(u,v)$ are observed. This organization highlights that the framework is not limited to inferring geometry as the hidden variable. It can also recover any other unobserved field variable from partial observations within the same latent representation. In the visual comparisons, the predicted obstacle boundaries remain close to the ground truth even when only a single velocity component is observed, whereas several baselines either oversmooth the boundary or introduce spurious high-frequency artifacts. 

Quantitatively, the Chamfer distance (CD) defined in Appendix~\ref{app:metrics} increases with the observation noise level for all methods, but GeoFunFlow consistently stays in the lowest-error group across the three conditioning settings (Fig.~\ref{fig:ldc-results}B). The advantage is especially clear when both velocity channels are used, indicating that the model can effectively use complementary physical information. The uncertainty bands in Fig.~\ref{fig:ldc-results}C further show that the posterior variability is concentrated near regions of geometric uncertainty and expands as the noise level increases, which is consistent with the ill-posedness of the inverse problem. When both $u$ and $v$ are observed, the uncertainty band is typically the narrowest and the recovered geometry is the most accurate, which is consistent with the stronger physical constraints available in this setting. Beyond geometry recovery, our framework also reconstructs the full $u$ and $v$ fields. The corresponding velocity-field comparisons are presented in Figs.~\ref{fig:LDC_ufield} and~\ref{fig:LDC_vfield}, and the relative errors are reported in Table~\ref{tab:relative_l2_all_noise} of Appendix~\ref{app:results}. These additional results show that GeoFunFlow achieves consistently strong SDF reconstruction and the best overall performance in the higher-noise regime, where robustness of both geometry and field recovery is most critical.

\subsubsection{Flow past objects}
\label{sec:joint_results_FPO}
We next consider the flow-past-objects (FPO) benchmark~\cite{tali2025flowbench}, which tests whether hidden geometry can be recovered from unsteady wake signatures. For each sample, the obstacle remains fixed while the surrounding wake evolves, and the model uses partial flow measurements from one or more temporal snapshots to recover the obstacle geometry. Here, each temporal snapshot provides a different indirect view of the same obstacle, making the benchmark useful for evaluating both robustness and accuracy of temporal aggregation.

\begin{figure}[!htb]
    \centering
    \includegraphics[width=\linewidth]{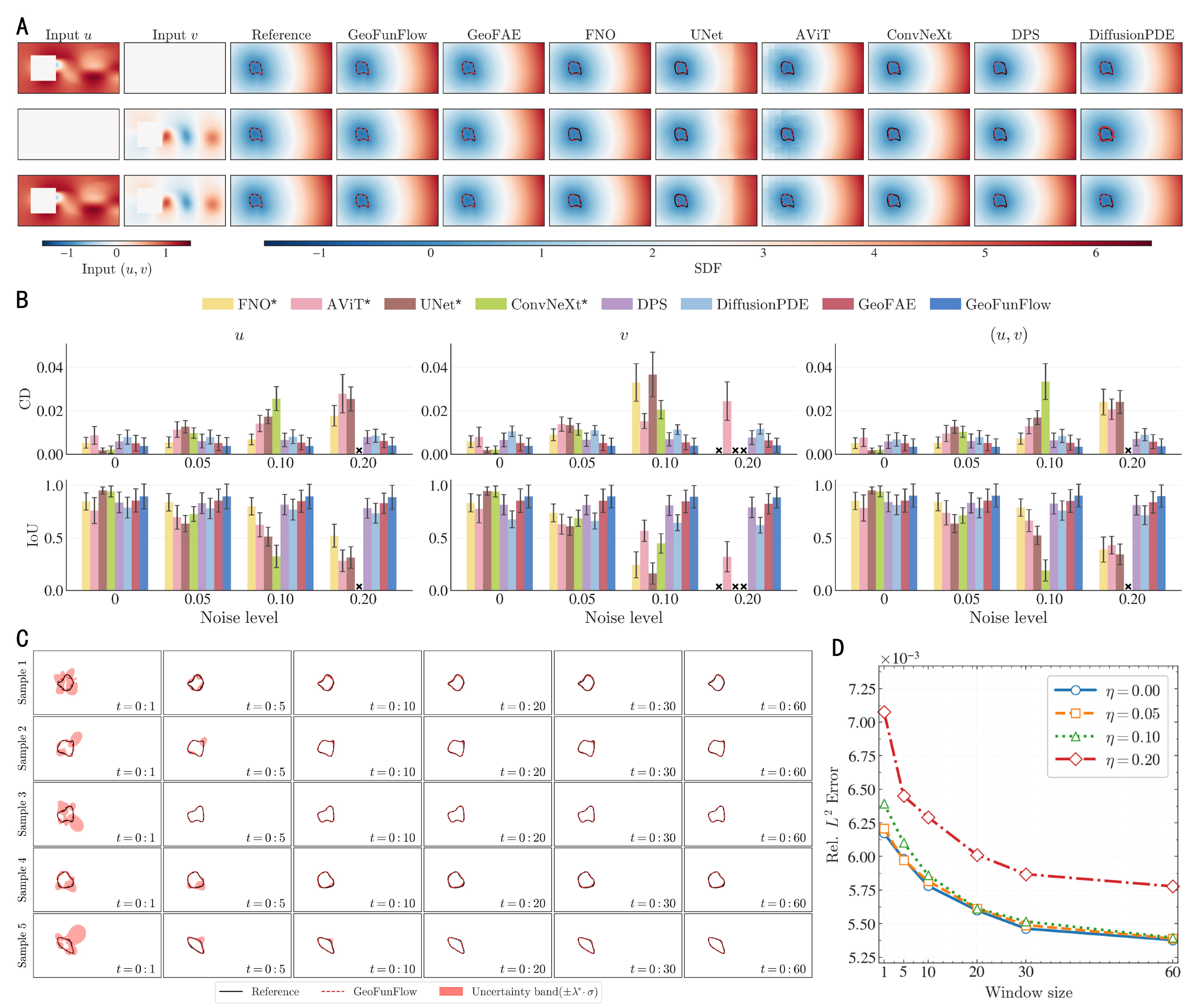}
    \caption{\textbf{Geometry inference results on the flow-past-objects (FPO) benchmark.} (\textbf{A}) Mean SDF predictions from different methods under three observation settings, corresponding to conditioning on only $u$, only $v$, or both velocity components $(u,v)$. GeoFunFlow recovers the obstacle shape most accurately across these settings. (\textbf{B}) Quantitative comparison using Chamfer distance (top) and IoU (bottom) across noise levels, demonstrating stronger robustness of GeoFunFlow. An asterisk ($\ast$) indicates that at least one test case for the corresponding method and noise level did not yield a converged geometry, whereas a cross ($\times$) indicates that none yielded a converged geometry. (\textbf{C}) Time-resolved predicted geometry boundaries and uncertainty bands for five test trajectories under joint conditioning on $(u,v)$, showing stable geometry estimates throughout the unsteady flow sequence. (\textbf{D}) Relative $L^2$ error of the predicted SDF versus temporal window size for different noise levels under joint conditioning on $(u,v)$, illustrating the accuracy-efficiency trade-off in aggregating temporal context.}
    \label{fig:fpo-results}
\end{figure}

The FPO results in Fig.~\ref{fig:fpo-results} demonstrate that GeoFunFlow remains robust in this more challenging unsteady setting. Fig.~\ref{fig:fpo-results}A evaluates GeoFunFlow in the more demanding unsteady FPO setting under different levels of velocity information. The recovered boundaries closely match the reference geometries whether the model is conditioned on either velocity component individually or on both components, showing that geometry recovery remains reliable even when only a single component is observed. In Fig.~\ref{fig:fpo-results}B, GeoFunFlow attains the lowest or near-lowest CD and the highest Intersection over Union (IoU) over nearly all noise levels and conditioning configurations, while the performance of several baselines deteriorates rapidly once the noise level reaches $10\%$ or $20\%$. This gap is particularly pronounced for single-channel conditioning, where relying on either $u$ or $v$ alone makes the inverse problem substantially less constrained. 

The temporal analysis in Figs.~\ref{fig:fpo-results}C and D, both based on conditioning with $(u,v)$, clarifies why windowed inference is beneficial in this benchmark. Because all snapshots within a trajectory correspond to the same obstacle, geometry predictions obtained at different times provide complementary evidence and can be ensembled to reduce snapshot-specific uncertainty. When only one or five snapshots are used, the predicted boundaries exhibit visibly broader uncertainty and larger variability. These effects diminish as the temporal window increases, resulting in sharper and more stable recovered contours. Fig.~\ref{fig:fpo-results}D further shows that enlarging the temporal window consistently reduces the relative $L^2$ error of the predicted SDF, with the largest gains at higher noise levels. Shorter observation windows are less costly, whereas longer windows provide richer temporal information and lead to more reliable geometry recovery. We therefore choose the window size of 10 for the reported FPO predictions, which captures most of the stability gain without incurring the cost of much longer temporal aggregation. In addition, the model reconstructs the associated $u$ and $v$ fields together with the hidden geometry. The corresponding field reconstructions and relative errors are reported in Appendix~\ref{app:results} (Figs.~\ref{fig:FPO_ufield} and~\ref{fig:FPO_vfield}; Table~\ref{tab:relative_l2_all_noise_case2}). These results show that GeoFunFlow is particularly strong for SDF recovery and becomes the most robust overall method under higher noise, where accurate field reconstruction and geometry inference must be achieved simultaneously.

\subsubsection{Optical tomography}

\begin{figure}[!ht]
    \centering
    \includegraphics[width=0.98\linewidth]{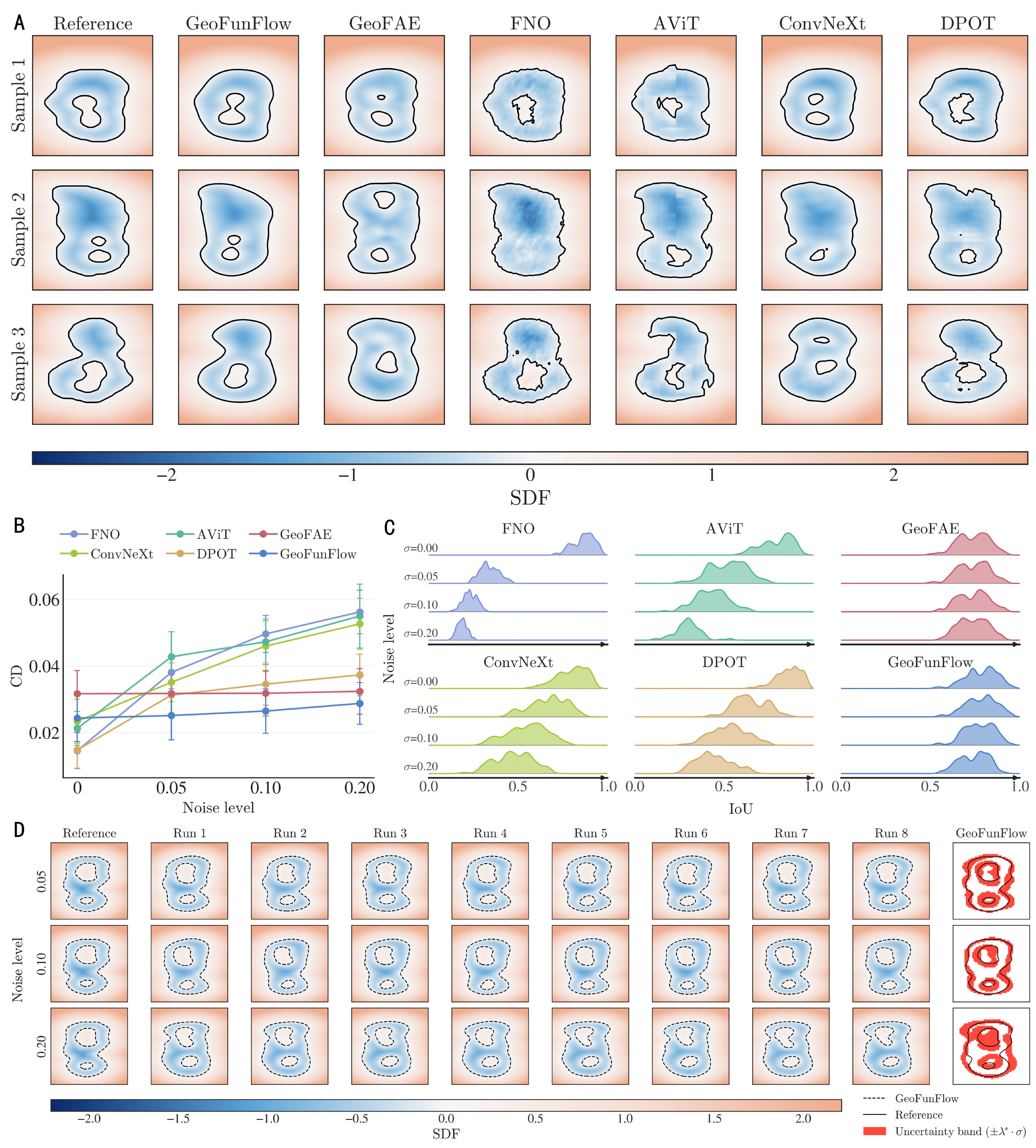}
    \caption{\textbf{Geometry recovery on the optical tomography (OT) benchmark.} (\textbf{A}) Visual comparison of reconstructed absorbing-inclusion geometries for three test samples. GeoFunFlow recovers both outer and inner boundaries more faithfully than the baselines. (\textbf{B}) Chamfer distance as a function of observation noise, showing that GeoFunFlow maintains the lowest or near-lowest reconstruction error. (\textbf{C}) Distribution of IoU scores across noise levels for all methods, where GeoFunFlow achieves the most favorable overlap with the ground-truth domain. (\textbf{D}) Multiple posterior samples from GeoFunFlow together with the resulting uncertainty band, illustrating calibrated shape uncertainty around the reconstructed inclusions.}
    \label{fig:ot-results}
\end{figure}

We finally evaluate GeoFunFlow on diffuse optical tomography (OT)~\cite{wang2025learning,carminati2021principles}. Boundary illumination generates a diffuse light field inside a scattering medium, and spatial variations in optical absorption alter the light measured at the boundary. Inferring the locations and morphologies of absorbing inclusions from these indirect, nonlocal responses is strongly ill-posed. Because the quantities of interest are the spatial extent, topology, and boundaries of the absorbing regions, this inverse problem can be viewed naturally as geometry inference. GeoFunFlow adopts this view and recovers the latent inclusion geometry directly from the boundary optical measurements.

Fig.~\ref{fig:ot-results} evaluates GeoFunFlow on the highly ill posed OT inverse problem, in which internal inclusions must be recovered solely from measurements collected by boundary sensors. These measurements are indirect responses to the unknown internal structures rather than pointwise observations of the reconstruction target. DPS and DiffusionPDE are therefore not directly comparable under this observation protocol. The qualitative results in Fig.~\ref{fig:ot-results}A show that GeoFunFlow reconstructs inclusions with more accurate topology and smoother boundaries than the deterministic baselines, particularly for samples containing multiple internal structures. As measurement noise increases, the CD curves in Fig.~\ref{fig:ot-results}B increase more slowly for GeoFunFlow, while the IoU distributions in Fig.~\ref{fig:ot-results}C remain shifted toward higher values. These results show that the learned latent representation retains stronger geometric consistency despite the indirect and noisy nature of the measurements. The posterior samples in Fig.~\ref{fig:ot-results}D further localize uncertainty around thin structures and boundary regions with high curvature, suggesting that these geometric features are less strongly constrained by the boundary measurements.

\subsection{Uncertainty quantification and calibration}

\begingroup
Uncertainty has a different practical role for physical fields and inferred geometries. For physical-field reconstruction, we ask whether the posterior standard deviation identifies regions with large reconstruction error. The error--uncertainty correlations in Table~\ref{tab:error_uq_main_compact} evaluate this property without assigning a coverage interpretation to the reported standard deviations. For a geometry, we first predict the corresponding SDF mean and standard deviation, from which we extract the uncertain region in which the boundary may plausibly lie. The raw uncertainty regions are consistently too narrow across LDC, FPO, and OT examples, so we apply post hoc calibration to bring their empirical coverage closer to the nominal levels. Specifically, for each dataset and noise level, we fit a constant $\lambda^\star$ to rescale the uncertainty on a held-out calibration subset by minimizing the root mean squared calibration error (RMSCE) over the nominal coverage levels~\cite{kuleshov2018accurate, moya2025conformalized, psaros2023uncertainty}. Scaling by $\lambda^\star$ substantially reduces RMSCE and brings empirical prediction interval coverage probability (PICP)~\cite{pearce2018high} closer to its nominal level. Details of the calibration procedure and evaluation are provided in Sec.~\ref{sec:uq-method}.
We first evaluate the uncertainty for the predicted SDF (Table~\ref{tab:uq_combined_by_dataset}, SDF rows).
The root mean squared calibration error RMSCE decreases from $0.358$--$0.673$ before calibration to $0.040$--$0.082$ after calibration. Avg.~PICP presents results averaged across the four nominal levels of $50\%$, $80\%$, $90\%$, and $95\%$, with an average nominal coverage of $78.75\%$. The corresponding level-specific results are provided in Table~\ref{tab:uq_picp_by_noise_level} of Appendix~\ref{app:results}. Raw PICP is below its nominal target in every setting, whereas calibration moves it substantially closer to the target. The accompanying increases in mean prediction interval width (MPIW)~\cite{pearce2018high} and boundary band width reflect the widening needed to correct the undercoverage of the raw uncertainty estimates. At the geometry level, the calibrated cover rate exceeds $0.93$ for LDC and OT at every noise level and remains between $0.899$ and $0.921$ for FPO.

Next, we evaluate the uncertainty for the predicted boundary or geometry (Table~\ref{tab:uq_combined_by_dataset}, Geometry rows). LDC and FPO are evaluated against reference contours, whereas OT is evaluated against the absorbing-inclusion region 
and therefore does not report a boundary band width. 
Calibration increases the boundary-based error--uncertainty correlation for both contour-based benchmarks at all reported noise levels. Overall, GeoFunFlow provides better-calibrated geometric coverage while retaining informative spatial alignment between uncertainty and reconstruction error across both contour- and region-based problems.
\endgroup

\begin{table*}[!ht]
\centering
\setlength{\tabcolsep}{3.5pt}
\caption{\textbf{Calibration of SDF and induced geometry uncertainty.} Entries are reported in the form raw $\rightarrow$ calibrated and as mean$\pm$std over 3 repeated calibration/holdout splits. The SDF block reports pointwise interval evaluation, and the geometry block reports the corresponding set-level evaluation.}
\label{tab:uq_combined_by_dataset}
\footnotesize
\begin{tabular}{@{}lllccc@{}}
\toprule
&  & Noise & \multicolumn{3}{c}{Metrics} \\
\midrule
\multirow{8}{*}{LDC}
& \multirow{4}{*}{SDF}
& & RMSCE $\downarrow$ & Avg.\ PICP & Avg.\ MPIW \\
& & 0.05 & $0.358 \pm 0.007 \rightarrow \mathbf{0.050 \pm 0.022}$ & $0.433 \pm 0.007 \rightarrow \mathbf{0.801 \pm 0.038}$ & $0.006 \pm 0.000 \rightarrow 0.017 \pm 0.002$ \\
& & 0.10 & $0.472 \pm 0.008 \rightarrow \mathbf{0.058 \pm 0.033}$ & $0.322 \pm 0.008 \rightarrow \mathbf{0.804 \pm 0.046}$ & $0.008 \pm 0.000 \rightarrow 0.033 \pm 0.004$ \\
& & 0.20 & $0.560 \pm 0.003 \rightarrow \mathbf{0.082 \pm 0.008}$ & $0.236 \pm 0.003 \rightarrow \mathbf{0.767 \pm 0.034}$ & $0.013 \pm 0.000 \rightarrow 0.086 \pm 0.011$ \\
\cmidrule(lr){2-6}
& \multirow{4}{*}{Geometry}
& & Cover Rate $\uparrow$ & Band Width & Correlation $\uparrow$ \\
& & 0.05 & $0.820 \pm 0.002 \rightarrow \mathbf{0.982 \pm 0.005}$ & $0.346 \pm 0.004 \rightarrow 0.971 \pm 0.111$ & $0.284 \pm 0.001 \rightarrow \mathbf{0.420 \pm 0.015}$ \\
& & 0.10 & $0.794 \pm 0.008 \rightarrow \mathbf{0.975 \pm 0.008}$ & $0.462 \pm 0.006 \rightarrow 1.842 \pm 0.218$ & $0.333 \pm 0.004 \rightarrow \mathbf{0.561 \pm 0.017}$ \\
& & 0.20 & $0.675 \pm 0.006 \rightarrow \mathbf{0.935 \pm 0.011}$ & $0.748 \pm 0.003 \rightarrow 4.410 \pm 0.464$ & $0.405 \pm 0.002 \rightarrow \mathbf{0.634 \pm 0.003}$ \\
\midrule
\multirow{8}{*}{FPO}
& \multirow{4}{*}{SDF}
& & RMSCE $\downarrow$ & Avg.\ PICP & Avg.\ MPIW \\
& & 0.05 & $0.649 \pm 0.003 \rightarrow \mathbf{0.059 \pm 0.010}$ & $0.151 \pm 0.003 \rightarrow \mathbf{0.764 \pm 0.017}$ & $0.006 \pm 0.000 \rightarrow 0.054 \pm 0.004$ \\
& & 0.10 & $0.627 \pm 0.002 \rightarrow \mathbf{0.061 \pm 0.007}$ & $0.172 \pm 0.002 \rightarrow \mathbf{0.769 \pm 0.008}$ & $0.006 \pm 0.000 \rightarrow 0.054 \pm 0.003$ \\
& & 0.20 & $0.617 \pm 0.003 \rightarrow \mathbf{0.059 \pm 0.003}$ & $0.181 \pm 0.003 \rightarrow \mathbf{0.770 \pm 0.004}$ & $0.007 \pm 0.000 \rightarrow 0.058 \pm 0.004$ \\
\cmidrule(lr){2-6}
& \multirow{4}{*}{Geometry}
& & Cover Rate $\uparrow$ & Band Width & Correlation $\uparrow$ \\
& & 0.05 & $0.389 \pm 0.015 \rightarrow \mathbf{0.899 \pm 0.012}$ & $0.210 \pm 0.019 \rightarrow 1.923 \pm 0.206$ & $0.219 \pm 0.008 \rightarrow \mathbf{0.428 \pm 0.014}$ \\
& & 0.10 & $0.427 \pm 0.014 \rightarrow \mathbf{0.905 \pm 0.008}$ & $0.219 \pm 0.018 \rightarrow 1.913 \pm 0.164$ & $0.228 \pm 0.008 \rightarrow \mathbf{0.460 \pm 0.015}$ \\
& & 0.20 & $0.503 \pm 0.012 \rightarrow \mathbf{0.921 \pm 0.003}$ & $0.264 \pm 0.018 \rightarrow 2.081 \pm 0.155$ & $0.257 \pm 0.008 \rightarrow \mathbf{0.515 \pm 0.015}$ \\
\midrule
\multirow{8}{*}{OT}
& \multirow{4}{*}{SDF}
& & RMSCE $\downarrow$ & Avg.\ PICP & Avg.\ MPIW \\
& & 0.05 & $0.596 \pm 0.005 \rightarrow \mathbf{0.056 \pm 0.011}$ & $0.201 \pm 0.004 \rightarrow \mathbf{0.805 \pm 0.031}$ & $0.099 \pm 0.001 \rightarrow 0.701 \pm 0.067$ \\
& & 0.10 & $0.618 \pm 0.005 \rightarrow \mathbf{0.040 \pm 0.004}$ & $0.181 \pm 0.004 \rightarrow \mathbf{0.788 \pm 0.022}$ & $0.094 \pm 0.001 \rightarrow 0.689 \pm 0.048$ \\
& & 0.20 & $0.673 \pm 0.002 \rightarrow \mathbf{0.041 \pm 0.007}$ & $0.128 \pm 0.002 \rightarrow \mathbf{0.769 \pm 0.016}$ & $0.081 \pm 0.001 \rightarrow 0.792 \pm 0.042$ \\
\cmidrule(lr){2-6}
& \multirow{4}{*}{Geometry}
& & Cover Rate $\uparrow$ & Band Width & Correlation $\uparrow$ \\
& & 0.05 & $0.671 \pm 0.004 \rightarrow \mathbf{0.973 \pm 0.007}$ &-& $0.444 \pm 0.003 \rightarrow \mathbf{0.579 \pm 0.011}$ \\
& & 0.10 & $0.648 \pm 0.002 \rightarrow \mathbf{0.970 \pm 0.005}$ &-& $0.442 \pm 0.002 \rightarrow \mathbf{0.585 \pm 0.007}$ \\
& & 0.20 & $0.563 \pm 0.003 \rightarrow \mathbf{0.962 \pm 0.006}$ &-& $0.400 \pm 0.003 \rightarrow \mathbf{0.561 \pm 0.007}$ \\
\bottomrule
\end{tabular}
\end{table*}

\section{Discussion}\label{sec3}

This work introduces GeoFunFlow as a unified geometry-aware generative framework for inverse problems on complex domains. Conceptually, it moves beyond the conventional inverse map from observations to a single physical state, $\mathbf{y}\mapsto\mathbf{u}$, and its geometry-conditioned form, $(\mathbf{y},\Omega)\mapsto\mathbf{u}$, by learning a joint conditional distribution over physical states and their supporting geometries, $\mathbf{y}\mapsto p(\mathbf{u},\Omega\mid\mathbf{y})$, when the domain is not fully known. GeoFunFlow realizes this formulation by combining a geometric function autoencoder with latent rectified-flow modeling, allowing field reconstruction on prescribed domains and joint field--geometry inference on unknown domains to be handled within the same representation and inference pipeline. Across benchmarks spanning regular and irregular discretizations, steady and unsteady settings, and fixed, variable, and unknown geometries, the results show that this formulation supports accurate reconstruction together with spatially resolved uncertainty estimates for both physical fields and inferred geometry. These findings suggest that modeling physical states and geometry as a coupled conditional distribution provides a useful direction for inverse operator learning beyond deterministic prediction.

At the same time, several limitations remain. First, the current SDF decoder is trained through reconstruction supervision without explicitly enforcing the Eikonal constraint $\|\nabla\varphi\|_2=1$. The predicted zero level set remains directly usable for geometry recovery, as demonstrated by the experiments, but the decoded field is not guaranteed to retain exact signed-distance structure away from the boundary. Second, the current geometry calibration uses one global multiplier for the SDF uncertainty in each dataset--noise setting. This corrects overall interval width but cannot address spatially heterogeneous miscalibration, local changes in boundary ambiguity, or distribution shift beyond the held-out calibration split. Third,  the current framework learns from data alone and does not explicitly enforce PDE or conservation laws during posterior generation. This improves flexibility across diverse discretizations, but it may also limit physical fidelity in regimes where training data are scarce or where extrapolation beyond the training distribution is required. This distinction is particularly relevant to the metric interpretation of SDF-derived uncertainty.

These limitations point to several directions for future work. First, Eikonal regularization or SDF reinitialization could promote signed-distance consistency while preserving the zero-level-set representation used for geometry recovery. Second, the global SDF multiplier could be replaced by spatially adaptive calibration to account for local variation in boundary ambiguity, or by conformal procedures that provide finite-sample coverage guarantees under suitable assumptions. Third, physics-aware constraints could be incorporated into latent generative inference through PDE residual regularization or differentiable solvers. Beyond these methodological extensions, the probabilistic posterior over physical fields and, in geometry-inference settings, domain geometry provides a basis for downstream tasks such as active sensing, experimental design, and decision-making under uncertainty. Together, these directions could extend GeoFunFlow toward more reliable, geometry-aware probabilistic inference for PDE-governed systems.

\section{Methods}\label{sec4}

\begingroup
The conceptual setting and an overview of GeoFunFlow are given in Sec.~\ref{sec:problem_overview}. Next, we formalize the inverse problem (Sec.~\ref{sec:problem_formulation}), specify the network architecture and algorithm (Sec.~\ref{sec:architecture}), and introduce SDF-uncertainty calibration (Sec.~\ref{sec:uq-method}). 

\subsection{Problem setup}\label{sec:problem_formulation}
We formulate field reconstruction on known domains and joint field--geometry inference as a common conditional inference problem. Let
$\mathbf u$ denote the physical fields, such as velocity and pressure.
In the known-geometry setting, the domain
$\Omega\subset\mathbb R^d$ is prescribed for each instance, although
its shape may vary between instances. The task is to reconstruct
$\mathbf u$ on this domain. In the unknown-geometry setting, $\Omega$
is a fixed ambient domain containing an unknown region
$\Omega_g\subset\Omega$, such as an obstacle or absorbing inclusion.
The task is to infer both $\mathbf u$ and the geometry of $\Omega_g$. Although geometry is typically provided as a discrete boundary or mesh, we represent $\Omega_g$ by its SDF $\varphi$,
\[
\varphi(\mathbf x)=
\begin{cases}
-\operatorname{dist}(\mathbf x,\partial\Omega_g),
  & \mathbf x\in\operatorname{int}(\Omega_g),\\
0, & \mathbf x\in\partial\Omega_g,\\
\operatorname{dist}(\mathbf x,\partial\Omega_g),
  & \mathbf x\notin\Omega_g,
\end{cases}
\]
where
$\operatorname{dist}(\mathbf x,\partial\Omega_g)
=\inf_{\mathbf y\in\partial\Omega_g}\|\mathbf x-\mathbf y\|_2$.
This gives a continuous spatial representation of geometry that can be inferred alongside the physical fields, and the zero level set of $\varphi$ defines the boundary of $\Omega_g$.

To express both tasks in a common form, let $\mathbf f=\mathbf u$ for
known geometry and $\mathbf f=(\mathbf u,\varphi)$ for unknown geometry. The condition $\mathbf c$
contains the available measurements, their locations, and the observed
variables. These measurements may be sparse, noisy, downsampled, or a
combination of these. Because the same condition may be consistent with
multiple target fields, we seek a conditional distribution rather than
a single reconstruction. The two inverse maps are
\[
\begin{aligned}
(\mathbf c,\Omega)
&\longmapsto
\operatorname{Law}\!\left(\mathbf u\mid\mathbf c,\Omega\right)
&&\text{(known geometry)},\\
\mathbf c
&\longmapsto
\operatorname{Law}\!\left((\mathbf u,\varphi)\mid\mathbf c\right)
&&\text{(unknown geometry)}.
\end{aligned}
\]
We denote the target conditional distribution in the two settings by $P^\star(\cdot\mid\mathbf c, \Omega)$ and
$P^\star(\cdot\mid\mathbf c)$. Furthermore, the prescribed domain can be understood as
an additional condition in the known-geometry setting, so we use $P^\star(\cdot\mid\mathbf c)$ for both tasks.

To handle different measurement scenarios, we represent the observations on a set of points. Specifically, let $X=\{\mathbf x_i\}_{i=1}^{n}$ denote a set of $n$ $d$-dimensional points in the computational domain, and let $C$ denote the dimension of $\mathbf f(\mathbf x_i)$. Then we define $\mathbf F=[\mathbf f(\mathbf x_i)]_{i=1}^{n}\in\mathbb R^{n\times C}.$ Because we only know partial observations for some fields, we use a binary observation mask
$\mathbf M\in\{0,1\}^{n\times C}$
to specify which of the $C$ variable channels are observed at each
location, such as whether a velocity component or pressure is available.
Therefore, the condition is
\[
\mathbf c
=
(X,\mathbf M,\mathbf F_{\text{obs}}),
\qquad
\mathbf F_{\text{obs}}
=
\mathbf M\odot\mathbf F,
\]
where $\odot$ denotes element-wise multiplication. 

For known geometry, the prescribed $X$ discretization provides the domain.
For unknown geometry, the SDF $\varphi$ is part of the target field $\mathbf f$ but is not
observed. The geometry must therefore be inferred from the available physical-field
observations.

\subsection{Network architecture}\label{sec:architecture}
GeoFunFlow approximates the conditional distributions defined above using a geometric function autoencoder (GeoFAE) and a conditional velocity model $\mathbf g_\psi$. GeoFAE consists of a geometry-aware encoder $E_\theta$ and a continuous-coordinate decoder $D_\phi$. The encoder maps variable-sized observations to a fixed number of latent tokens, which is used as the condition of a rectified flow $\mathbf g_\psi$ to generate latent samples of the full target field. The decoder then maps these latent samples to physical fields and geometries at arbitrary query coordinates.

\subsubsection{Geometric function encoder}
The encoder follows the same three-step pipeline for both regular and irregular discretizations (Fig.~\ref{fig:framework}B):
\[
\mathbf c
\xrightarrow{\text{Tokenization}}
\mathbf H\in\mathbb R^{m\times d_z}
\xrightarrow{\text{Perceiver}}
\mathbf Z_{\rm agg}\in\mathbb R^{N_\ell\times d_z}
\xrightarrow{\text{Transformer}}
\mathbf z\in\mathbb R^{N_\ell\times d_z}.
\]
Here $m$ is the number of input tokens, $d_z$ is the token width, and $N_\ell$ is a fixed number of latent tokens independent of the input resolution. A Perceiver module \cite{jaegle2021perceiver} uses $N_\ell$ learned latent queries to aggregate the variable-size token set, after which $L_{\rm enc}$ Transformer blocks refine the latent representation. Regular and irregular inputs therefore differ only in their initial tokenization, with both subsequently processed by the same Perceiver aggregation and Transformer refinement. Thus, the cost of latent self-attention is controlled by $N_\ell$ rather than by the number of mesh nodes or grid patches. The underlying attention operations are summarized in Appendix~\ref{app:preliminaries}. 

\par\medskip\noindent\textbf{Tokenization on irregular discretizations.}
For a mesh or point cloud, each retained coordinate in $X$ defines one token, so $m=|S|$. We embed the coordinate and the masked observations separately and concatenate the resulting features:
\[
\mathbf H=
\left[
\gamma_{\rm geo}(\mathbf X),
\gamma_{\rm obs}([\mathbf M,\mathbf F_{\text{obs}}])
\right].
\]
The map $\gamma_{\rm geo}$ is a random Fourier feature ~\cite{tancik2020fourier} embedding of the coordinates, whereas $\gamma_{\rm obs}$ is a multilayer perceptron applied to the mask--value pair. The coordinates carry the sampled geometry of an irregular domain, while the observation features identify the physical or geometric variables available at each point. 

\par\medskip\noindent\textbf{Tokenization on regular grids.}
For a field sampled on an $H\times W$ grid of $\Omega$, we stack $\mathbf M$ and $\mathbf F_{\text{obs}}$ along the channel dimension and partition the resulting array into non-overlapping $P_h\times P_w$ patches. A convolutional patch embedding maps each patch to $\mathbb R^{d_z}$, producing
\[
m=\frac{H}{P_h}\times\frac{W}{P_w}
\]
tokens, to which a two-dimensional sinusoidal positional encoding is added. 

\subsubsection{Continuous-coordinate decoder}
The decoder builds on CViT~\cite{wang2025cvit} and evaluates the generated function at arbitrary query coordinates. For a query set $\mathcal Q=\{\mathbf q_j\}_{j=1}^{M_q}$, random Fourier features produce the initial query embeddings $\mathbf Q_0$. The embeddings are then updated through $L_{\rm dec}$ cross-attention blocks using the encoder output $\mathbf z$ as keys and values:
\begin{align*}
\mathbf Q'_{\ell}
&=\mathbf Q_{\ell-1}
+\operatorname{MHA}\!\left(
\operatorname{LN}(\mathbf Q_{\ell-1}),
\operatorname{LN}(\mathbf z),
\operatorname{LN}(\mathbf z)\right),\\
\mathbf Q_{\ell}
&=\mathbf Q'_{\ell}
+\operatorname{MLP}\!\left(\operatorname{LN}(\mathbf Q'_{\ell})\right),
\qquad \ell=1,\ldots,L_{\rm dec}.
\end{align*}
Here MHA, LN, and MLP denote multi-head attention, layer normalization, and a feedforward network,
respectively. A final feedforward projection maps each row of $\mathbf Q_{L_{\rm dec}}$ to the $C$ modeled variables, giving
\[
D_\phi(\mathbf z,\mathcal Q)
=\left[\widehat{\mathbf f}(\mathbf q_j)\right]_{j=1}^{M_q}
\in\mathbb R^{M_q\times C}.
\]
Because $\mathcal Q$ is independent of the input discretization, the same latent state can be decoded on the original points, at sensor locations, or on a different output resolution.

\subsubsection{Conditional latent rectified flow}
GeoFunFlow uses rectified flow~\cite{liu2023flow} to model the conditional distribution. The encoder summarizes the partial observations as $\mathbf z_{\mathbf c}=E_\theta(\mathbf c)$. Conditioned on this representation, a Diffusion Transformer (DiT)~\cite{peebles2023scalable} parameterizes the time-dependent velocity
\[
\mathbf g_\psi(\mathbf z_t,t,\mathbf z_{\mathbf c})
\in\mathbb R^{N_\ell\times d_z}.
\]
Here $\mathbf z_t$ is the latent state transported by the flow at time $t$, whereas $\mathbf z_{\mathbf c}$ remains fixed as the conditioning representation. Because $\mathbf z_{\mathbf c}$ encodes both the observed values and their availability mask $\mathbf M$, the same conditional-flow interface applies to supplied geometry, missing physical variables, and latent geometry, eliminating the need for setting-specific conditioning rules. Appendix~\ref{app:theory} formalizes this conditional transport in function space and gives the corresponding posterior-approximation bounds.

\subsection{Training and inference}
\subsubsection{Two-stage training}\label{sec:twostage_training}
GeoFunFlow is trained in two stages. First, GeoFAE learns to reconstruct the target function from both partial and full inputs. For each instance
$i$ in a minibatch of size $B$, we sample a partial condition
$\mathbf c_i$ using the observation process defined above, along with
query coordinates $\mathcal Q_i=\{\mathbf q_{i,j}\}_{j=1}^{M_q}.$
The corresponding full input is
$\mathbf c_i^{\mathrm{full}}=(X_i,\mathbf 1,\mathbf F_i)$,
where $\mathbf 1$ is an all-ones observation mask. GeoFAE encodes both the partial and full inputs and reconstructs the same
target function from the resulting latent representations,
\begin{equation*}
\mathbf z_{i,\mathbf c} = E_\theta(\mathbf c_i),
\qquad
\mathbf z_{i,1} = E_\theta(\mathbf c_i^{\mathrm{full}}),
\end{equation*}
with the corresponding reconstructions
\begin{equation*}
\widehat{f}_{i,\mathbf c}(q_{i,j})
=
D_\phi(z_{i,\mathbf c},q_{i,j}),
\qquad
\widehat{f}_{i,1}(q_{i,j})
=
D_\phi(z_{i,1},q_{i,j}).
\end{equation*}
The reconstruction loss is
\begin{equation*}
\mathcal{L}_{\mathrm{FAE}}
=
\frac{1}{2BM_qC}
\sum_{i=1}^{B}
\sum_{j=1}^{M_q}
\left[
\left\|
f_i(q_{i,j})
-
\widehat{f}_{i,\mathbf c}(q_{i,j})
\right\|_2^2
+
\left\|
f_i(q_{i,j})
-
\widehat{f}_{i,1}(q_{i,j})
\right\|_2^2
\right].
\end{equation*}
The two reconstruction terms are weighted equally. Partial inputs cover
varying observation settings, including sparse or downsampled measurements,
missing variable channels, and measurement noise. This joint training enables
GeoFAE to reconstruct the target function from partial observations while
also providing latent representations of the complete target functions
needed in the second stage.

We then freeze GeoFAE and train the conditional latent flow using $\mathbf z_{i,1}$ as the target latent and $\mathbf z_{i,\mathbf c}$ as the conditioning latent.
When geometry is unknown, the latent $\mathbf z_{i,1}$ encodes the full physical fields together with the SDF. The latent $\mathbf z_{i,c}$ encodes
the available observations. The rectified flow learns to generate the former
conditioned on the latter. To train the flow, we draw
\[
\mathbf z_{i,0}\sim\mathcal N(\mathbf 0,\mathbf I),
\qquad
t_i\sim\operatorname{U}(0,1),
\]
and interpolate between noise and the target latent:
\[
\mathbf z_{i,t_i}
=
(1-t_i)\mathbf z_{i,0}
+
t_i\mathbf z_{i,1}.
\]
The conditional velocity model minimizes the rectified-flow loss
\cite{liu2023flow}
\[
\mathcal L_{\rm CRF}
=
\frac{1}{B N_\ell d_z}
\sum_{i=1}^{B}
\left\|
\mathbf g_\psi
\left(
\mathbf z_{i,t_i},
t_i,
\mathbf z_{i,c}
\right)
-
\left(
\mathbf z_{i,1}-\mathbf z_{i,0}
\right)
\right\|^2.
\]
Thus, the model learns the distribution of full target latents
conditioned on the encoded partial observations.

\subsubsection{Latent-space regularization}\label{sec:latentspace_regularization}
The conditional flow may generate latents that differ slightly from
those produced by the GeoFAE encoder on training data. To make the
decoder robust to these deviations, we perturb
$\mathbf z=E_\theta(\mathbf c)$ before reconstruction,
\[
\overline{\mathbf z}
=\operatorname{Drop}_{\rm channel}\!\left(
\mathbf z+\rho\,s(\mathbf z)\odot\boldsymbol\xi
\right),
\qquad
\boldsymbol\xi\sim\mathcal N(\mathbf0,\mathbf I),
\quad
\rho\sim\operatorname{U}(0.01,0.1).
\]
Here $s(\mathbf z)$ is the feature-wise standard deviation across latent tokens, which adapts the perturbation magnitude to each latent channel. The operator $\operatorname{Drop}_{\rm channel}$ applies a random channel-wise dropout mask shared across tokens, discouraging the decoder from depending on a small subset of latent features. During GeoFAE training, $\overline{\mathbf z}$ replaces $\mathbf z$ in the reconstruction loss. No latent perturbation or dropout is applied at inference.

\subsubsection{Posterior sampling}\label{sec:posteriorsampling}
At inference, the partial observations are encoded once as $\mathbf z_{\mathbf c}=E_\theta(\mathbf c)$. For each posterior sample, we draw independent Gaussian noise and integrate the learned ordinary differential equation
\begin{align*}
\mathbf z(0)&\sim\mathcal N(\mathbf0,\mathbf I),
&
\frac{d\mathbf z(t)}{dt}
&=\mathbf g_\psi(\mathbf z(t),t,\mathbf z_{\mathbf c}),
\qquad t\in[0,1].
\end{align*}
The terminal state is decoded at the desired coordinates,
\[
\widehat{\mathbf f}(\mathbf q)=D_\phi(\mathbf z(1),\mathbf q).
\]
Repeating the integration from independent initial states produces samples from the learned conditional posterior. If the output includes the SDF $\varphi$, its zero level set yields a sample of the recovered geometry boundary. This execution path is unchanged across field reconstruction, missing-variable inference, and joint geometry--field recovery.

\subsection{Uncertainty quantification}\label{sec:uq-method}
The sampling procedure above maps a fixed partial observation to an ensemble of field--geometry realizations. For one scalar output channel, let $\widehat f^{(s)}(\mathbf q)$ denote the $s$-th decoded sample, with $s=1,\ldots,S$. We summarize the learned conditional posterior by
\[
\mu(\mathbf q)=\frac{1}{S}\sum_{s=1}^{S}\widehat f^{(s)}(\mathbf q),
\qquad
\sigma^{2}(\mathbf q)
=\frac{1}{S}\sum_{s=1}^{S}
\left(\widehat f^{(s)}(\mathbf q)-\mu(\mathbf q)\right)^{2}.
\]
Here $\mu(\mathbf q)$ is the posterior-mean estimate and $\sigma(\mathbf q)$ describes local variation across posterior samples. When the decoder returns multiple physical and geometric variables, these quantities are computed separately for each output component, measuring field uncertainty for physical outputs and geometric uncertainty of the SDF.

\par\medskip\noindent\textbf{Physical-field uncertainty.}
For physical-field reconstruction, we assess whether posterior uncertainty is spatially aligned with reconstruction error. Specifically, we compute the Pearson correlation between the predictive standard deviation $\sigma(\mathbf q)$ and the pointwise absolute reconstruction error. A larger positive correlation indicates that locations with larger reconstruction errors tend to be assigned greater posterior uncertainty. This metric therefore measures spatial error–uncertainty alignment rather than reconstruction accuracy or the calibration of uncertainty magnitudes.

\par\medskip\noindent\textbf{Calibrated geometry uncertainty.}
Whereas physical-field uncertainty is evaluated through its spatial alignment with reconstruction error, geometry uncertainty must additionally support a coverage interpretation. When geometry is represented by the SDF $\varphi$, the zero level set of each posterior sample defines one plausible boundary. The spatial variation of $\sigma(\mathbf q)$ therefore identifies where the boundary is less constrained, but its raw magnitude does not necessarily provide the intended coverage. We calibrate this magnitude through a single positive scale factor $\lambda$. For a target miscoverage level $\alpha$, we construct the normal-reference SDF interval
\[
I_{\lambda}^{(\alpha)}(\mathbf q)
=\left[
\mu(\mathbf q)-z_{1-\alpha/2}\lambda\sigma(\mathbf q),\;
\mu(\mathbf q)+z_{1-\alpha/2}\lambda\sigma(\mathbf q)
\right].
\]
The coefficient $z_{1-\alpha/2}$ is the standard-normal quantile associated with a two-sided interval. For example, $1-\alpha=95\%$ gives $z_{0.975}\approx1.96$. For each dataset and noise level, $\lambda^\star$ is fitted on a held-out calibration subset by minimizing the root mean squared calibration error (RMSCE) over the nominal coverage levels used in evaluation, and all reported calibration results are computed on a disjoint holdout subset.

The plausible boundary band is defined as the set of coordinates whose calibrated posterior SDF intervals $I_{\lambda^{\star}}^{(\alpha)}(\mathbf q)$ include zero, indicating geometry uncertainty. Because $\lambda$ is shared across coordinates, calibration adjusts the overall width of this band without changing the spatial ordering of $\sigma(\mathbf q)$. We characterize the resulting uncertainty using complementary metrics for calibration, interval sharpness, geometric coverage, and spatial alignment, including PICP, RMSCE, MPIW, geometry cover rate, band width, and correlation with geometric error. Definitions and computation details for these metrics are provided in Appendix~\ref{app:metrics}. The global scale calibration is intentionally lightweight, using only aggregate empirical coverage to adjust a single scale factor without imposing pointwise constraints or exploiting additional spatial information.
\endgroup

\bibliography{reference.bib}

\backmatter

\section*{Data availability}

The datasets used in this study are publicly available or generated using the numerical procedures described in the Methods and Supplementary Information. The processed datasets used for training and evaluation will be made publicly available upon publication.

\section*{Code availability}

The source code for GeoFunFlow will be made publicly available upon publication at \url{https://github.com/lu-group/geofunflow}.

\section*{Supplementary information}

Supplementary information is available for this paper.

\section*{Acknowledgments}

This work was supported by the U.S. Department of Energy Office of Advanced Scientific Computing Research under Grants No.~DE-SC0025593 and No.~DE-SC0025592 and the U.S. National Science Foundation under Grants No.~DMS-2347833 and No.~DMS-2527294. This research used resources of the National Energy Research Scientific Computing Center (NERSC), a Department of Energy User Facility (project m4946).

\begin{appendices}
\renewcommand{\theHequation}{app.\Alph{section}.\arabic{equation}}
\renewcommand{\theHfigure}{app.\Alph{section}.\arabic{figure}}
\renewcommand{\theHtable}{app.\Alph{section}.\arabic{table}}
\clearpage
\section{Technical background}
\label{app:technical}

\subsection{Notation}
\label{app:notations}

Table~\ref{notation} summarizes the main symbols and notations used in this work.

\begingroup
\small
\renewcommand{\arraystretch}{1.05}
\begin{longtable}{|>{\centering\arraybackslash}m{0.23\textwidth}|>{\arraybackslash}m{0.7\textwidth}|}
\caption{\textbf{Notation used throughout the main text and appendices.} }
\label{notation}\\
\hline
\rowcolor{gray!30}
\textbf{Notation} & \multicolumn{1}{c}{\textbf{Description}} \\ \hline
\endfirsthead
\hline
\rowcolor{gray!30}
\textbf{Notation} & \multicolumn{1}{c}{\textbf{Description}} \\ \hline
\endhead
\multicolumn{2}{|>{\columncolor{gray!15}}c|}{\textbf{Problem Setup}} \\ \hline
$\Omega \subset \mathbb{R}^d$ & Bounded spatial domain embedded in $d$-dimensional Euclidean space. \\

$X=\{\mathbf x_i\}_{i=1}^{n}$
& Set of $n$ input points in the computational domain. \\

$\mathbf f(\mathbf x),\ \mathbf F$
& Target function and its values on $X$,
  $\mathbf F=[\mathbf f(\mathbf x_i)]_{i=1}^{n}
  \in\mathbb R^{n\times C}$. \\

$C$
& Number of variable channels in $\mathbf f$. \\

$\varphi$
& Signed distance function whose zero level set defines the inferred boundary. \\

$\mathbf M,\ \mathbf F_{\rm obs}$
& Observation mask and masked values,
  $\mathbf F_{\rm obs}=\mathbf M\odot\mathbf F$. \\

$\mathbf c=(X,\mathbf M,\mathbf F_{\rm obs})$
& Partial condition used to construct the conditioning latent $z_c$. \\

$\mathbf c_{\rm full}=(X,\mathbf1,\mathbf F)$
& Complete input used to construct the complete-state target latent $z_1$. \\

$P^\star(\cdot\mid\mathbf c,\Omega),\
 P^\star(\cdot\mid\mathbf c)$
& Target conditional distributions for known and unknown geometry,
  respectively. In the known-geometry setting, the supplied domain
  is implicit when writing $P^\star(\cdot\mid\mathbf c)$. \\
\hline
\multicolumn{2}{|>{\columncolor{gray!15}}c|}{\textbf{Rectified Flow}} \\ \hline
$x_t$ & Generic state along an interpolation or flow at time $t\in[0,1]$. \\
$v_\vartheta$  & Generic velocity field with parameters $\vartheta$. \\
$\pi_0, \pi_1$ & Base distribution, typically $\mathcal N(0,I)$, and data distribution. \\
$\mathcal L_{\rm RF}(\vartheta),\ \mathcal L_{\rm CRF}(\vartheta)$ & Generic unconditional and conditional rectified-flow objectives. \\
$p(x\mid c),\ p_\vartheta(x\mid c)$ & True and modeled conditional distributions in the generic formulation. \\
\hline
\multicolumn{2}{|>{\columncolor{gray!15}}c|}{\textbf{GeoFunFlow}} \\ \hline
$m,\ N_\ell,\ d_z$ & Numbers of input and latent tokens, and token width, respectively. \\
$\gamma_{\rm geo},\ \gamma_{\rm obs},\ \mathbf h_i,\ \mathbf H$ & Coordinate and observation embeddings, an input token, and the resulting input-token array. \\
$L_{\rm enc},\ L_{\rm dec}$ & Numbers of encoder Transformer blocks and decoder cross-attention blocks. \\
$\mathbf Z_q,\ \mathbf Z_{\rm agg},\ \mathbf U_\ell$ & Learned Perceiver queries, aggregated tokens, and encoder Transformer states. \\
$\mathcal Q,\ M_q,\ \mathbf Q_\ell$ & Query-coordinate set, number of queries, and decoder query embeddings at block $\ell$. \\
$\mathrm{MHA},\ \mathrm{MSA},\ \mathrm{MLP},\ \mathrm{LN}$ & Multi-head attention, multi-head self-attention, multilayer perceptron, and layer normalization. \\
$E_\theta,\ D_\phi$& GeoFAE encoder and continuous-coordinate decoder.\\
$\mathbf z,\ \mathbf z_0,\ \mathbf z_1,\ \mathbf z_t,\ \mathbf z_{\mathbf c}$ & Encoded, base-noise, target, interpolated, and conditioning latents. \\
$\mathbf g_\psi$ & DiT-parameterized conditional rectified-flow velocity field.\\
$\mathcal L_{\rm FAE},\ \mathcal L_{\rm CRF}$ & GeoFAE reconstruction and conditional rectified-flow objectives.\\
$\overline{\mathbf z},\ \operatorname{Drop}_{\rm channel},\ \rho,\ \boldsymbol\xi,\ s(\mathbf z)$ & Regularized latent, channel-wise dropout, perturbation scale, Gaussian perturbation, and feature-wise latent standard deviation.\\
\hline
\multicolumn{2}{|>{\columncolor{gray!15}}c|}{\textbf{Uncertainty Quantification}} \\ \hline
$S,\ \widehat f^{(s)}(\mathbf q)$ & Posterior-ensemble size and the $s$-th decoded sample of a scalar output channel. \\
$\mu(\mathbf q),\ \sigma(\mathbf q)$ & Empirical posterior mean and predictive standard deviation at query coordinate $\mathbf q$. \\
$\alpha,\ z_{1-\alpha/2}$ & Target miscoverage and the corresponding standard-normal quantile. \\
$I_{\lambda}^{(\alpha)}(\mathbf q)$ & Normal-reference interval for the SDF after global spread scaling. Setting $\lambda=1$ gives the raw interval. \\
$\lambda,\ \lambda^\star$ & Global SDF-uncertainty scale and the value selected on calibration data by minimizing RMSCE. \\
\hline
\multicolumn{2}{|>{\columncolor{gray!15}}c|}{\textbf{Theory}} \\ \hline
$\mathcal H(\Omega),\ a_{\mathbf u},\ a_\varphi$ & Complete-state function space and the physical-field and SDF weights in its product norm. \\
$\mathcal Z,\ \|\cdot\|_{\mathcal Z},\ T_\Omega,\ L_D$ & Latent space, normalized Frobenius norm, decode-to-function map, and its Lipschitz constant. \\
$P^\star,\ P_Z^\star$ & True function posterior and target complete-state latent posterior. \\
$Q_Z^\psi,\ Q_{Z,h}^\psi,\ \widehat P_h$ & Continuous-flow latent law, its numerical approximation with step size $h$, and the decoded model posterior. \\
$\mathcal P_2(\cdot),\ W_2^\Omega,\ W_2^{\mathcal Z}$ & Finite-second-moment probability measures and 2-Wasserstein distances on the function and latent spaces.\\
$\mathcal H_{\rm reg},\ \mathcal A_X$ & Pointwise-regular function space and the sampling--reconstruction composition on $X$. \\
$\mathcal S_X,\ \mathcal R_X,\ \epsilon_{\rm disc}$ & Sampling and reconstruction operators and the resulting discrete-to-continuous consistency error. \\
$P_X^\star,\ \widehat P_{h,X}$ & Reconstructed target and model posteriors associated with the finite set $X$. \\
$\epsilon_{\rm cond},\ \epsilon_{\rm flow},\ \epsilon_{\rm ODE},\ \epsilon_{\rm rec}$ & Four end-to-end error terms.\\
$H^s,\ d_H$ & Sobolev space and Hausdorff distance used for the geometry result.\\
\hline
\end{longtable}
\endgroup

\subsection{Preliminaries}\label{app:preliminaries}

We briefly review the key concepts underlying our approach.

\subsubsection{Perceiver}\label{app:perceiver}
The Perceiver module~\cite{jaegle2021perceiver} introduces trainable latent queries
$\mathbf Z_q\in\mathbb R^{N_\ell\times d_z}$, which interact with the input-token array
$\mathbf H$ defined in Sec.~\ref{sec4} through cross-attention:
\begingroup
\begin{align*}
\widetilde{\mathbf Z}_{\rm agg}
&=\mathbf Z_q+
  \operatorname{MHA}\bigl(\operatorname{LN}(\mathbf Z_q),
                          \operatorname{LN}(\mathbf H),
                          \operatorname{LN}(\mathbf H)\bigr),
\\
\mathbf Z_{\rm agg}
&=\widetilde{\mathbf Z}_{\rm agg}+
  \operatorname{MLP}\bigl(\operatorname{LN}(\widetilde{\mathbf Z}_{\rm agg})\bigr).
\end{align*}
\endgroup
Here MHA, LN, and MLP denote multi-head attention, layer normalization, and a feedforward network, respectively. $N_\ell$ is the number of latent queries and $d_z$ is the latent token dimension.

\subsubsection{Transformer}\label{app:transformer}
The aggregated latent tokens $\mathbf Z_{\rm agg}$ are processed by
$L_{\rm enc}$ pre-norm Transformer blocks~\citep{vaswani2017attention,xiong2020layer}:
\begingroup
\begin{align*}
\mathbf U_0 &= \operatorname{LN}(\mathbf Z_{\rm agg}), \\
\mathbf U_\ell^{\prime}
&= \operatorname{MSA}\bigl(\operatorname{LN}(\mathbf U_{\ell-1})\bigr)
   + \mathbf U_{\ell-1}, \qquad \ell=1,\dots,L_{\rm enc}, \\
\mathbf U_\ell
&= \operatorname{MLP}\bigl(\operatorname{LN}(\mathbf U_\ell^{\prime})\bigr)
   + \mathbf U_\ell^{\prime}, \qquad \ell=1,\dots,L_{\rm enc}.
\end{align*}
\endgroup
MSA denotes multi-head self-attention. The final state $\mathbf U_{L_{\rm enc}}$ is the fixed-size encoder representation denoted by $\mathbf z$ in the main text.

\subsubsection{Rectified flow}
Rectified flow (RF)~\cite{liu2023flow} replaces the stochastic dynamics of diffusion models with a deterministic ordinary differential equation (ODE) that maps noise to data along straighter trajectories. This yields faster and more stable generation. In this work, we use RF for conditional generation, focusing on posterior estimation $p(x \mid c)$, where $x$ denotes the unknown data and $c$ the observed condition. To distinguish this generic background notation from the GeoFAE parameters, let $v_\vartheta(x_t,t)$ denote a generic velocity field. Its ODE is
\begin{equation*}
    \frac{d x_t}{d t} = v_\vartheta(x_t, t), \quad t \in [0,1],
\end{equation*}
with initial condition $x_0 \sim \pi_0$ (e.g., Gaussian noise). The flow is trained by minimizing the objective
\begin{equation*}
    \mathcal{L}_{\mathrm{RF}}(\vartheta)
    = \mathbb{E}_{t \sim \operatorname{U}(0,1), (x_0,x_1) \sim (\pi_0,\pi_1)}
    \left\| v_\vartheta(x_t, t) - (x_1 - x_0) \right\|^2,
\end{equation*}
where $x_t = (1-t)x_0 + t x_1$ interpolates between $x_0$ and data $x_1 \sim \pi_1$. This yields simpler transport paths compared to score-based diffusion models.

For conditional generation, the velocity is extended to $v_\vartheta(x_t,t,c)$ with objective
\begin{equation*}
    \mathcal{L}_{\mathrm{CRF}}(\vartheta)
    = \mathbb{E}_{t, x_0, x_1, c}
    \left\| v_\vartheta(x_t, t, c) - (x_1 - x_0) \right\|^2.
\end{equation*}
Given a trained velocity field, posterior sampling is performed by solving
\begin{equation*}
    \frac{d x_t}{d t} = v_\vartheta(x_t, t, c), \quad x_0 \sim \pi_0,
\end{equation*}
and the terminal state follows the modeled conditional distribution $x_1\sim p_\vartheta(x\mid c)$. This enables efficient posterior sampling without explicit likelihood evaluation or Markov Chain Monte Carlo (MCMC).

\subsection{Metrics}\label{app:metrics}
\subsubsection{Evaluation metrics}
We evaluate GeoFunFlow from two complementary perspectives aligned with its problem setting: reconstruction of continuous fields and recovery of geometry. For continuous outputs, including physical fields such as velocity components \(u,v\), pressure \(p\), and signed distance functions (SDFs) viewed as continuous scalar fields, we use the relative \(L^2\) error to quantify function-level prediction accuracy over the domain. For geometry recovery, where shape is represented implicitly through a distance function, we use metrics designed for geometric comparison: the Chamfer distance measures discrepancy between the predicted and reference boundaries, and the Intersection over Union (IoU) measures overlap between the corresponding induced domains. These metrics together provide a comprehensive assessment of both field accuracy and geometric fidelity.

\paragraph{Relative $L^2$ norm}
For continuous field reconstruction and prediction, we use the relative $L^2$ error as the evaluation metric:
\begin{equation*}
\mathrm{Rel}\text{.}~L^2
=
\frac{\|u_{\mathrm{pred}}-u_{\mathrm{true}}\|_{L^2(\Omega)}}{\|u_{\mathrm{true}}\|_{L^2(\Omega)}}.
\end{equation*}
Here, $u_{\mathrm{pred}}$ and $u_{\mathrm{true}}$ denote the predicted and reference fields, respectively, and $\Omega$ is the underlying domain. A smaller value indicates a more accurate prediction.

\paragraph{Chamfer distance (CD)}
To quantify the discrepancy between the predicted and reference boundaries, we use the symmetric Chamfer distance between their sampled point sets. Let
\[
\Gamma_{\rm pred}^{(N)}=\{\mathbf a_i\}_{i=1}^{N}\subset \mathbb{R}^d,
\qquad
\Gamma_{\rm true}^{(M)}=\{\mathbf b_j\}_{j=1}^{M}\subset \mathbb{R}^d
\]
denote the sampled predicted and reference boundaries, respectively. The Chamfer distance is defined as
\begin{equation*}
\mathrm{CD}(\Gamma_{\rm pred}^{(N)},\Gamma_{\rm true}^{(M)})
=
\frac{1}{N}\sum_{i=1}^{N}\min_{1\le j\le M}\|\mathbf a_i-\mathbf b_j\|_2^2
+
\frac{1}{M}\sum_{j=1}^{M}\min_{1\le i\le N}\|\mathbf b_j-\mathbf a_i\|_2^2.
\end{equation*}
This metric measures the bidirectional average nearest-neighbor discrepancy between the two boundary point sets. A smaller Chamfer distance indicates better geometric agreement between the predicted and reference boundaries.

\paragraph{Intersection over Union (IoU)}
To measure the overlap between the predicted and ground-truth domains, we use the Intersection over Union (IoU). Let
\[
\Omega_{\mathrm{pred}},\ \Omega_{\mathrm{true}} \subset \mathbb{R}^d
\]
denote the predicted and reference domains, respectively. The IoU is defined as
\begin{equation*}
\mathrm{IoU}(\Omega_{\mathrm{pred}},\Omega_{\mathrm{true}})
=
\frac{|\Omega_{\mathrm{pred}}\cap \Omega_{\mathrm{true}}|}
{|\Omega_{\mathrm{pred}}\cup \Omega_{\mathrm{true}}|},
\end{equation*}
where $|\cdot|$ denotes the Lebesgue measure, i.e., area in two dimensions and volume in three dimensions. The IoU takes values in $[0,1]$, with larger values indicating better agreement. In particular, $\mathrm{IoU}=1$ if and only if the two domains coincide up to a set of measure zero.

For SDF-defined geometries, the domains are induced by the sign of the level-set function. Specifically, given the predicted and reference signed distance functions $\varphi_{\mathrm{pred}}$ and $\varphi_{\mathrm{true}}$, we define
\begin{equation*}
\Omega_{\mathrm{pred}}=\{\mathbf x\in\Omega:\varphi_{\mathrm{pred}}(\mathbf x)<0\},
\qquad
\Omega_{\mathrm{true}}=\{\mathbf x\in\Omega:\varphi_{\mathrm{true}}(\mathbf x)<0\},
\end{equation*}
and compute the IoU accordingly.

\subsubsection{Uncertainty quantification metrics}
The uncertainty criteria follow the interpretation of each output. For physical fields, we evaluate spatial informativeness through the Pearson correlation between pointwise reconstruction error and posterior standard deviation. For inferred geometries, we additionally evaluate the calibration of the SDF intervals and the coverage of the induced boundary or inclusion region. PICP and RMSCE measure interval coverage, whereas MPIW records the accompanying loss of sharpness.

\paragraph{Pearson correlation}
To assess whether the predicted uncertainty is informative of the actual prediction error, we compute the Pearson correlation coefficient between the pointwise absolute error and the associated uncertainty estimate. Let
\[
e_n = |y_n-\mu_n|,
\qquad
s_n = \sigma_n,
\qquad n=1,\dots,N,
\]
where \(y_n\) is the reference value, \(\mu_n\) is the predictive mean, and \(\sigma_n\) is the predicted standard deviation at the \(n\)-th evaluation point. The Pearson correlation coefficient is defined as
\begin{equation*}
r
=
\frac{\sum_{n=1}^{N}(e_n-\bar e)(s_n-\bar s)}
{\sqrt{\sum_{n=1}^{N}(e_n-\bar e)^2}\sqrt{\sum_{n=1}^{N}(s_n-\bar s)^2}},
\end{equation*}
where
\[
\bar e=\frac{1}{N}\sum_{n=1}^{N}e_n,
\qquad
\bar s=\frac{1}{N}\sum_{n=1}^{N}s_n.
\]
A larger positive value of \(r\) indicates that regions with larger prediction errors tend to be assigned higher uncertainty, suggesting that the uncertainty estimate is more informative.

\paragraph{Prediction Interval Coverage Probability (PICP)}
To evaluate the reliability of predictive intervals at a given nominal coverage level \(1-\alpha\), we use the prediction interval coverage probability (PICP). Let
\[
I_n^{(\alpha)}=[\,l_n^{(\alpha)},\,u_n^{(\alpha)}\,]
\]
denote the predictive interval for the \(n\)-th prediction. The PICP is defined as
\begin{equation*}
\mathrm{PICP}(1-\alpha)
=
\frac{1}{N}\sum_{n=1}^{N}
\mathbf{1}\!\left\{
y_n \in I_n^{(\alpha)}
\right\}.
\end{equation*}
This metric measures the proportion of reference values contained within the predictive intervals. Ideally, \(\mathrm{PICP}(1-\alpha)\) should be close to the nominal coverage \(1-\alpha\).

\paragraph{Mean Prediction Interval Width (MPIW)}
To quantify the sharpness of predictive intervals, we use the mean prediction interval width (MPIW). For a nominal coverage level \(1-\alpha\), MPIW is defined as
\begin{equation*}
\mathrm{MPIW}(1-\alpha)
=
\frac{1}{N}\sum_{n=1}^{N}
\Bigl(
u_n^{(\alpha)}-l_n^{(\alpha)}
\Bigr).
\end{equation*}
MPIW measures the average width of the predictive intervals, with smaller values indicating sharper predictions. However, MPIW should be interpreted together with PICP, since overly wide intervals may achieve high coverage while providing limited practical informativeness.

\paragraph{Root Mean Squared Calibration Error (RMSCE)}
To assess calibration across multiple confidence levels, we compare the empirical coverage of predictive intervals against their nominal coverage levels. Let $\{1-\alpha_k\}_{k=1}^{K}$
be the set of nominal coverage levels under consideration. Here we select  $1-\alpha_k = 50\%, 80\%, 90\%, 95\%$. The root mean squared calibration error is defined as
\begin{equation*}
\mathrm{RMSCE}
=
\left(
\frac{1}{K}
\sum_{k=1}^{K}
\Bigl(
\mathrm{PICP}(1-\alpha_k) - (1-\alpha_k)
\Bigr)^2
\right)^{1/2}.
\end{equation*}
RMSCE summarizes the discrepancy between empirical and nominal coverage across confidence levels. Smaller values indicate better calibration, with \(\mathrm{RMSCE}=0\) corresponding to perfect agreement between empirical and nominal coverage.

\paragraph{Geometry cover rate}
Let $\mathcal T_g$ denote the reference geometric set on the evaluation grid and $\mathcal U_g$ the reported geometric uncertainty region. Under strict matching, the cover rate is
\[
\mathrm{Cover~Rate}
=\frac{|\mathcal T_g\cap\mathcal U_g|}{|\mathcal T_g|},
\]
where $|\cdot|$ counts evaluation pixels. For LDC and FPO, $\mathcal T_g$ consists of the reference boundary pixels. For OT, it consists of the pixels in the reference absorbing-inclusion region. In the evaluation for LDC and FPO, boundary locations separated by at most one pixel are treated as matched. A reference boundary pixel is therefore counted as covered if it lies in $\mathcal U_g$ or within one pixel of this region, while the denominator remains the total number of reference boundary pixels. This tolerance accommodates small contour displacements on the evaluation grid without changing the inferred geometry. 

\paragraph{Boundary band width}
For LDC and FPO, let $\mathcal U_g$ be the geometric uncertainty region
and $\mathcal T_g$ the reference boundary pixels, as defined above. We
report
\[
W_{\mathrm{band}}=\frac{|\mathcal U_g|}{|\mathcal T_g|},
\]
where $|\cdot|$ counts evaluation pixels. This ratio measures the spatial
extent of the uncertainty band relative to the boundary length, whereas
MPIW measures interval width in SDF values. 

\paragraph{Geometry error--uncertainty correlation}
Let $\mathcal U_g$ be the uncertainty band and $\mathcal E_g$ the set of
pixels where the predicted mean SDF and reference SDF disagree on the
inside--outside classification, relaxed by one pixel. We report
\[
\rho_{\mathrm{EU}}
=
\operatorname{Corr}_{x\in\Omega_g}
\bigl(
\mathbf 1[x\in\mathcal U_g],
\mathbf 1[x\in\mathcal E_g]
\bigr).
\]
A larger positive value indicates stronger spatial alignment between
the uncertainty band and geometric reconstruction errors. We interpret
this correlation alongside geometry cover rate.

\clearpage
\section{Theoretical results}
\label{app:theory}
\begingroup
We provide the proof of the error decomposition in Sec.~\ref{sec:theory-results}. We analyze the conditional posterior generated by GeoFunFlow following its inference pipeline. We first define the target and latent distributions (Sec.~\ref{app:theorm1}), analyze latent rectified flow and numerical integration (Sec.~\ref{app:theorm2}), and establish the posterior error bound (Sec.~\ref{app:theorm3}). We then extend the analysis to field and geometry posteriors (Sec.~\ref{app:theorm4}), finite discretizations, and posterior statistics (Sec.~\ref{app:theorm5}):
\[
\widehat{P}_h (\cdot\mid \mathbf c)
\;\xrightarrow{\;\epsilon_{\mathrm{ODE}}\;}\;
(T_\Omega)_{\#} Q_Z^\psi
\;\xrightarrow{\;\epsilon_{\mathrm{flow}}\;}\;
(T_\Omega)_{\#} P_Z^\star(\,\cdot\mid \mathbf z_\mathbf c)
\;\xrightarrow{\;\epsilon_{\mathrm{cond}}\;}\;
(T_\Omega)_{\#} P_Z^\star(\,\cdot\mid \mathbf c)
\;\xrightarrow{\;\epsilon_{\mathrm{rec}}\;}\;
P^\star(\,\cdot\mid \mathbf c).
\]
Here, $P^\star(\cdot\mid \mathbf c)$ is the target conditional distribution of the complete function given the partial observation $c$, while $\widehat{P}_h(\cdot\mid \mathbf c)$ is the posterior generated by GeoFunFlow using a numerical ODE solver with maximum step size $h$. The map $T_\Omega$ denotes the continuous decoder from the latent space to the function space, and $(T_\Omega)_{\#}$ denotes the corresponding pushforward of a latent distribution. $Q_Z^\psi(\cdot\mid \mathbf z_\mathbf c)$ is the terminal latent distribution produced by the learned continuous flow conditioned on $z_c=E_\theta(\mathbf c)$. The distributions $P_Z^\star(\cdot\mid \mathbf z_\mathbf c)$ and $P_Z^\star(\cdot\mid \mathbf c)$ are the target latent distributions conditioned on the encoded condition $\mathbf z_\mathbf c$ and the original condition $\mathbf c$, respectively.

The four transitions identify distinct sources of approximation error. Specifically, $\epsilon_{\mathrm{ODE}}$ measures numerical integration error, $\epsilon_{\mathrm{flow}}$ measures the error of the learned latent transport, $\epsilon_{\mathrm{cond}}$ measures information lost when $\mathbf c$ is compressed into $\mathbf z_\mathbf c$, and $\epsilon_{\mathrm{rec}}$ measures the error introduced by encoding and decoding the complete target function. Precise definitions are given in the following sections.

\subsection{Function-space formulation and conditional latent laws}
\label{app:theorm1}
Fix a supplied or ambient domain $\Omega\subset\mathbb R^d$. When supplied
domains vary between instances, the statements below apply conditionally for
each $\Omega$.  Let
$\mathcal H(\Omega)$ be a separable Hilbert space containing the complete
modeled functions.  For the standard field result, one may take
$\mathcal H(\Omega)=L^2(\Omega;\mathbb R^C)$.  When geometry is inferred, a
complete realization is
\[
\mathbf f=(\mathbf u,\varphi)
\in\mathcal H_{\mathbf u}(\Omega)\times\mathcal H_\varphi(\Omega),
\]
where $\mathbf u$ collects the physical channels and $\varphi$ is the SDF.
We equip this product with
\begin{equation}
\|\mathbf f\|_{\mathcal H(\Omega)}^2
=a_{\mathbf u}\|\mathbf u\|_{\mathcal H_{\mathbf u}(\Omega)}^2
+a_\varphi\|\varphi\|_{\mathcal H_\varphi(\Omega)}^2,
\qquad a_u,a_\varphi>0,
\label{eq:product_norm}
\end{equation}
with weights matching the channel normalization used in GeoFAE.  The notation
$W_2^\Omega$ refers to the 2-Wasserstein distance induced by this norm.

\revred{Let \(\mathbf F\) be the random complete modeled function, with a realization denoted by \(\mathbf f\), and let \(C\) be the random partial condition. For an observed realization \(c\), define the target conditional law by}
\[
\revred{
P^\star(\cdot\mid c)
=
\operatorname{Law}(\mathbf F\mid C=\mathbf c)
\in\mathcal P_2(\mathcal H(\Omega)).
}
\]
\revred{For geometry inference, this is the joint law of \((\mathbf u,\varphi)\), allowing dependence between the physical and geometric components.}

\revred{Let \(C_{\mathrm{full}}\) be the complete input paired jointly with \((\mathbf F,C)\) on the data-generating probability space:}
\[
\revred{
C_{\mathrm{full}}
=
\left(
X,\mathbf 1,
[\mathbf F(\mathbf x_i)]_{i=1}^{n}
\right).
}
\]
\revred{The shared GeoFAE encoder produces}
\[
\revred{
Z_c=E_\theta(C),
\qquad
Z_1=E_\theta(C_{\mathrm{full}}).
}
\]
Their realizations are the conditioning and target latents \(\mathbf z_{\mathbf c}\) and \(z_1\) in Sec.~4. Importantly, \(C_{\mathrm{full}}\) is retained as a random variable jointly paired with \(C\) and \(\mathbf F\). 
\[
P_Z^\star(\cdot\mid \mathbf c)
=
\operatorname{Law}(Z_1\mid C=\mathbf c),
\qquad
P_Z^\star(\cdot\mid \mathbf z_{\mathbf c})
=
\operatorname{Law}(Z_1\mid Z_{\mathbf c}=\mathbf z_{\mathbf c}).
\]
The distinction matters because the conditional flow receives
\(\mathbf z_{\mathbf c}=E_\theta(c)\), rather than the uncompressed condition itself.

The continuous decoder induces the map
\[
T_\Omega:\mathcal Z\to\mathcal H(\Omega),
\qquad
[T_\Omega(\mathbf z)](\mathbf q)=D_\phi(\mathbf z,\mathbf q).
\]
Let $Q_Z^\psi(\cdot\mid\mathbf z_{\mathbf c})$ be the terminal law of the learned ODE
\[
\frac{d\mathbf z(t)}{dt}
=\mathbf g_\psi(\mathbf z(t),t,\mathbf z_{\mathbf c}),
\qquad
\mathbf z(0)\sim\mathcal N(\mathbf0,\mathbf I),
\qquad t\in[0,1],
\]
and let $Q_{Z,h}^\psi(\cdot\mid\mathbf z_{\mathbf c})$ be the law returned by a numerical
solver with maximum step size $h$.  The generated function posterior is
\[
\widehat P_h(\cdot\mid\mathbf c)
=(T_\Omega)_\#Q_{Z,h}^\psi(\cdot\mid\mathbf z_{\mathbf c}).
\]
When the target contains an SDF, each latent draw is decoded jointly into the
physical and geometric channels.

\subsection{Conditional rectified flow}
\label{app:theorm2}
Fix $\mathbf z_{\mathbf c}$, draw
$\mathbf Z_0\sim\mathcal N(\mathbf0,\mathbf I)$ independently of
$\mathbf Z_1\sim P_Z^\star(\cdot\mid\mathbf z_{\mathbf c})$, and set
$\mathbf Z_t=(1-t)\mathbf Z_0+t\mathbf Z_1$.  The conditional population
counterpart of $\mathcal L_{\rm CRF}$ is
\begin{equation}
\mathcal J_{\mathbf z_{\mathbf c}}(\mathbf g)
=\int_0^1
\mathbb E\!\left[
\|\mathbf g(\mathbf Z_t,t,\mathbf z_{\mathbf c})-(\mathbf Z_1-\mathbf Z_0)\|_{\mathcal Z}^2
\,\middle|\,\mathbf z_{\mathbf c}
\right]dt.
\label{eq:population_crf}
\end{equation}

\begin{proposition}[Population-optimal conditional velocity]
\label{prop:optimal_velocity}
If $\mathbf Z_0$ and $\mathbf Z_1$ have finite second moments, a minimizer of
Eq.~(\plaineqref{eq:population_crf}) is
\[
\mathbf g^\star(\mathbf z,t,\mathbf z_{\mathbf c})
=\mathbb E[\mathbf Z_1-\mathbf Z_0
\mid\mathbf Z_t=\mathbf z,\mathbf z_{\mathbf c}].
\]
Moreover,
\[
\mathcal J_{\mathbf z_{\mathbf c}}(\mathbf g)
-\mathcal J_{\mathbf z_{\mathbf c}}(\mathbf g^\star)
=\int_0^1
\mathbb E_{\mathbf Z\sim P_t^\star(\cdot\mid\mathbf z_{\mathbf c})}
\|\mathbf g(\mathbf Z,t,\mathbf z_{\mathbf c})
-\mathbf g^\star(\mathbf Z,t,\mathbf z_{\mathbf c})\|_{\mathcal Z}^2dt,
\]
where $P_t^\star(\cdot\mid\mathbf z_{\mathbf c})$ is the law of $\mathbf Z_t$.
\end{proposition}

\begin{proof}
For fixed $t$ and $\mathbf z_{\mathbf c}$, conditional expectation is the orthogonal
projection of $\mathbf Z_1-\mathbf Z_0$ onto the square-integrable functions
of $\mathbf Z_t$.  The conditional bias--variance identity separates the
integrand in Eq.~(\plaineqref{eq:population_crf}) into the squared distance from
$\mathbf g^\star$ and a term independent of $\mathbf g$.  Integration over
$t$ gives both statements.
\end{proof}

Under the standard regularity conditions for the continuity
equation, the ODE driven by $\mathbf g^\star$ has time-$t$ law
$P_t^\star(\cdot\mid\mathbf z_{\mathbf c})$ and therefore transports the Gaussian base
law to $P_Z^\star(\cdot\mid\mathbf z_{\mathbf c})$.  We separate continuous transport
and numerical integration by defining
\begin{align*}
\epsilon_{\rm flow}(\mathbf z_{\mathbf c})
&=W_2^{\mathcal Z}\!\left(
Q_Z^\psi(\cdot\mid\mathbf z_{\mathbf c}),
P_Z^\star(\cdot\mid\mathbf z_{\mathbf c})
\right),\\
\epsilon_{\rm ODE}(\mathbf z_{\mathbf c};h)
&=W_2^{\mathcal Z}\!\left(
Q_{Z,h}^\psi(\cdot\mid\mathbf z_{\mathbf c}),
Q_Z^\psi(\cdot\mid\mathbf z_{\mathbf c})
\right).
\end{align*}

\begin{proposition}[A sufficient flow-loss bound]
\label{prop:flow_loss_bound}
Fix $\mathbf z_{\mathbf c}$.  Suppose that the exact and learned ODEs have
unique flows, the ODE driven by $\mathbf g^\star$ has time-$t$ law
$P_t^\star(\cdot\mid\mathbf z_{\mathbf c})$, and
$\mathbf g_\psi(\cdot,t,\mathbf z_{\mathbf c})$ is $L_\psi$-Lipschitz in its transported
state uniformly over $t\in[0,1]$.  With
$\Delta_{\rm CRF}(\mathbf z_{\mathbf c})
=\mathcal J_{\mathbf z_{\mathbf c}}(\mathbf g_\psi)
-\mathcal J_{\mathbf z_{\mathbf c}}(\mathbf g^\star)$,
\begin{equation}
\epsilon_{\rm flow}(\mathbf z_{\mathbf c})
\le K(L_\psi)\sqrt{\Delta_{\rm CRF}(\mathbf z_{\mathbf c})},
\qquad
K(L_\psi)=
\begin{cases}
\left[(e^{2L_\psi}-1)/(2L_\psi)\right]^{1/2},&L_\psi>0,\\
1,&L_\psi=0.
\end{cases}
\label{eq:flow_loss_bound}
\end{equation}
\end{proposition}

\begin{proof}
Start the exact and learned ODEs from the same Gaussian draw and
denote their solutions by $\mathbf Z_t^\star$ and
$\widehat{\mathbf Z}_t$.  Adding and subtracting
$\mathbf g_\psi(\mathbf Z_t^\star,t,\mathbf z_{\mathbf c})$ gives
\begin{align*}
&\|\mathbf g_\psi(\widehat{\mathbf Z}_t,t,\mathbf z_{\mathbf c})
-\mathbf g^\star(\mathbf Z_t^\star,t,\mathbf z_{\mathbf c})\|_{\mathcal Z}\\
&\quad\le L_\psi
\|\widehat{\mathbf Z}_t-\mathbf Z_t^\star\|_{\mathcal Z}
+\|\mathbf g_\psi(\mathbf Z_t^\star,t,\mathbf z_{\mathbf c})
-\mathbf g^\star(\mathbf Z_t^\star,t,\mathbf z_{\mathbf c})\|_{\mathcal Z}.
\end{align*}
Gronwall's inequality therefore gives
\[
\|\widehat{\mathbf Z}_1-\mathbf Z_1^\star\|_{\mathcal Z}
\le\int_0^1 e^{L_\psi(1-t)}
\|\mathbf g_\psi(\mathbf Z_t^\star,t,\mathbf z_{\mathbf c})
-\mathbf g^\star(\mathbf Z_t^\star,t,\mathbf z_{\mathbf c})\|_{\mathcal Z}\,dt.
\]
After squaring and applying Cauchy--Schwarz,
\begin{align*}
\mathbb E\|\widehat{\mathbf Z}_1-\mathbf Z_1^\star\|_{\mathcal Z}^2
&\le K(L_\psi)^2
\int_0^1
\mathbb E_{\mathbf Z\sim P_t^\star(\cdot\mid\mathbf z_{\mathbf c})}
\|\mathbf g_\psi(\mathbf Z,t,\mathbf z_{\mathbf c})
-\mathbf g^\star(\mathbf Z,t,\mathbf z_{\mathbf c})\|_{\mathcal Z}^2dt\\
&=K(L_\psi)^2\Delta_{\rm CRF}(\mathbf z_{\mathbf c}),
\end{align*}
where the equality follows from
Proposition~\ref{prop:optimal_velocity}.  The shared initial draw couples the
terminal laws and proves Eq.~(\plaineqref{eq:flow_loss_bound}).
\end{proof}

This result concerns the population \emph{excess} loss.  The raw population
loss also contains the irreducible term
$\mathcal J_{\mathbf z_{\mathbf c}}(\mathbf g^\star)$, and its empirical estimate
additionally reflects approximation, generalization, and optimization error.
Consequently, the reported training loss is not itself a Wasserstein error.
For a stable order-$r$ solver and a sufficiently smooth learned velocity, the
pathwise global truncation estimate gives
\[
\epsilon_{\rm ODE}(\mathbf z_{\mathbf c};h)
\le C_{\rm ODE}(\mathbf z_{\mathbf c})h^r,
\]
provided the corresponding error constant is square-integrable under the
initial Gaussian law.

\subsection{Posterior approximation}
\label{app:theorm3}
Condition compression and complete-state reconstruction contribute
\begin{align*}
\epsilon_{\rm cond}(\mathbf c)
&=W_2^{\mathcal Z}\!\left(
P_Z^\star(\cdot\mid\mathbf z_{\mathbf c}),
P_Z^\star(\cdot\mid\mathbf c)
\right),\\
\epsilon_{\rm rec}(\mathbf c)
&=\left(
\mathbb E_{\mathbf f\sim P^\star(\cdot\mid\mathbf c)}
\|T_\Omega(E_\theta(\mathbf c_{\rm full}(\mathbf f;\mathbf c)))
-\mathbf f\|_{\mathcal H(\Omega)}^2
\right)^{1/2}.
\end{align*}
The first term measures the information about the complete-state latent that
is lost when $\mathbf c$ is represented only by $\mathbf z_{\mathbf c}$.  The second is
the root-mean-square error of encoding and decoding the complete target.  For
$\mathbf f=(\mathbf u,\varphi)$, it includes both physical and SDF errors in
the product norm of Eq.~(\plaineqref{eq:product_norm}).

\begin{assumption}[Decoder regularity]
\label{ass:decoder}
The decode-to-function map is $L_D$-Lipschitz:
\[
\|T_\Omega(\mathbf z)-T_\Omega(\mathbf z')\|_{\mathcal H(\Omega)}
\le L_D\|\mathbf z-\mathbf z'\|_{\mathcal Z},
\qquad \mathbf z,\mathbf z'\in\mathcal Z.
\]
\end{assumption}

\begin{theorem}[End-to-end conditional posterior approximation]
\label{thm:joint_posterior}
Suppose that $T_\Omega$ is $L_D$-Lipschitz and that the conditional laws above
exist and have finite second moments. Then, for $P_{\mathbf C}$-almost every
$\mathbf c$,
\begin{equation}
W_2^\Omega\!\left(
\widehat P_h(\cdot\mid\mathbf c),
P^\star(\cdot\mid\mathbf c)
\right)
\le
L_D\!\left[
\epsilon_{\rm cond}(\mathbf c)
+\epsilon_{\rm flow}(\mathbf z_{\mathbf c})
+\epsilon_{\rm ODE}(\mathbf z_{\mathbf c};h)
\right]
+\epsilon_{\rm rec}(\mathbf c).
\label{eq:main_bound}
\end{equation}
\end{theorem}

\begin{proof}[Proof of Theorem~\ref{thm:joint_posterior}]
Insert the continuous learned latent law and the two target latent laws, and
apply the triangle inequality:
\begin{align*}
&W_2^\Omega(\widehat P_h,P^\star)\\
&\le
W_2^\Omega\!\left(
(T_\Omega)_\#Q_{Z,h}^\psi,
(T_\Omega)_\#Q_Z^\psi
\right)
+
W_2^\Omega\!\left(
(T_\Omega)_\#Q_Z^\psi,
(T_\Omega)_\#P_Z^\star(\cdot\mid\mathbf z_{\mathbf c})
\right)\\
&\quad+
W_2^\Omega\!\left(
(T_\Omega)_\#P_Z^\star(\cdot\mid\mathbf z_{\mathbf c}),
(T_\Omega)_\#P_Z^\star(\cdot\mid\mathbf c)
\right)\\
&\quad+
W_2^\Omega\!\left(
(T_\Omega)_\#P_Z^\star(\cdot\mid\mathbf c),
\revred{P^\star(\cdot\mid\mathbf c)}
\right),
\end{align*}
where conditioning arguments shared within a term are suppressed. A
Lipschitz pushforward contracts Wasserstein distance up to its Lipschitz
constant, so the first three terms are bounded by
$L_D\epsilon_{\rm ODE}$, $L_D\epsilon_{\rm flow}$, and
$L_D\epsilon_{\rm cond}$.

\revred{For the final term, use the conditional joint law of
$(\mathbf F, C_{\rm full})$ given $C=\mathbf c$. Under this
conditional law, pair $\mathbf F$ with
\[
T_\Omega(\mathbf Z_1)
=
T_\Omega\!\left(E_\theta(C_{\rm full})\right).
\]
The first component has law $P^\star(\cdot\mid\mathbf c)$, whereas the second
has law
\[
(T_\Omega)_\#P_Z^\star(\cdot\mid\mathbf c).
\]
Therefore, this conditional coupling gives
\[
W_2^\Omega\!\left(
(T_\Omega)_\#P_Z^\star(\cdot\mid\mathbf c),
P^\star(\cdot\mid\mathbf c)
\right)
\le
\left[
\mathbb E\!\left[
\left\|
T_\Omega(\mathbf Z_1)-\mathbf F
\right\|_{\mathcal H(\Omega)}^2
\,\middle|\,
C=\mathbf c
\right]
\right]^{1/2}
=
\epsilon_{\rm rec}(\mathbf c).
\]
Summing the four bounds proves
Eq.~(\plaineqref{eq:main_bound}).}
\end{proof}

Theorem~\ref{thm:joint_posterior} separates errors with different practical
origins.  In particular, increasing the number of ODE steps can reduce only $\epsilon_{\rm ODE}$ and cannot address errors arising from information loss in the condition encoder, misspecification of the continuous flow, or incomplete GeoFAE reconstruction.

For compactness, define the right-hand side of Eq.~(\plaineqref{eq:main_bound}) as
\[
\mathcal E_{\rm total}(\mathbf c;h)
=L_D\!\left[
\epsilon_{\rm cond}(\mathbf c)
+\epsilon_{\rm flow}(\mathbf z_{\mathbf c})
+\epsilon_{\rm ODE}(\mathbf z_{\mathbf c};h)
\right]
+\epsilon_{\rm rec}(\mathbf c).
\]

\begin{corollary}[Physical-field and SDF marginals]
\label{cor:marginals}
For $\mathbf f=(\mathbf u,\varphi)$,
\begin{align*}
W_2^{\mathcal H_u}\!\left(
(\pi_u)_\#\widehat P_h,(\pi_u)_\#P^\star
\right)
&\le a_u^{-1/2}\mathcal E_{\rm total}(\mathbf c;h),\\
W_2^{\mathcal H_\varphi}\!\left(
(\pi_\varphi)_\#\widehat P_h,(\pi_\varphi)_\#P^\star
\right)
&\le a_\varphi^{-1/2}\mathcal E_{\rm total}(\mathbf c;h),
\end{align*}
where conditioning on $\mathbf c$ is suppressed.
\end{corollary}

\begin{proof}
Under Eq.~(\plaineqref{eq:product_norm}), the coordinate projections have Lipschitz
constants $a_u^{-1/2}$ and $a_\varphi^{-1/2}$.  Applying the Lipschitz
pushforward inequality to Theorem~\ref{thm:joint_posterior} proves the result.
\end{proof}

Control of the joint law is stronger than separate control of the two
marginals: distributions can share field and SDF marginals while coupling
different fields to different geometries.  Theorem~\ref{thm:joint_posterior}
retains this statistical dependence.  When geometry is supplied, the target
space contains only the unknown physical channels and the same theorem applies
on $\mathcal H_u(\Omega)$.  If a fixed geometry channel is retained, exact
conditioning on that channel gives a degenerate geometry marginal and leaves
the physical-field bound unchanged.

\subsection{From SDF posteriors to geometry posteriors}
\label{app:theorm4}
For an SDF $\varphi$, define its boundary through the zero-level-set map
\[
\mathfrak G(\varphi)
=
\Gamma_\varphi
=
\left\{
\mathbf q\in\overline\Omega:
\varphi(\mathbf q)=0
\right\}.
\]
Because an $L^2$ error alone does not control this set, establishing a boundary result requires uniform control and additional regularity near the zero level set.

\begin{proposition}[SDF-to-boundary stability under exact-SDF assumptions]
\label{prop:sdf_boundary}
\revred{Assume that $\overline\Omega$ is compact. Let $\varphi$ and
$\widehat\varphi$ be exact signed distance functions on $\overline\Omega$,
with the same sign convention and nonempty compact zero sets
$\Gamma_\varphi$ and $\Gamma_{\widehat\varphi}$. Then}
\begin{equation}
d_H
\left(
\Gamma_{\widehat\varphi},
\Gamma_\varphi
\right)
\le
\left\|
\widehat\varphi-\varphi
\right\|_{L^\infty(\Omega)}.
\label{eq:sdf_boundary}
\end{equation}
\end{proposition}

\begin{proof}
Let
\[
\delta
=
\left\|
\widehat\varphi-\varphi
\right\|_{L^\infty(\Omega)}.
\]
For any $\mathbf q\in\Gamma_\varphi$, exact signed-distance structure gives
\[
\operatorname{dist}
\left(
\mathbf q,\Gamma_{\widehat\varphi}
\right)
=
\left|
\widehat\varphi(\mathbf q)
\right|
=
\left|
\widehat\varphi(\mathbf q)-\varphi(\mathbf q)
\right|
\le
\delta.
\]
Therefore,
\[
\sup_{\mathbf q\in\Gamma_\varphi}
\operatorname{dist}
\left(
\mathbf q,\Gamma_{\widehat\varphi}
\right)
\le
\delta.
\]
Interchanging $\varphi$ and $\widehat\varphi$ gives
\[
\sup_{\mathbf q\in\Gamma_{\widehat\varphi}}
\operatorname{dist}
\left(
\mathbf q,\Gamma_\varphi
\right)
\le
\delta.
\]
Taking the maximum of the two directed distances proves
Eq.~(\plaineqref{eq:sdf_boundary}).
\end{proof}

\revred{Proposition~\ref{prop:sdf_boundary} applies to exact signed distance
functions. It does not imply that the zero-level-set map is Lipschitz on an
arbitrary class of continuous level-set functions, and it cannot be obtained
from an $L^2$ error bound alone.}

Let
\[
P_\varphi^\star
=
(\pi_\varphi)_\#P^\star,
\qquad
\widehat P_{\varphi,h}
=
(\pi_\varphi)_\#\widehat P_h,
\]
and define the induced geometry distributions by
\[
P_\Gamma^\star
=
\mathfrak G_\#P_\varphi^\star,
\qquad
\widehat P_{\Gamma,h}
=
\mathfrak G_\#\widehat P_{\varphi,h}.
\]
The boundaries are regarded as nonempty compact subsets of
$\overline\Omega$, and $W_2^{d_H}$ denotes the 2-Wasserstein distance induced
by the Hausdorff metric.

\begin{corollary}[Geometry-posterior error under exact-SDF support]
\label{cor:geometry}
\revred{Assume that both the true and generated SDF laws are supported on
exact signed distance functions on $\overline\Omega$, with the same sign
convention and nonempty compact zero sets. Assume further that}
\[
\mathcal H_\varphi(\Omega)
\hookrightarrow
C(\overline\Omega),
\qquad
\|v\|_{L^\infty(\Omega)}
\le
C_{\rm emb}
\|v\|_{\mathcal H_\varphi(\Omega)}.
\]
Then
\[
W_2^{d_H}\!\left(
\widehat P_{\Gamma,h},
P_\Gamma^\star
\right)
\le
C_{\rm emb}
a_\varphi^{-1/2}
\mathcal E_{\rm total}(\mathbf c;h).
\]
\revred{One possible choice is
$\mathcal H_\varphi(\Omega)=H^s(\Omega)$ with $s>d/2$, provided that the
target and generated SDFs belong to this space and that decoder Lipschitzness
and reconstruction error are controlled in the corresponding $H^s$ norm.
Membership of an exact SDF in such a space is an additional regularity
assumption and is not automatic.}
\end{corollary}

\begin{proof}
On the exact-SDF class, Proposition~\ref{prop:sdf_boundary} and the embedding
inequality give
\[
d_H
\left(
\mathfrak G(\widehat\varphi),
\mathfrak G(\varphi)
\right)
\le
C_{\rm emb}
\left\|
\widehat\varphi-\varphi
\right\|_{\mathcal H_\varphi(\Omega)}.
\]
The Lipschitz pushforward property and
Corollary~\ref{cor:marginals} therefore yield
\[
W_2^{d_H}\!\left(
\widehat P_{\Gamma,h},
P_\Gamma^\star
\right)
\le
C_{\rm emb}
W_2^{\mathcal H_\varphi}\!\left(
\widehat P_{\varphi,h},
P_\varphi^\star
\right)
\le
C_{\rm emb}
a_\varphi^{-1/2}
\mathcal E_{\rm total}(\mathbf c;h).
\]
\end{proof}

The target geometry data are represented by exact signed distance functions. 
Accordingly, Corollary~\ref{cor:geometry} characterizes the ideal exact-SDF regime. For finite learned models, exact signed-distance structure is encouraged through SDF supervision but is not enforced by the decoder.

\subsection{Discrete representations and posterior statistics}
\label{app:theorm5}
The preceding results concern functions, whereas training and
evaluation data contain their values on a finite reconstruction set
$X=\{\mathbf x_i\}_{i=1}^n$.  To make point evaluation well defined, let
$\mathcal H_{\rm reg}(\Omega)$ be a separable Banach function space continuously
embedded in both $\mathcal H(\Omega)$ and
$C(\overline\Omega;\mathbb R^C)$, and assume that
$P^\star(\cdot\mid\mathbf c)$ and
$\widehat P_h(\cdot\mid\mathbf c)$ are supported on
$\mathcal H_{\rm reg}(\Omega)$.  Define the sampling operator
\[
\mathcal S_X:\mathcal H_{\rm reg}(\Omega)\to\mathbb R^{n\times C},
\qquad
\mathcal S_X\mathbf f=[\mathbf f(\mathbf x_i)]_{i=1}^n,
\]
and let
$\mathcal R_X:\mathbb R^{n\times C}\to\mathcal H(\Omega)$ be a measurable
reconstruction operator.  Set
$\mathcal A_X=\mathcal R_X\circ\mathcal S_X:
\mathcal H_{\rm reg}(\Omega)\to\mathcal H(\Omega)$ and assume that
$\mathcal A_X\mathbf f$ has a finite second moment under both conditional
laws.  Define
\begin{equation}
\epsilon_{\rm disc}(\mathbf c;X)
=\left(
\mathbb E_{\mathbf f\sim P^\star(\cdot\mid\mathbf c)}
\|\mathcal R_X\mathcal S_X\mathbf f-\mathbf f\|_{\mathcal H(\Omega)}^2
\right)^{1/2}.
\end{equation}

The reconstructed target and model posteriors are the pushforward
measures
\[
P_X^\star(\cdot\mid\mathbf c)
=(\mathcal A_X)_\#P^\star(\cdot\mid\mathbf c),
\qquad
\widehat P_{h,X}(\cdot\mid\mathbf c)
=(\mathcal A_X)_\#\widehat P_h(\cdot\mid\mathbf c),
\]
which belong to $\mathcal P_2(\mathcal H(\Omega))$ by the assumptions above.

\begin{proposition}[Discrete-to-continuous posterior estimate]
Under the assumptions above,
\begin{equation}
W_2^\Omega\!\left(
\widehat P_{h,X}(\cdot\mid\mathbf c),P^\star(\cdot\mid\mathbf c)
\right)
\le
W_2^\Omega\!\left(
\widehat P_{h,X}(\cdot\mid\mathbf c),P_X^\star(\cdot\mid\mathbf c)
\right)
+\epsilon_{\rm disc}(\mathbf c;X).
\end{equation}
\end{proposition}

\begin{proof}
Apply the triangle inequality through
$P_X^\star(\cdot\mid\mathbf c)$.  If
$\mathbf f\sim P^\star(\cdot\mid\mathbf c)$, then the pair
$(\mathcal R_X\mathcal S_X\mathbf f,\mathbf f)$ couples
$P_X^\star(\cdot\mid\mathbf c)$ and $P^\star(\cdot\mid\mathbf c)$ with
root-mean-square cost $\epsilon_{\rm disc}(\mathbf c;X)$.
\end{proof}

This term measures the error introduced by representing a complete
function on finitely many nodes and reconstructing it in function space.  It
is distinct from $\epsilon_{\rm cond}$, which measures information lost when
the realized partial condition $\mathbf c=(X_\omega,\mathbf M_\omega,
\mathbf Y_\omega)$ is compressed into $\mathbf z_{\mathbf c}$.  A convergence rate for
$\epsilon_{\rm disc}$ requires additional approximation assumptions on
$X$ and $\mathcal R_X$.

\begin{corollary}[Posterior moments of observables]
\label{cor:uq}
Let $G:\mathcal H(\Omega)\to\mathbb R$ be $L_G$-Lipschitz.  Then
\begin{equation}
\left|\mathbb E_{\widehat P_h}G-\mathbb E_{P^\star}G\right|
\le L_G\mathcal E_{\rm total}(\mathbf c;h).
\label{eq:mean_bound}
\end{equation}
If $|G|\le B_G$, then
\begin{equation}
\left|\operatorname{Var}_{\widehat P_h}(G)
-\operatorname{Var}_{P^\star}(G)\right|
\le4B_GL_G\mathcal E_{\rm total}(\mathbf c;h).
\label{eq:variance_bound}
\end{equation}
\end{corollary}

\begin{proof}
The Kantorovich--Rubinstein inequality and $W_1\le W_2$ give
Eq.~(\plaineqref{eq:mean_bound}).  Since $G^2$ is $2B_GL_G$-Lipschitz, its
expectation differs by at most
$2B_GL_G\mathcal E_{\rm total}$.  The squared means satisfy the same bound,
which proves Eq.~(\plaineqref{eq:variance_bound}).
\end{proof}

Finally, the calibration scale $\lambda$ in
$I_{\lambda}^{(\alpha)}$ is applied after posterior sampling.  It changes the reported SDF interval and the induced boundary uncertainty region, but not the generated samples, $\widehat P_h$, their empirical mean, or the Wasserstein error in Theorem~\ref{thm:joint_posterior}.  The theorem and calibration therefore address complementary questions: posterior approximation concerns the generated joint distribution, whereas calibration adjusts the empirical coverage of a derived uncertainty region.
\endgroup

\clearpage
\section{Experimental details}
\label{app:experiments}

\subsection{Benchmarks}\label{app:benchmarks}

We evaluate GeoFunFlow on seven benchmarks encompassing the main sources of complexity in geometry-aware inverse modeling, including sparse and noisy measurements, fixed, variable, and unknown geometries, structured and unstructured discretizations, and both steady and unsteady physical systems. The benchmarks span porous media flow, internal and external fluid mechanics, and optical tomography, with substantial variation in spatial resolution and temporal sampling. For Cylinder and Airfoil, each snapshot is treated as an independent sample, with no temporal history or rollout prediction. In FPO, multiple snapshots from each trajectory provide temporal information for inferring the geometry, which remains fixed throughout the trajectory. Following the organization of the main text, we divide the benchmarks into field reconstruction on known domains and geometry inference from incomplete or indirect measurements.

\begin{table}[htbp]
\centering
\caption{\textbf{Benchmark characteristics and reconstruction targets.} Columns identify the simulator or data source, spatial discretization, mesh size, temporal sampling, train/test split, and reconstructed physical and geometry variables; ``traj.'' denotes trajectories.}
\label{tab:benchmark_summary}
\setlength{\tabcolsep}{3.8pt}
\begin{tabular}{l l l l l l l l}
\toprule
\textbf{Dataset}  & \textbf{Solver} & \textbf{Mesh type} & \textbf{Meshing} & \textbf{Mesh size} &\textbf{\# steps} & \textbf{Splits (train/test)} & \textbf{Variables} \\
\vspace{-0.8em}\\
\hline
\vspace{-0.8em}\\
Darcy  & Matlab PDE & Triangle 2D & irregular & $\sim$2,000 & Steady & 1800 / 200                    & $p$ \\
Cylinder    & COMSOL             & Triangle 2D     & irregular & $\sim$2,000 & 600  & 1000 / 100 (traj.)     & $u,v,p$ \\
Airfoil             & SU2                & Triangle 2D     & irregular & $\sim$5,000 & 600  & 1000 / 100 (traj.)     & $u,v,p$ \\
Ahmed body              & OpenFOAM           & Triangle 3D     & irregular & $\sim$50,000 & Steady     & 500 / 51                      & $p$ \\
LDC  & SBM           & Uniform 2D      & regular & 16,384 & Steady     & 900 / 100                      & $u,v,\varphi$ \\
FPO  & SBM          & Uniform 2D      & regular & 65,536 & 60     & 400 / 53                      & $u,v,\varphi$ \\
OT  & Matlab PDE           & Uniform 2D     & regular & 1,800 & Steady     & 900 / 100                      & $u,\varphi$ \\
\bottomrule
\end{tabular}
\end{table}

\paragraph*{Field reconstruction on known geometries}
The field-reconstruction benchmarks are Darcy, Cylinder, Airfoil, and Ahmed body. Darcy follows the setting of \citet{lu2021learning}, with $K(x,y)=0.1$ and $f=-1$, and uses Gaussian-process boundary conditions on triangular domains with a notch. It provides a controlled elliptic inverse problem on irregular meshes, with 2,000 total samples and a train/test split of 1,800/200. Cylinder and Airfoil, both adopted from \citet{pfaff2020learning}, provide snapshots from unsteady external-flow simulations over varying two-dimensional geometries. The Cylinder dataset contains simulations of incompressible flow around a cylinder, whereas the Airfoil dataset describes compressible aerodynamic flow around airfoil cross-sections. Each dataset comprises 1,000 training trajectories and 100 test trajectories, with 600 time steps per trajectory. Although the source data are organized as trajectories, each snapshot is treated as an independent reconstruction sample. The models receive no temporal history and are neither trained nor evaluated for rollout prediction. These benchmarks, therefore, assess reconstruction across varying geometries and instantaneous flow states rather than the learning of temporal dynamics. Ahmed body, generated by \citet{li2023geometry}, is the most demanding benchmark in this group. It involves turbulent three-dimensional vehicle aerodynamics with much larger CFD discretizations, more than 500 geometries, and a 500/51 train/test split. 

\paragraph*{Geometry inference from partial field information}
The geometry-inference benchmarks are lid-driven cavity (LDC), flow past objects (FPO), and optical tomography (OT). Together, these three datasets assess whether GeoFunFlow can recover latent geometry across distinct physical mechanisms while also producing uncertainty estimates that meaningfully reflect geometric ambiguity.

LDC and FPO are adopted from \citet{tali2025flowbench}. In LDC, the task is to infer internal obstacle geometry from partial observations of the velocity field in a two-dimensional cavity flow with coupled thermo-fluid effects. The top wall moves at a constant velocity while the other walls remain fixed over the computational domain $[0,2]\times[0,2]$. Different temperature boundary conditions are imposed on the bottom wall and obstacle surfaces, creating an interplay between shear-driven forced convection and buoyancy-driven natural convection. The flow regime is controlled by $Re\in[10^{1},10^{3}]$ and $Gr\in[10^{1},10^{7}]$, spanning forced, mixed, and free convection. We use the Harmonics geometry family, with 900 training samples and 100 test samples.

FPO probes geometry inference from unsteady external flow past a stationary complex object. The original computational domain is $[0,16]\times[0,4]$, and we focus on the spatial region $[0,8]\times[0,4]$. Unlike LDC, the obstacle geometry is fixed within a trajectory but the observations are time dependent: vortex shedding and wake interactions generate instantaneous flow fields that provide complementary indirect views of the same latent obstacle. We subsample 60 snapshots from the original 240 time frames, use 400 training trajectories and 53 test trajectories, and consider $Re\in[10^{2},10^{3}]$. Unless otherwise noted, FPO inference uses a temporal window size of 10, which balances the stability gained from aggregating multiple snapshots with the computational cost of longer temporal windows.

OT differs qualitatively from both fluid benchmarks because it seeks to recover absorbing inclusions from indirect boundary optical responses. The computational domain has physical dimensions $6\,\mathrm{cm}\times3\,\mathrm{cm}$ and is discretized with uniform grid spacing $0.1\,\mathrm{cm}$, resulting in a $60\times30$ grid. The underlying physics is governed by the diffusion equation
\begin{equation*}
- D \Delta u + \sigma(\mathbf{r}) u = 0 \quad \text{in } \Omega,
\end{equation*}
subject to boundary illumination conditions, where $u$ is the diffuse light field, $\sigma(\mathbf{r})$ is the spatially varying absorption coefficient, and $D$ is the diffusion constant. Absorption contrasts perturb photon propagation throughout the domain, so boundary measurements of $u$ contain only indirect information about the interior. In this benchmark, the contrasts form one or more absorbing inclusions, whose union is denoted by $\Omega_{\mathrm{abs}}$. Recovering their locations, topology, and boundaries can therefore be formulated as inference of a latent geometry. GeoFunFlow estimates $\Omega_{\mathrm{abs}}$ from the boundary optical responses and internally represents its geometry using the same SDF encoding as in the other geometry-inference benchmarks. We use 900 training samples and 100 test samples. This benchmark is strongly ill-posed because the target absorption geometry must be inferred from indirect, nonlocal boundary measurements~\citep{wang2025learning,carminati2021principles}.

\subsection{GeoFunFlow}

\paragraph{GeoFAE}

For the four reconstruction tasks with known geometry, the Perceiver uses an embedding dimension of $256$ and Fourier coordinate embeddings initialized from $\mathcal{N}(0,10)$. The number of trainable latent queries is fixed at $256$. The encoder consists of $8$ transformer blocks, each with $8$ attention heads and a two-layer MLP. The decoder consists of $4$ transformer blocks, each with $8$ attention heads and a two-layer MLP. It uses the same coordinate embedding scheme as the Perceiver. Finally, an MLP with one hidden layer projects each final query embedding to the target field dimension.

For the three tasks with unknown geometry, GeoFAE uses a grid-based encoder with an embedding dimension of $256$. The input grid resolutions for LDC, FPO and OT are $128\times128$, $128\times256$ and $60\times60$, respectively. The corresponding patch sizes are $16\times16$, $16\times32$ and $3\times3$. Patch embeddings are combined with fixed two-dimensional sinusoidal positional encodings. Each input field is encoded using $256$ trainable latent queries. The encoder contains $2$ Perceiver cross-attention blocks followed by $8$ latent self-attention blocks. Each block uses $8$ attention heads and a two-layer MLP.

For these unknown-geometry tasks, the decoder uses Fourier coordinate embeddings with a dimension of $256$. The Fourier frequency matrix is initialized from $\mathcal{N}(0,5^2)$ and optimized during training. The decoder consists of $4$ cross-attention blocks, each with $8$ attention heads and a two-layer MLP. Finally, an MLP with two hidden layers of width $256$ and GELU activations projects each query embedding to the target output channels.

\paragraph*{GeoFunFlow}

For the diffusion part of GeoFunFlow,  the embedding dimension of DiTs is $256$, matching the latent dimension of GeoFAE. The DiTs consist of $8$ transformer blocks with $8$ attention heads and  a  MLP ratio of 2. 

\subsection{Interpolation-based baselines}
For known-geometry reconstruction on complex domains, we consider two groups of baselines for irregular point-cloud data: interpolation-based methods and geometry-aware models. We include two non-parametric interpolation baselines and several grid-based neural baselines for irregular point-cloud data.
For the grid-based neural baselines (ViT, UNet, FNO, DPS), we use a common interpolation pipeline.
We first interpolate each input point cloud to a regular $128\times128$ grid using an exponential distance weighting with parameter $\beta = 10,000$, which assigns each grid cell a weighted average of nearby point values while limiting oversmoothing.
We then apply the baseline model on the gridded field and finally interpolate the output back to the query coordinates.

Interpolation error is empirically negligible on benchmarks with smooth fields and near-uniform sampling (Darcy, Cylinder). In contrast, Airfoil exhibits highly uneven sampling and Ahmed-body points lie on 3D surfaces, where interpolation onto a regular grid incurs non-negligible error. We therefore omit only the neural-network baselines that rely on this regular-grid interpolation pipeline. Interpolant and Kriging operate directly on the original coordinates and remain included.

\paragraph*{Interpolant~\cite{zhang2026operator}}
The Interpolant baseline is a non-parametric method with no trainable network. For each test sample, it fits a thin-plate-spline radial basis function interpolator from the observed sensor coordinates and field values, and evaluates the interpolant at the target query coordinates. We interpolate all output channels jointly, using $128$ nearest observed points for each local RBF evaluation and a smoothing parameter of $10^{-8}$.

\paragraph*{Kriging~\cite{oliver1990kriging}}
Kriging was applied independently to each snapshot using the isotropic
RBF kernel
$k(\mathbf{x},\mathbf{x}')
=
\exp\!\left(-\frac{\|\mathbf{x}-\mathbf{x}'\|_2^2}{2\ell^2}\right).$
Coordinates were rescaled using the training-domain minima and maxima,
and observed values were standardized per channel within each snapshot.
For each dataset and noise--sensor-density condition, a single length scale $\ell$ was selected using 50 held-out validation snapshots. Specifically, 10 logarithmically spaced candidates in $[0.05,3.0]$ were evaluated by minimizing the mean channel-wise relative $L_2$ error over the complete validation fields. The selected length scale was then fixed for all test snapshots under the corresponding condition. All channels
shared the kernel and optimized length scale, with no cross-channel
covariance. A diagonal term of $2\times10^{-6}I$ stabilized the covariance
matrix.

\paragraph*{Vision transformer (ViT)~\cite{dosovitskiy2021an}}
For ViT, it patchifies the input tensor first with a patch size $16\times16$, and embeds each patch token with the embedding dimension of $384$. Then it uses $12$ transformer blocks with $8$ attention heads and an MLP ratio of $2$.  Finally, a patchwise decoder maps the final tokens back to $128\times 128$ to reconstruct the output field at the original resolution. 

\paragraph*{Axial vision transformer (AViT)~\cite{mccabe2023multiple}}
An axial ViT variant that sequentially attends over spatial and temporal dimensions, enabling efficient modeling of dynamic physical systems. Our implementation follows the base model, consisting of 8 AViT blocks with hidden dimension 512, 8 attention heads, and grouped axial attention across spatial partitions.

\paragraph*{UNet~\cite{ronneberger2015u}}
We adopt a standard 2-D UNet which consists of $4$ levels of downsampling and expansion processes with a downsampling ratio $2$, where the maximum width is $48$. In each level, a 3-layer CNN is employed to downsample or expand the shape, and a 2-layer CNN is used to incorporate skip connections.

\paragraph*{CNextU-Net (ConvNeXt)~\cite{ohana2024well}}
A modernized U-Net backbone built upon ConvNeXt blocks, providing a high-capacity convolutional baseline competitive with transformer-based architectures for spatiotemporal field regression. Our model employs a four-stage encoder--decoder hierarchy with two ConvNeXt blocks per stage and two additional blocks at the bottleneck, initialized with 32 feature channels and doubling features at each downsampling level.

\paragraph*{Fourier neural operator (FNO)~\cite{li2021fourier}}
The Fourier Neural Operator (FNO) applies a Fast Fourier Transform (FFT) along spatial dimensions and multiplies learned complex-valued weights on a limited number of Fourier modes to perform global convolutions in the frequency domain. We implement a 2-D FNO with width $32$, using $32$ Fourier modes per dimension across $4$ spectral layers. 

\paragraph*{DPOT~\cite{hao2024dpot}}
A PDE foundation operator trained via autoregressive denoising and Fourier attention, designed to generalize across diverse PDE systems and boundary conditions. In our implementation, DPOT processes inputs of size 256 × 256 with a patch size of 16, embedding dimension 512 and an MLP ratio of 2. The core architecture consists of a Fourier-attention transformer stack with 8 layers, interleaved with 4 Fourier mixer blocks (via AFNO) for spatial-frequency mixing.

\paragraph*{Senseiver~\cite{santos2023development}}
Senseiver is implemented as a Perceiver-style deterministic baseline for sparse-observation field reconstruction. Each observed token concatenates the normalized coordinate and observed field value, and a set of $128$ learnable latent vectors with dimension $64$ cross-attends to these sensor tokens. The encoder uses $3$ cross-attention blocks with $3$ self-attention layers per block, $2$ attention heads, and preprocessing width $64$. The coordinate-conditioned decoder uses one cross-attention head and predicts the target field at query locations.

\paragraph*{Diffusion posterior sampling (DPS)~\cite{chung2022diffusion}}
Diffusion posterior sampling (DPS) formulates reconstruction as sampling from $p(x\mid y)$, where $y$ denotes the observations and $x$ the unknown field. DPS uses a pre-trained DDPM as a prior for $p(x)$, and modifies the reverse diffusion by a guidance term depending on $y$.
We adopt a conditional $4$-level UNet with the width of $64$ as the backbone of DDPM, pre-trained on  the interpolated dataset. 
During inference, a time-dependent scaling of $\nabla_{x_t} \log p(y \mid x_t)$ is added at each of the 1000 denoising steps, steering the trajectory toward consistency with the measurements.

Although DPS is designed for inverse problems with partial observations, the image-space implementation used here operates on regular-grid representations. We therefore interpolate the irregular point-cloud measurements onto a grid before applying DPS, using the same preprocessing pipeline described above. Accordingly, DPS is treated as an interpolation-based baseline in our comparisons.

\paragraph*{Diffusion PDE~\cite{huang2024diffusionpde}} A conditional diffusion model that jointly learns the distribution of PDE coefficients and solutions, guided by both observation consistency and PDE residual constraints during denoising. We follow the official implementation with a four-stage U-Net backbone using feature channels (64, 128, 256, 512), two residual blocks per stage, and self-attention in the bottleneck layer. Training is conducted for 1,000 diffusion steps with a cosine noise schedule and dual guidance weights tuned to balance data fidelity and physics consistency, selected for optimal reconstruction accuracy.

\subsection{Geometry-based baselines}
We also evaluate models designed for complex geometries with point-cloud inputs or geometry-aware discretizations. We include three competitive, representative methods: Geo-FNO, Transolver, and Voronoi-CNN.

\paragraph*{Geo-FNO~\cite{li2023fourier}}
Geo-FNO  extends the FNO to arbitrary geometries by learning a coordinate transformation that maps the irregular input domain into a latent regular grid, then an FNO-2D or FNO-3D can be applied.
Geo-FNO uses a positional encoding of point coordinates to project input into a latent grid of $40\times40$ (2-D) or $24\times24\times12$ (3-D). Then we use an FNO-2D with $20$ modes or an FNO-3D with $10$ modes to process the data in the projected domain. The number of FNO layers is $5$.
Finally, an inverse deformation is applied to project the output back onto the querying geometry. 

\paragraph*{Transolver~\cite{wu2024transolver}}
Transolver  is a transformer-based neural PDE solver for unstructured meshes and point clouds. It learns Physics-Attention layers to group the multitude of mesh points into a smaller set of latent tokens, representing distinct physical states or regions.
In our implementation, each input coordinate, together with its values and mask, is embedded into a $512$-dimensional feature. The model has $12$ Transolver blocks, each consisting of a learnable grouping aggregation and a self-attention with $8$ heads. The grouping aggregation will aggregate points to $8$ latent slices, where each slice can be seen as a cluster of points that likely share a similar physical behavior. Then the self-attention will learn global interactions across tokens.
We follow the standard baseline configuration with minor tuning to match our datasets.

\paragraph*{Voronoi-CNN~\cite{fukami2021global}}
Voronoi-CNN converts sparse observations on an irregular mesh into a structured grid before applying a convolutional network. Each grid cell takes the value of its nearest observed sensor, forming a Voronoi-style piecewise-constant field, and a binary sensor mask is concatenated to the grid features. For two-dimensional datasets, we use seven same-padding $7\times7$ convolutional layers with $48$ hidden channels and ReLU activations, followed by a $3\times3$ output convolution. For Ahmed body, we use the corresponding 3-D convolutional model on a $24\times32\times64$ grid. The predicted grid field is sampled back to the original mesh or target query points.

\clearpage
\section{Additional results and visualizations}
\label{app:results}

This appendix complements the main-text results with benchmark-specific quantitative summaries and additional qualitative visualizations. For field-reconstruction benchmarks, we report full test-set error statistics and sample-wise error distributions in order to assess both average performance and stability across instances. For geometry-inference benchmarks, we further provide representative field reconstructions and exhaustive tables across observation settings and noise levels, which help clarify how geometry recovery and field recovery interact under increasingly ill-posed conditions.

\begin{table}[ht]
\centering
\caption{\textbf{Field-reconstruction performance with $5\%$ observation noise and a $5\%$ observation fraction.} Entries report relative $L^2$ test error ($\downarrow$), and $N$ denotes the number of observed input points. Values in parentheses give the Pearson correlation coefficient $r$ ($\uparrow$) between pointwise absolute error and predictive standard deviation for Kriging, DPS and GeoFunFlow. Bold and underlined values indicate the best and second-best results, respectively. A dash indicates that a baseline is not applicable.}
\label{tab:noise05_frac05}
\setlength{\tabcolsep}{13pt}
\begin{tabular}{lllll}
\toprule
 & Darcy &Cylinder &Airfoil & Ahmed body \\
& {($N=114$)} & {($N=50$)} & {($N=250$)} & {($N=2000$)} \\
\midrule
Interpolant     & 0.2388 & 0.6989 & 0.1764 & 0.9224 \\
Kriging            & 0.1741 (0.4102) & 0.7434 (0.5466) & 0.3929 (0.0930) & 0.4704 (0.4930) \\
ViT (Interp)       & 0.4043 & 0.3555 & -      & -      \\
UNet (Interp)      & 0.2780 & 0.5030 & -      & -      \\
FNO (Interp)       & 0.3846 & 0.3895 & -      & -      \\
Geo-FNO            & 0.3532 & 1.5937 & 0.4088 & 0.5427 \\
Transolver         & 0.1867 & 0.6511 & 0.4136 & 0.2199 \\
Voronoi-CNN        & 0.1800 & 0.3357 & 0.3309 & 0.2299 \\
Senseiver          & \textbf{0.0995} & \underline{0.2709} & 0.0636 & 0.2873 \\
DPS (Interp)       & 0.8420 (0.1108) & 0.7469 (0.3945) & -      & -      \\
\midrule
GeoFAE (ours)      & 0.1130 & 0.3156 & \underline{0.0525} & \underline{0.1066} \\
GeoFunFlow (ours)  & \underline{0.1106} (0.4935) & \textbf{0.1882} (0.6507) & \textbf{0.0508} (0.4690) & \textbf{0.1049} (0.6080) \\
\bottomrule
\end{tabular}
\end{table}

\subsection{Field reconstruction on known geometries}
The appendix reports two complementary observation settings for the field-reconstruction benchmarks. Tables~\ref{tab:darcy_cylinder} and~\ref{tab:airfoil_ahmed} correspond to the main-text setting with $1\%$ noise and a $25\%$ observation fraction. Tables~\ref{tab:darcy_cylinder_noise005_frac005} and~\ref{tab:airfoil_ahmed_noise005_frac005} report a more ill-posed setting with $5\%$ noise and only a $5\%$ observation fraction. Together, these results separate standard sparse reconstruction from the substantially harder low-observation regime.

\begin{figure}[!ht]
    \centering
    \includegraphics[width=\linewidth]{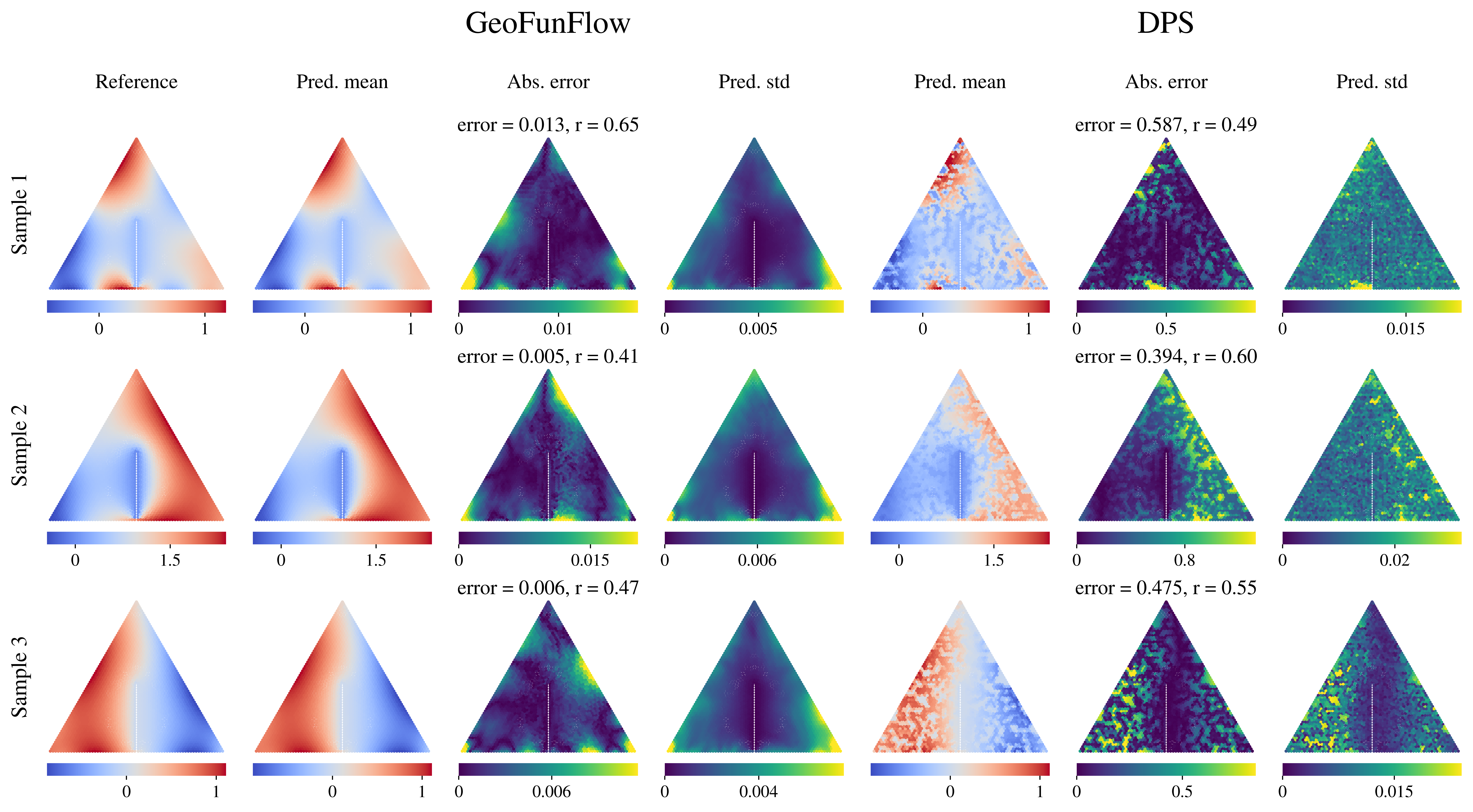}
    \caption{\textbf{Posterior reconstruction and uncertainty on three representative Darcy test samples.} Each row compares GeoFunFlow with DPS and shows the reference pressure field, posterior mean, pointwise absolute error, and predictive standard deviation. Annotations report the relative $L^2$ error and the Pearson correlation $r$ between pointwise error and predictive standard deviation. GeoFunFlow gives lower posterior-mean errors while retaining spatial alignment between uncertainty and reconstruction error.}
    \label{fig:darcy_dps_geofunflow_uq}
\end{figure}

\paragraph*{Darcy}
Tables~\ref{tab:darcy_cylinder} and~\ref{tab:darcy_cylinder_noise005_frac005} show that GeoFunFlow remains among the most accurate Darcy methods when the observation noise increases and the observed fraction decreases. Fig.~\ref{fig:darcy_dps_geofunflow_uq} further separates uncertainty localization from reconstruction fidelity. DPS can attain a moderate, and in aggregate slightly higher, error--uncertainty correlation even though its posterior mean is substantially less accurate. Pearson $r$ quantifies spatial covariation between the uncertainty and error maps but does not assess uncertainty magnitude or interval calibration. The higher DPS value therefore does not by itself indicate more reliable posterior inference. GeoFunFlow achieves substantially lower posterior-mean error while retaining clear spatial correspondence between predictive uncertainty and reconstruction error.

\begin{table}[ht]
\centering
\small
\renewcommand{\arraystretch}{1.25}
\setlength{\tabcolsep}{5pt}
\caption{\textbf{Relative $L^2$ test error on Darcy and Cylinder with $1\%$ observation noise and a $25\%$ observation fraction.} Columns report pressure for Darcy and horizontal velocity, vertical velocity, and pressure for Cylinder. Values report the test-set mean $\pm$ standard deviation. Bold and underlined means denote the best and second-best results, respectively.}
\label{tab:darcy_cylinder}
\begin{tabular}{lcccc}
\toprule
 & Darcy & \multicolumn{3}{c}{Cylinder} \\
\cmidrule(lr){2-2}\cmidrule(lr){3-5}
 & $p$ & $u$ & $v$ & $p$ \\
\midrule
Interpolant & $0.0550 \pm 0.0542$& $0.1046 \pm 0.0293$ & $0.5062 \pm 0.1137$  & $0.2416 \pm 0.0813$ \\
Kriging &$0.0477 \pm 0.0252$ &$0.2014 \pm 0.0276$ &$0.7533 \pm 0.1138$ & $0.5949\pm 0.1794$\\
ViT (Interp) 
& $0.0290 \pm 0.0099$ & $0.1076 \pm 0.0157$ & $0.2774 \pm 0.0318$ & $0.0885 \pm 0.0428$ \\
UNet (Interp) 
& $0.0256 \pm 0.0125$ & $0.1058 \pm 0.0161$ & $0.2678 \pm 0.0280$ & $0.0891 \pm 0.0456$ \\
FNO (Interp) 
& $0.0177 \pm 0.0085$ & $0.1089 \pm 0.0160$ & $0.2806 \pm 0.0388$ & $0.0990 \pm 0.0615$ \\
Geo-FNO 
& $\mathbf{0.0101 \pm 0.0050}$ & $0.2403 \pm 0.0538$ & $0.3483 \pm 0.0987$ & $0.5642 \pm 0.2131$ \\
Transolver 
& $0.0289 \pm 0.0153$ & $0.0630 \pm 0.0220$ & $0.2168 \pm 0.0750$ & $0.1275 \pm 0.0683$ \\
Voronoi-CNN 
& $0.0440 \pm 0.0192$ & $0.0382 \pm 0.0083$ & $0.1348 \pm 0.0234$ & $0.0712 \pm 0.0345$ \\
Senseiver 
& $0.0293 \pm 0.0190$ & $0.0501\pm0.0130$ & $0.1673\pm0.0413$ & $0.0991\pm0.0451$ \\
DPS (Interp) 
& $0.4007 \pm 0.0344$ & $0.1974 \pm 0.0275$ & $0.4162 \pm 0.1445$ & $0.2776 \pm 0.2028$ \\
\midrule
GeoFAE (ours) 
& \underline{$0.0103 \pm 0.0091$} & $\mathbf{0.0327 \pm 0.0304}$ & $\mathbf{0.1074 \pm 0.1004}$ & $\mathbf{0.0524 \pm 0.0516}$ \\
GeoFunFlow (ours) 
& $0.0131 \pm 0.0101$ & \underline{$0.0364 \pm 0.0308$} & \underline{$0.1164 \pm 0.1031$} & \underline{$0.0544 \pm 0.0480$} \\
\bottomrule
\end{tabular}
\end{table}

\begin{table}[ht]
\centering
\small
\renewcommand{\arraystretch}{1.25}
\setlength{\tabcolsep}{5pt}
\caption{\textbf{Relative $L^2$ test error on Darcy and Cylinder with $5\%$ observation noise and a $5\%$ observation fraction.} Columns report pressure for Darcy and horizontal velocity, vertical velocity, and pressure for Cylinder. Values are the test-set mean $\pm$ standard deviation, and lower values indicate more accurate reconstruction.}
\label{tab:darcy_cylinder_noise005_frac005}
\begin{tabular}{lcccc}
\toprule
 & Darcy & \multicolumn{3}{c}{Cylinder} \\
\cmidrule(lr){2-2}\cmidrule(lr){3-5}
 & $p$ & $u$ & $v$ & $p$ \\
\midrule
Interpolant & $0.2388 \pm 0.1814$ & $0.3000 \pm 0.0729$  & $1.0271 \pm 0.2311$  & $0.7686 \pm 0.2197$ \\
Kriging & $0.1741 \pm 0.0938$& $0.5412 \pm 0.0484$  & $0.8920 \pm 0.1306$  & $0.7969 \pm 0.1086$ \\
ViT 
& $0.4043 \pm 0.1092$ & $0.1831 \pm 0.0717$ & $0.4770 \pm 0.1790$ & $0.4065 \pm 0.3379$ \\
UNet 
& $0.2780 \pm 0.0870$ & $0.2296 \pm 0.1262$ & $0.5763 \pm 0.3726$ & $0.7030 \pm 0.8797$ \\
FNO 
& $0.3846 \pm 0.0805$ & $0.2187 \pm 0.0600$ & $0.4986 \pm 0.1656$ & $0.4511 \pm 0.3118$ \\
Geo-FNO 
& $0.3532 \pm 0.1385$ & $1.1309 \pm 0.4032$ & $1.0987 \pm 0.5068$ & $2.5516 \pm 2.0192$ \\
Transolver 
& $0.1867 \pm 0.1382$ & $0.2971 \pm 0.1423$ & $0.7351 \pm 0.3117$ & $0.9211 \pm 1.1222$ \\
Senseiver 
& $\mathbf{0.0995 \pm 0.0576}$ & $0.1241 \pm 0.0497$ & $0.3823 \pm 0.1506$ & $0.3062 \pm 0.1407$ \\
Voronoi-CNN 
& $0.1800 \pm 0.1015$ & $0.1760 \pm 0.0434$ & $0.4583 \pm 0.1280$ & $0.3730 \pm 0.1491$ \\
DPS 
& $0.8420 \pm 0.0244$ & $0.6033 \pm 0.0593$ & $0.8180 \pm 0.1588$ & $0.8194 \pm 0.1451$ \\
\midrule
GeoFAE (ours) 
& $0.1130 \pm 0.0700$ & $0.1642 \pm 0.0812$ & $0.4706 \pm 0.2163$ & $0.3120 \pm 0.2577$ \\
GeoFunFlow (ours) 
& \underline{$0.1106 \pm 0.0643$} & $\mathbf{0.1011 \pm 0.0619}$ & $\mathbf{0.2966 \pm 0.1780}$ & $\mathbf{0.1668 \pm 0.1489}$ \\
\bottomrule
\end{tabular}
\end{table}

\begin{figure}[!ht]
    \centering
    \includegraphics[width=\linewidth]{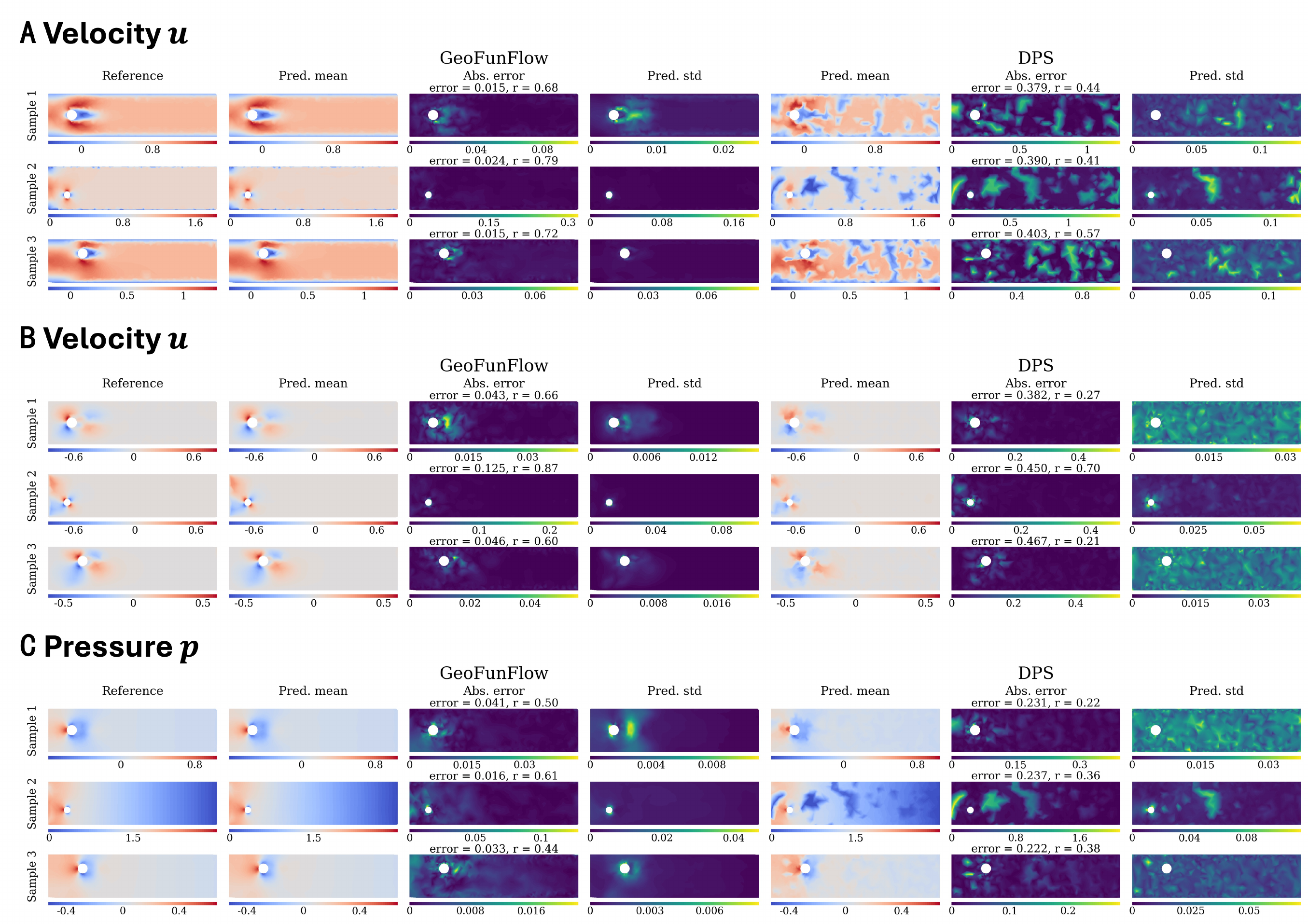}
    \caption{\textbf{Posterior reconstruction and uncertainty on representative Cylinder test samples.} (\textbf{A})--(\textbf{C}) compare GeoFunFlow with DPS for horizontal velocity $u$, vertical velocity $v$, and pressure $p$, respectively. Each panel shows the reference field, posterior mean, pointwise absolute error, and predictive standard deviation for three representative samples. The annotations report the relative $L^2$ error and the spatial Pearson correlation $r$ between pointwise absolute error and predictive standard deviation. GeoFunFlow yields lower posterior-mean errors while preserving spatially informative uncertainty around the cylinder and wake.}
    \label{fig:cylinder_dps_geofunflow_uq}
\end{figure}

\paragraph*{Cylinder}
For Cylinder, sparse sensors only partially constrain the velocity and pressure structures around the body and in the downstream wake. Tables~\ref{tab:darcy_cylinder} and~\ref{tab:darcy_cylinder_noise005_frac005} quantify reconstruction in the standard and low-observation settings, while Fig.~\ref{fig:cylinder_dps_geofunflow_uq} compares the posterior means, errors, and predictive standard deviations of GeoFunFlow and DPS for all three variables. GeoFunFlow gives lower posterior-mean errors, and its uncertainty remains spatially aligned with residual errors around the cylinder and wake.

\begin{figure}[htbp]
    \centering
    \includegraphics[width=1.0\linewidth]{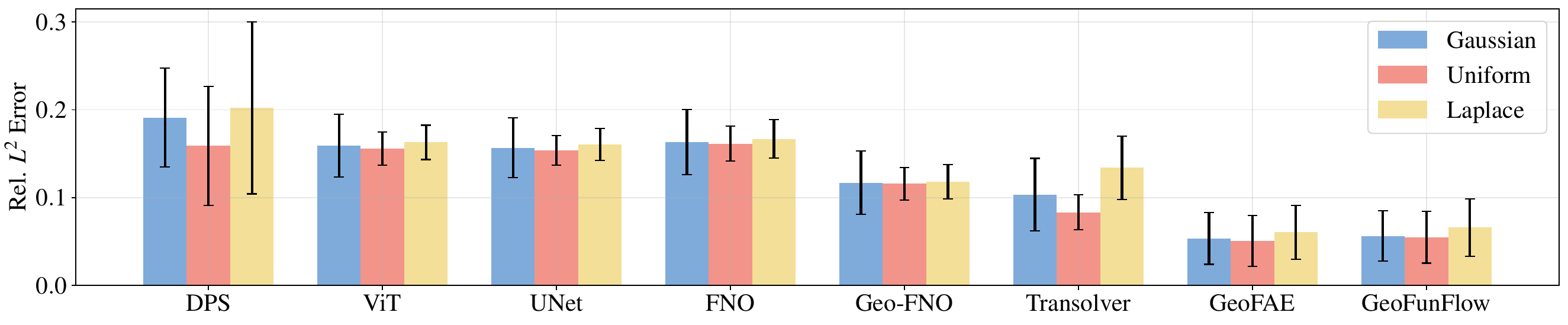}
\caption{\textbf{Sensitivity of Cylinder reconstruction to the observation-noise distribution.} Bars report relative $L^2$ test error for Gaussian, uniform, and Laplace perturbations at a $50\%$ observation fraction and a $1\%$ noise level. Error bars indicate the standard deviation across test samples.}\label{fig:combined_noise_study}
\end{figure}

Fig.~\ref{fig:combined_noise_study} tests sensitivity to the observation-noise distribution at a fixed noise scale and observation fraction. GeoFAE and GeoFunFlow form the lowest-error group under Gaussian, uniform, and Laplace perturbations, and GeoFunFlow varies only modestly across the three distributions in this setting.

\begin{figure}[!htb]
    \vspace{-2mm}
    \centering
    \includegraphics[width=1.0\linewidth]{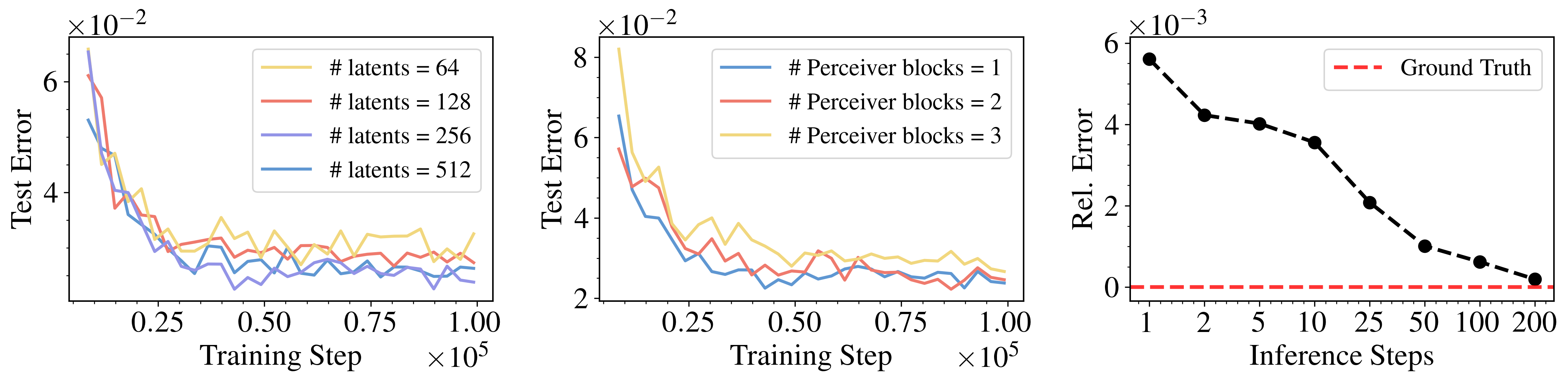}
\caption{\textbf{GeoFAE architecture and GeoFunFlow sampling ablations on Cylinder.} (\textbf{Left}) Relative test error during training for different numbers of Perceiver latent queries. (\textbf{Center}) Relative test error for different numbers of Perceiver blocks. (\textbf{Right}) Relative reconstruction discrepancy from a 1,000-step reference solution as the number of rectified-flow integration steps varies. All other settings are held fixed.}
    \label{fig:cylinder_ablation}
\vskip -0.05in
\end{figure}

Fig.~\ref{fig:cylinder_ablation} examines the Perceiver configuration in GeoFAE and the numerical integration cost of GeoFunFlow on Cylinder. The left and center panels vary the number of latent queries and Perceiver blocks, respectively, while holding the remaining settings fixed. The converged errors differ only modestly across the tested configurations, with one block performing comparably to deeper alternatives. The right panel measures the discrepancy from a 1,000-step numerical reference as the number of rectified-flow integration steps changes. This discrepancy decreases steadily and is already small with few steps, a trend consistent with the near-straight transport paths targeted by rectified flow.

\begin{table}[!htb]
\centering
\small
\renewcommand{\arraystretch}{1.25}
\setlength{\tabcolsep}{5pt}
\caption{\textbf{Relative $L^2$ test error on Airfoil and Ahmed body with $1\%$ observation noise and a $25\%$ observation fraction.} Columns report horizontal velocity, vertical velocity, and pressure for Airfoil and pressure for Ahmed body. Values report the test-set mean $\pm$ standard deviation. Bold and underlined means denote the best and second-best results, respectively.}
\label{tab:airfoil_ahmed}
\begin{tabular}{lcccc}
\toprule
 & \multicolumn{3}{c}{Airfoil} & Ahmed body \\
\cmidrule(lr){2-4}\cmidrule(lr){5-5}
 & $u$ & $v$ & $p$ & $p$ \\
\midrule
Interpolant &$0.0309 \pm 0.0199$  & $0.0714 \pm 0.0246$ & $0.0123 \pm 0.0031$ & $0.3741 \pm 0.0730$\\
Kriging & $0.2272 \pm 0.1673$& $0.7645 \pm 0.1178$  & $0.0942 \pm 0.0667$  & $0.1521 \pm 0.0752$ \\
Geo-FNO 
& $0.1503 \pm 0.0566$ & $0.5589 \pm 0.3059$ & $0.0920 \pm 0.0538$ & $0.2795 \pm 0.0688$ \\
Transolver 
& $0.1499 \pm 0.0564$ & $0.5659 \pm 0.3095$ & $0.0917 \pm 0.0538$ & $0.1262 \pm 0.0245$ \\
Senseiver 
& $0.0139 \pm 0.0104$ & $0.0411 \pm 0.0175$ & $0.0102 \pm 0.0075$ & $0.1070 \pm 0.0444$ \\
Voronoi-CNN 
& $0.2503 \pm 0.1531$ & $0.5880 \pm 0.1602$ & $0.0679 \pm 0.0422$ & $0.0860 \pm 0.0148$ \\
\midrule
GeoFAE (ours) 
& $\mathbf{0.0066 \pm 0.0064}$ & $\mathbf{0.0182 \pm 0.0123}$ & $\mathbf{0.0040 \pm 0.0041}$ & \underline{$0.0819 \pm 0.0405$} \\
GeoFunFlow (ours) 
& \underline{$0.0069 \pm 0.0068$} & \underline{$0.0189 \pm 0.0128$} & \underline{$0.0042 \pm 0.0044$} & $\mathbf{0.0814 \pm 0.0400}$ \\
\bottomrule
\end{tabular}
\end{table}

\begin{table}[!ht]
\centering
\small
\renewcommand{\arraystretch}{1.25}
\setlength{\tabcolsep}{5pt}
\caption{\textbf{Relative $L^2$ test error on Airfoil and Ahmed body with $5\%$ observation noise and a $5\%$ observation fraction.} Columns report horizontal velocity, vertical velocity, and pressure for Airfoil and pressure for Ahmed body. Values report the test-set mean $\pm$ standard deviation. Bold and underlined means denote the best and second-best results, respectively.}
\label{tab:airfoil_ahmed_noise005_frac005}
\begin{tabular}{lcccc}
\toprule
 & \multicolumn{3}{c}{Airfoil} & Ahmed body \\
\cmidrule(lr){2-4}\cmidrule(lr){5-5}
 & $u$ & $v$ & $p$ & $p$ \\
\midrule
Interpolant &  $0.1166 \pm 0.0507$ & $0.3459 \pm 0.1044$  & $0.0666 \pm 0.0122$ & $0.9224 \pm 0.5713$\\
Kriging &$0.2796 \pm 0.1733$  &$0.7992 \pm 0.1143$ &$0.0999 \pm 0.0683$& $0.4704 \pm 0.2325$\\
Geo-FNO 
& $0.2291 \pm 0.0578$ & $0.8854 \pm 0.5861$ & $0.1118 \pm 0.0307$ & $0.5427 \pm 0.1909$ \\
Transolver 
& $0.2296 \pm 0.0582$ & $0.8989 \pm 0.6034$ & $0.1123 \pm 0.0306$ & $0.2199 \pm 0.0521$ \\
Senseiver 
& $0.0398 \pm 0.0200$ & $0.1282 \pm 0.0292$ & $0.0228 \pm 0.0110$ & $0.2873 \pm 0.2522$ \\
Voronoi-CNN 
& $0.2601 \pm 0.1493$ & $0.6553 \pm 0.1832$ & $0.0774 \pm 0.0479$ & $0.2299 \pm 0.0477$ \\
\midrule
GeoFAE (ours) 
& \underline{$0.0355 \pm 0.0221$} & \underline{$0.1079 \pm 0.0944$} & \underline{$0.0140 \pm 0.0115$} & \underline{$0.1066 \pm 0.0491$} \\
GeoFunFlow (ours) 
& $\mathbf{0.0346 \pm 0.0213}$ & $\mathbf{0.1040 \pm 0.0951}$ & $\mathbf{0.0138 \pm 0.0118}$ & $\mathbf{0.1049 \pm 0.0471}$ \\
\bottomrule
\end{tabular}
\end{table}

\paragraph*{Airfoil} The Airfoil benchmark combines varying aerodynamic geometries with highly nonuniform surface-adjacent sampling. Tables~\ref{tab:airfoil_ahmed} and~\ref{tab:airfoil_ahmed_noise005_frac005} show that GeoFunFlow reconstructs the velocity and pressure fields accurately in both observation settings. Its errors remain close to those of the deterministic GeoFAE in the standard setting and are the lowest among the compared methods in the low-observation setting, indicating that posterior sampling does not impose a substantial loss of reconstruction fidelity.

\begin{figure}[!htb]
    \centering
    \includegraphics[width=0.8\linewidth]{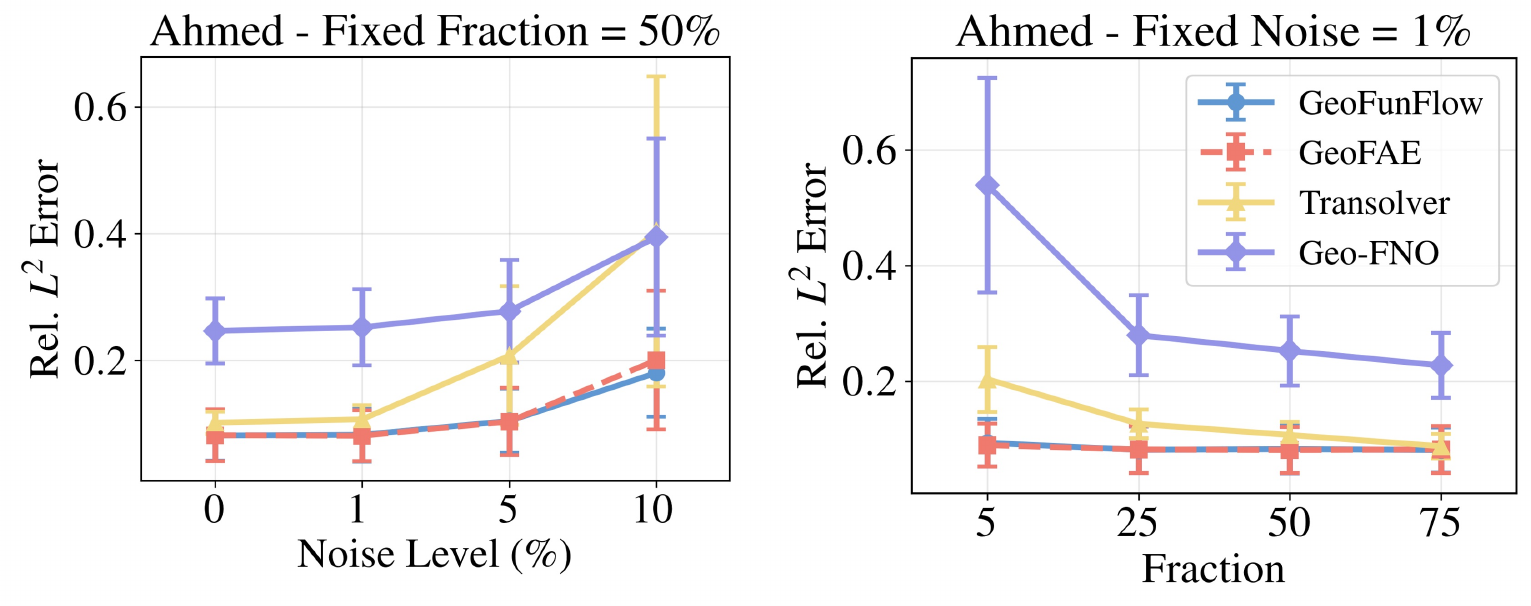}
    \caption[Sensitivity of Ahmed body reconstruction to observation noise and sparsity.]{\textbf{Sensitivity of Ahmed body pressure reconstruction to observation noise and sparsity.} The curves report relative $L^2$ test error for GeoFunFlow, GeoFAE, Transolver, and Geo-FNO. (\textbf{Left}) Noise level varies at a fixed $50\%$ observation fraction. (\textbf{Right}) Observation fraction varies at a fixed $1\%$ noise level. Error bars indicate the standard deviation across test samples.}
    \label{fig:ahmed_violin}
\end{figure}

\paragraph*{Ahmed}
The Ahmed body benchmark evaluates pressure reconstruction on a large, irregular three-dimensional surface discretization. Tables~\ref{tab:airfoil_ahmed} and~\ref{tab:airfoil_ahmed_noise005_frac005} show that GeoFunFlow attains the lowest mean error in both observation settings among the methods evaluated on this mesh. Fig.~\ref{fig:ahmed_violin} complements these two operating points by showing how the error changes with observation noise at a fixed observation fraction and with observation fraction at a fixed noise level.

\clearpage
\subsection{Joint field--geometry inference from partial field information}
\begin{figure}[!htb]
    \centering
    \includegraphics[width=\linewidth]{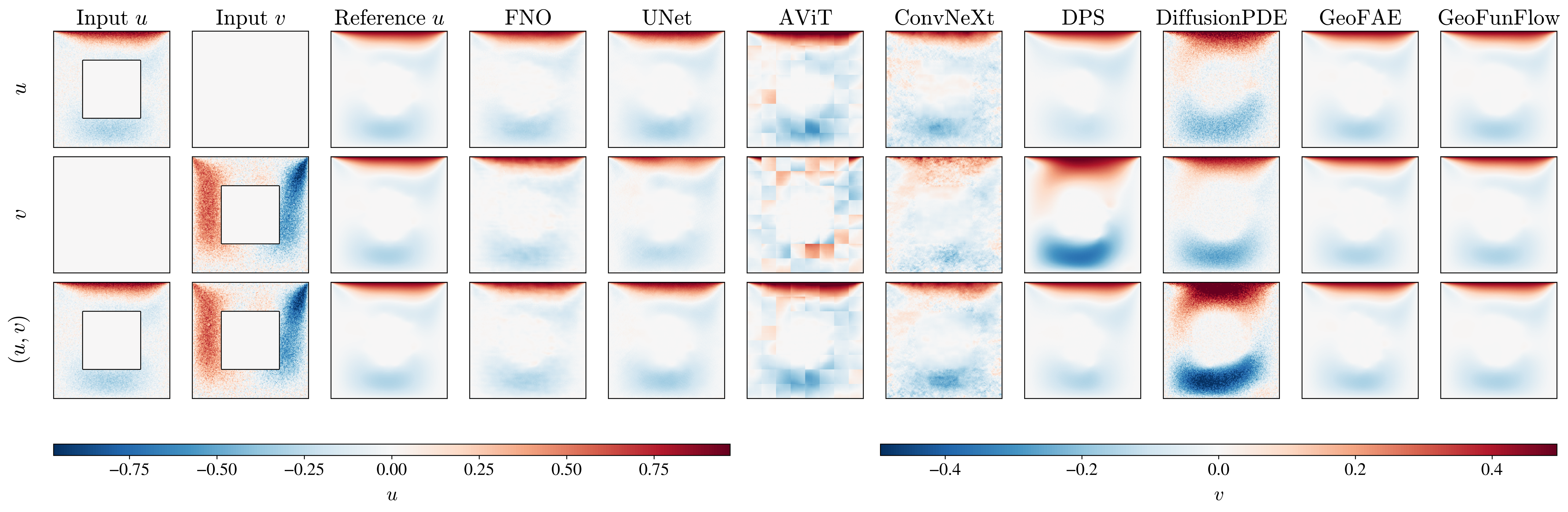}
    \includegraphics[width=\linewidth]{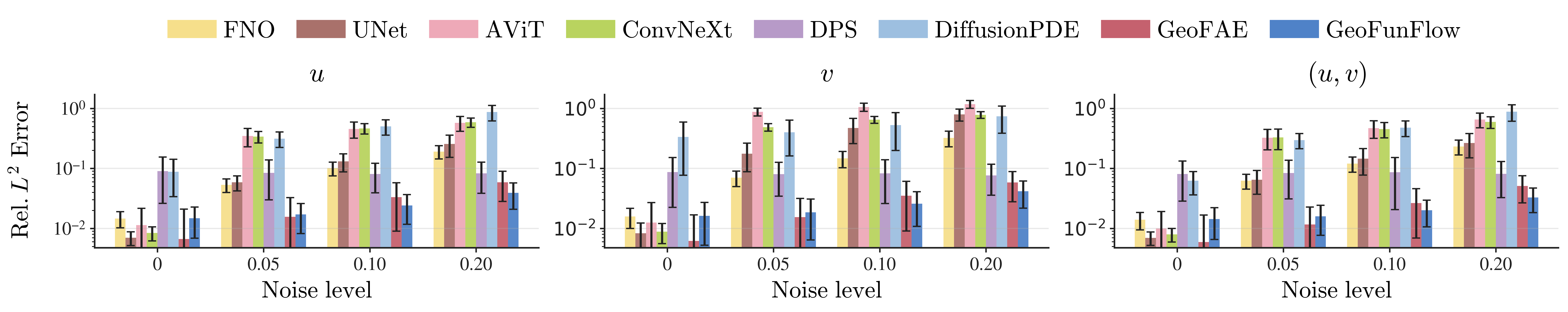}
    \caption{\textbf{Horizontal-velocity reconstruction on LDC across observation settings and noise levels.} (\textbf{Top}) For one representative test case, rows condition on $u$, $v$, or $(u,v)$, and columns show the available inputs, reference $u$ field, and predictions from each method. (\textbf{Bottom}) Bar charts show the relative $L^2$ test error across noise levels for the same three observation settings. Bar heights represent test-set means, and error bars represent standard deviations.}
    \label{fig:LDC_ufield}
\end{figure}

\begin{figure}[!htb]
    \centering
    \includegraphics[width=\linewidth]{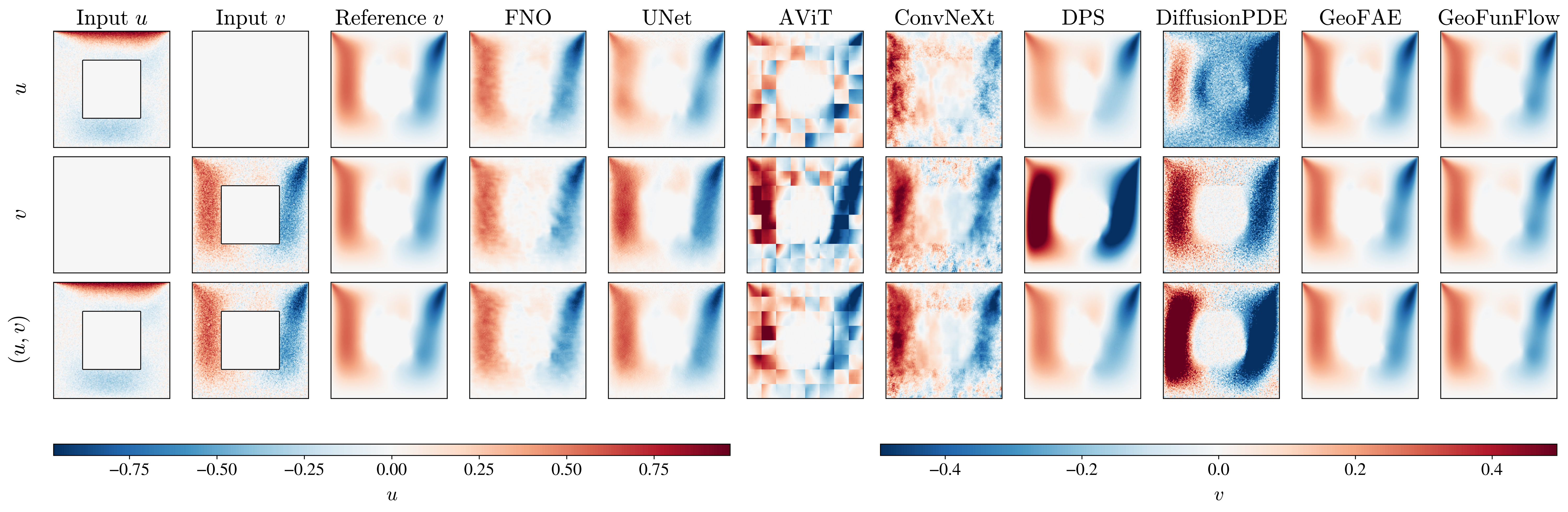}
    \includegraphics[width=\linewidth]{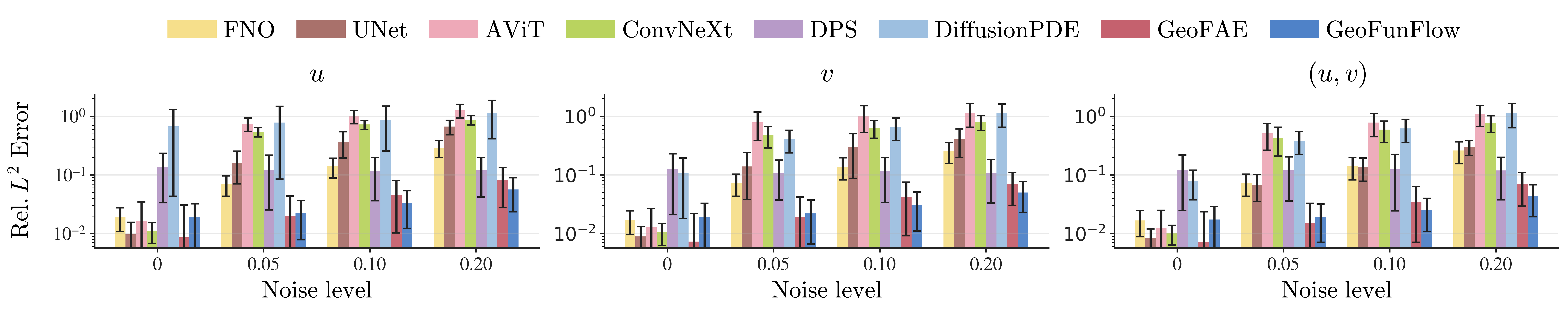}
    \caption{\textbf{Vertical-velocity reconstruction on LDC across observation settings and noise levels.} (\textbf{Top}) For one representative test case, rows condition on $u$, $v$, or $(u,v)$, and columns show the available inputs, reference $v$ field, and predictions from each method. (\textbf{Bottom}) Bar charts show the relative $L^2$ test error across noise levels for the same three observation settings. Bar heights represent test-set means, and error bars represent standard deviations.}
    \label{fig:LDC_vfield}
\end{figure}

\paragraph*{Lid-driven cavity flow}
For LDC, Sec.~\ref{sec:joint_results_LDC} focuses on geometry recovery from partial velocity information. Figs.~\ref{fig:LDC_ufield} and~\ref{fig:LDC_vfield} complement that analysis with representative reconstructions of both velocity components and test-set error summaries across observation settings and noise levels. Table~\ref{tab:relative_l2_all_noise} provides the corresponding numerical results for $u$, $v$, and the SDF, allowing field and geometry reconstruction to be assessed together without conflating their distinct error scales.

\begin{table*}[!htb]
\centering
\caption{\textbf{Relative $L^2$ test error on LDC across observation settings and noise levels.} Results are grouped by conditioning on $u$, $v$, or $(u,v)$ at noise levels of $5\%$, $10\%$, and $20\%$. Each group reports reconstruction of $u$, $v$, and the SDF $\varphi$. Values are the test-set mean $\pm$ standard deviation, with the best and second-best mean in each column shown in bold and underline, respectively.}
\label{tab:relative_l2_all_noise}
\newcommand{\best}[1]{\textbf{#1}}
\newcommand{\second}[1]{\underline{#1}}
\setlength{\tabcolsep}{4pt}
\renewcommand{\arraystretch}{1.05}
\resizebox{\textwidth}{!}{%
\begin{tabular}{lccccccccc}
\toprule
 & \multicolumn{3}{c}{$u$ \textbf{observed}} & \multicolumn{3}{c}{$v$ \textbf{observed}} & \multicolumn{3}{c}{$(u,v)$ \textbf{observed}} \\
\cmidrule(lr){2-4} \cmidrule(lr){5-7} \cmidrule(lr){8-10}
& $u$ & $v$ & SDF & $u$ & $v$ & SDF & $u$ & $v$ & SDF \\
\midrule

\multicolumn{10}{c}{\textbf{Noise level = 0.05}} \\
\midrule
FNO
& 0.0533 $\pm$ 0.0136 & 0.0698 $\pm$ 0.0265 & 0.1170 $\pm$ 0.0186
& 0.0700 $\pm$ 0.0202 & 0.0732 $\pm$ 0.0298 & 0.1319 $\pm$ 0.0234
& 0.0625 $\pm$ 0.0173 & 0.0733 $\pm$ 0.0299 & 0.1155 $\pm$ 0.0180 \\
UNet
& 0.0586 $\pm$ 0.0165 & 0.1623 $\pm$ 0.0919 & 0.1024 $\pm$ 0.0236
& 0.1760 $\pm$ 0.0877 & 0.1395 $\pm$ 0.1012 & 0.1362 $\pm$ 0.0533
& 0.0647 $\pm$ 0.0277 & 0.0681 $\pm$ 0.0330 & 0.0895 $\pm$ 0.0221 \\
AViT
& 0.3471 $\pm$ 0.1176 & 0.7429 $\pm$ 0.1885 & 0.3837 $\pm$ 0.0714
& 0.8795 $\pm$ 0.1299 & 0.7851 $\pm$ 0.3986 & 0.4153 $\pm$ 0.0476
& 0.3255 $\pm$ 0.1216 & 0.5089 $\pm$ 0.2450 & 0.3557 $\pm$ 0.0713 \\
ConvNeXt
& 0.3396 $\pm$ 0.0732 & 0.5419 $\pm$ 0.0982 & 0.5580 $\pm$ 0.0898
& 0.4877 $\pm$ 0.0704 & 0.4761 $\pm$ 0.1886 & 0.5855 $\pm$ 0.1014
& 0.3284 $\pm$ 0.1235 & 0.4292 $\pm$ 0.2199 & 0.5216 $\pm$ 0.1224 \\
DPS
& 0.0842 $\pm$ 0.0542 & 0.1215 $\pm$ 0.0963 & 0.1959 $\pm$ 0.1709
& 0.0805 $\pm$ 0.0460 & 0.1080 $\pm$ 0.0707 & 0.1819 $\pm$ 0.1179
& 0.0840 $\pm$ 0.0531 & 0.1197 $\pm$ 0.0837 & 0.1981 $\pm$ 0.1699 \\
Diffusion PDE
& 0.3145 $\pm$ 0.0903 & 0.7827 $\pm$ 0.6975 & 0.2963 $\pm$ 0.2098
& 0.3999 $\pm$ 0.2382 & 0.4074 $\pm$ 0.1698 & 0.2755 $\pm$ 0.1866
& 0.2975 $\pm$ 0.0843 & 0.3849 $\pm$ 0.1601 & 0.2620 $\pm$ 0.1841 \\
GeoFAE
& \best{0.0156 $\pm$ 0.0169} & \best{0.0202 $\pm$ 0.0235} & \second{0.0186 $\pm$ 0.0258}
& \best{0.0154 $\pm$ 0.0162} & \best{0.0195 $\pm$ 0.0226} & \second{0.0162 $\pm$ 0.0242}
& \best{0.0116 $\pm$ 0.0111} & \best{0.0154 $\pm$ 0.0176} & \best{0.0105 $\pm$ 0.0160} \\
GeoFunFlow
& \second{0.0171 $\pm$ 0.0089} & \second{0.0222 $\pm$ 0.0143} & \best{0.0138 $\pm$ 0.0182}
& \second{0.0186 $\pm$ 0.0122} & \second{0.0220 $\pm$ 0.0153} & \best{0.0133 $\pm$ 0.0161}
& \second{0.0159 $\pm$ 0.0083} & \second{0.0195 $\pm$ 0.0124} & \second{0.0113 $\pm$ 0.0159} \\

\midrule
\multicolumn{10}{c}{\textbf{Noise level = 0.10}} \\
\midrule
FNO
& 0.1011 $\pm$ 0.0262 & 0.1408 $\pm$ 0.0519 & 0.2196 $\pm$ 0.0354
& 0.1474 $\pm$ 0.0448 & 0.1388 $\pm$ 0.0563 & 0.2447 $\pm$ 0.0425
& 0.1206 $\pm$ 0.0340 & 0.1400 $\pm$ 0.0576 & 0.2175 $\pm$ 0.0339 \\
UNet
& 0.1310 $\pm$ 0.0437 & 0.3685 $\pm$ 0.1747 & 0.1933 $\pm$ 0.0463
& 0.4722 $\pm$ 0.2124 & 0.2962 $\pm$ 0.2086 & 0.3144 $\pm$ 0.1438
& 0.1455 $\pm$ 0.0683 & 0.1364 $\pm$ 0.0579 & 0.1826 $\pm$ 0.0538 \\
AViT
& 0.4564 $\pm$ 0.1375 & 1.0002 $\pm$ 0.2589 & 0.4411 $\pm$ 0.0537
& 1.0591 $\pm$ 0.1616 & 1.0143 $\pm$ 0.4893 & 0.4486 $\pm$ 0.0400
& 0.4702 $\pm$ 0.1522 & 0.7801 $\pm$ 0.3334 & 0.4119 $\pm$ 0.0573 \\
ConvNeXt
& 0.4656 $\pm$ 0.0921 & 0.7224 $\pm$ 0.1249 & 0.7278 $\pm$ 0.1138
& 0.6500 $\pm$ 0.0850 & 0.6318 $\pm$ 0.2098 & 0.7492 $\pm$ 0.1178
& 0.4537 $\pm$ 0.1300 & 0.5915 $\pm$ 0.2371 & 0.7191 $\pm$ 0.1291 \\
DPS
& 0.0806 $\pm$ 0.0414 & 0.1171 $\pm$ 0.0810 & 0.1916 $\pm$ 0.1700
& 0.0826 $\pm$ 0.0566 & 0.1154 $\pm$ 0.0815 & 0.1825 $\pm$ 0.1285
& 0.0865 $\pm$ 0.0658 & 0.1241 $\pm$ 0.0997 & 0.1863 $\pm$ 0.1470 \\
Diffusion PDE
& 0.5032 $\pm$ 0.1440 & 0.8714 $\pm$ 0.6148 & 0.3590 $\pm$ 0.2216
& 0.5269 $\pm$ 0.3277 & 0.6591 $\pm$ 0.2715 & 0.3261 $\pm$ 0.2149
& 0.4800 $\pm$ 0.1398 & 0.6204 $\pm$ 0.2656 & 0.3000 $\pm$ 0.1623 \\
GeoFAE
& \second{0.0333 $\pm$ 0.0243} & \second{0.0451 $\pm$ 0.0349} & \second{0.0547 $\pm$ 0.0434}
& \second{0.0350 $\pm$ 0.0259} & \second{0.0423 $\pm$ 0.0331} & \second{0.0477 $\pm$ 0.0399}
& \second{0.0264 $\pm$ 0.0195} & \second{0.0351 $\pm$ 0.0280} & \second{0.0336 $\pm$ 0.0368} \\
GeoFunFlow
& \best{0.0242 $\pm$ 0.0125} & \best{0.0330 $\pm$ 0.0207} & \best{0.0293 $\pm$ 0.0303}
& \best{0.0259 $\pm$ 0.0150} & \best{0.0310 $\pm$ 0.0200} & \best{0.0283 $\pm$ 0.0275}
& \best{0.0201 $\pm$ 0.0095} & \best{0.0253 $\pm$ 0.0146} & \best{0.0186 $\pm$ 0.0190} \\

\midrule
\multicolumn{10}{c}{\textbf{Noise level = 0.20}} \\
\midrule
FNO
& 0.1912 $\pm$ 0.0477 & 0.2922 $\pm$ 0.0958 & 0.3999 $\pm$ 0.0640
& 0.3242 $\pm$ 0.0944 & 0.2556 $\pm$ 0.0994 & 0.4305 $\pm$ 0.0750
& 0.2325 $\pm$ 0.0647 & 0.2610 $\pm$ 0.1059 & 0.3937 $\pm$ 0.0606 \\
UNet
& 0.2558 $\pm$ 0.1026 & 0.6660 $\pm$ 0.1845 & 0.3653 $\pm$ 0.0713
& 0.7946 $\pm$ 0.1789 & 0.4040 $\pm$ 0.2047 & 0.5440 $\pm$ 0.1657
& 0.2653 $\pm$ 0.1153 & 0.2985 $\pm$ 0.0851 & 0.3788 $\pm$ 0.1008 \\
AViT
& 0.5752 $\pm$ 0.1598 & 1.2595 $\pm$ 0.3316 & 0.4729 $\pm$ 0.0408
& 1.1828 $\pm$ 0.1780 & 1.1539 $\pm$ 0.5025 & 0.4814 $\pm$ 0.0422
& 0.6558 $\pm$ 0.1789 & 1.0996 $\pm$ 0.4275 & 0.4534 $\pm$ 0.0404 \\
ConvNeXt
& 0.5892 $\pm$ 0.1037 & 0.8721 $\pm$ 0.1594 & 0.8728 $\pm$ 0.1323
& 0.7817 $\pm$ 0.0998 & 0.7974 $\pm$ 0.2276 & 0.8794 $\pm$ 0.1325
& 0.5954 $\pm$ 0.1302 & 0.7725 $\pm$ 0.2454 & 0.8728 $\pm$ 0.1408 \\
DPS
& 0.0827 $\pm$ 0.0442 & 0.1200 $\pm$ 0.0779 & 0.1923 $\pm$ 0.1436
& 0.0766 $\pm$ 0.0409 & 0.1082 $\pm$ 0.0750 & 0.1908 $\pm$ 0.1413
& 0.0814 $\pm$ 0.0488 & 0.1192 $\pm$ 0.0815 & 0.1920 $\pm$ 0.1752 \\
Diffusion PDE
& 0.8719 $\pm$ 0.2514 & 1.1400 $\pm$ 0.7268 & 0.4469 $\pm$ 0.2060
& 0.7412 $\pm$ 0.3558 & 1.1348 $\pm$ 0.4825 & 0.4329 $\pm$ 0.1836
& 0.8834 $\pm$ 0.2698 & 1.1485 $\pm$ 0.5120 & 0.4167 $\pm$ 0.1651 \\
GeoFAE
& \second{0.0589 $\pm$ 0.0308} & \second{0.0811 $\pm$ 0.0535} & \second{0.0920 $\pm$ 0.0518}
& \second{0.0583 $\pm$ 0.0305} & \second{0.0704 $\pm$ 0.0400} & \second{0.0829 $\pm$ 0.0456}
& \second{0.0511 $\pm$ 0.0244} & \second{0.0699 $\pm$ 0.0404} & \second{0.0641 $\pm$ 0.0493} \\
GeoFunFlow
& \best{0.0391 $\pm$ 0.0182} & \best{0.0564 $\pm$ 0.0329} & \best{0.0599 $\pm$ 0.0430}
& \best{0.0418 $\pm$ 0.0200} & \best{0.0502 $\pm$ 0.0272} & \best{0.0572 $\pm$ 0.0380}
& \best{0.0327 $\pm$ 0.0144} & \best{0.0436 $\pm$ 0.0243} & \best{0.0476 $\pm$ 0.0359} \\

\bottomrule
\end{tabular}%
}
\end{table*}

\begin{figure}[!htb]
    \centering
    \includegraphics[width=\linewidth]{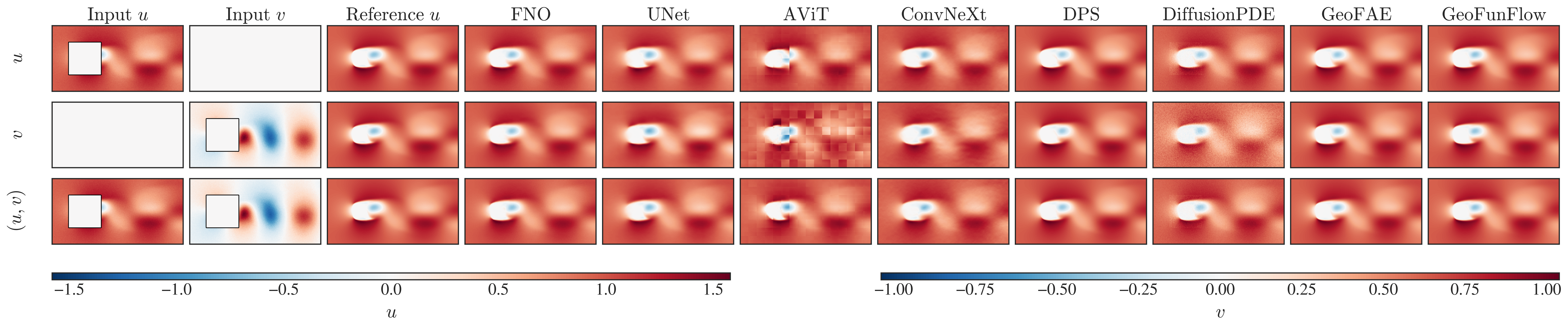}
    \includegraphics[width=\linewidth]{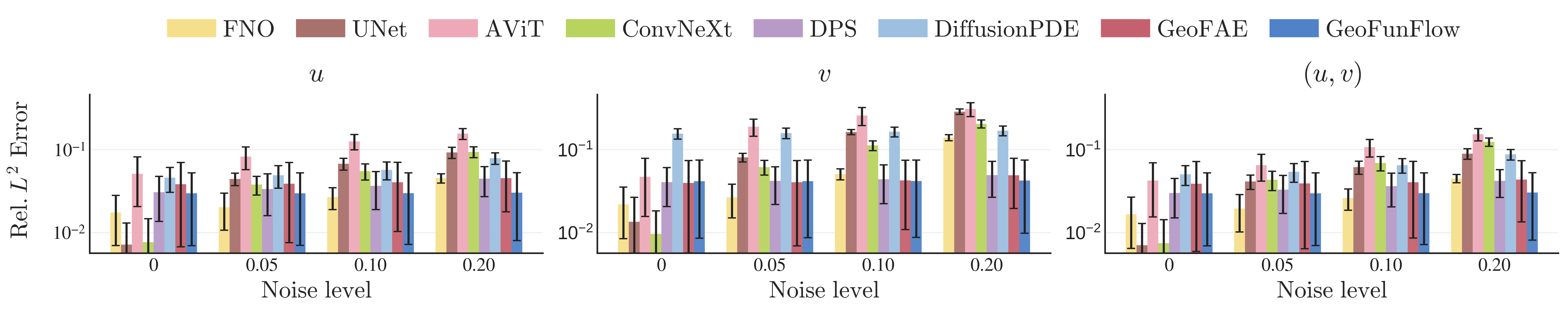}
    \caption{\textbf{Horizontal-velocity reconstruction on FPO across observation settings and noise levels.} (\textbf{Top}) For one representative snapshot, rows condition on $u$, $v$, or $(u,v)$, and columns show the available inputs, reference $u$ field, and predictions from each method. (\textbf{Bottom}) Bar charts show the relative $L^2$ test error across noise levels for the same three observation settings. Bar heights represent test-set means, and error bars represent standard deviations.}
    \label{fig:FPO_ufield}
\end{figure}

\begin{figure}[!htb]
    \centering
    \includegraphics[width=\linewidth]{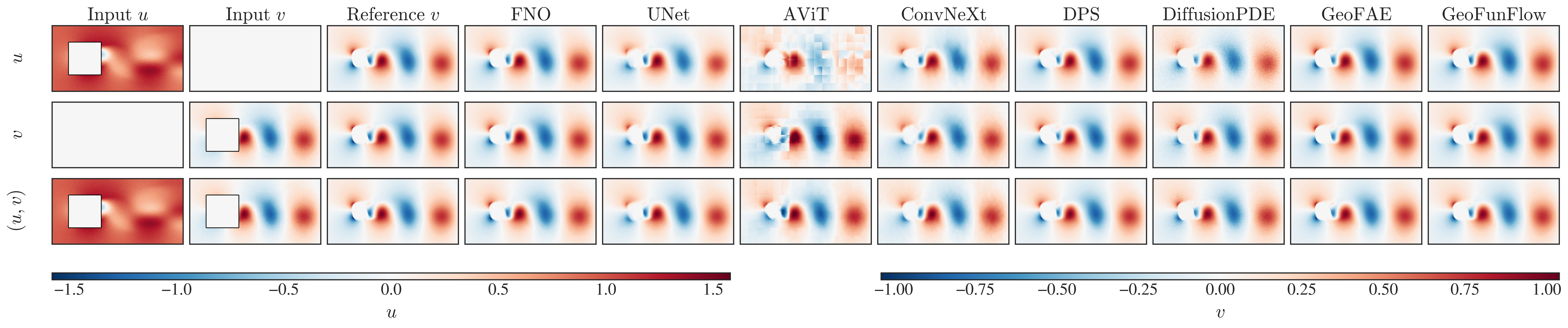}
    \includegraphics[width=\linewidth]{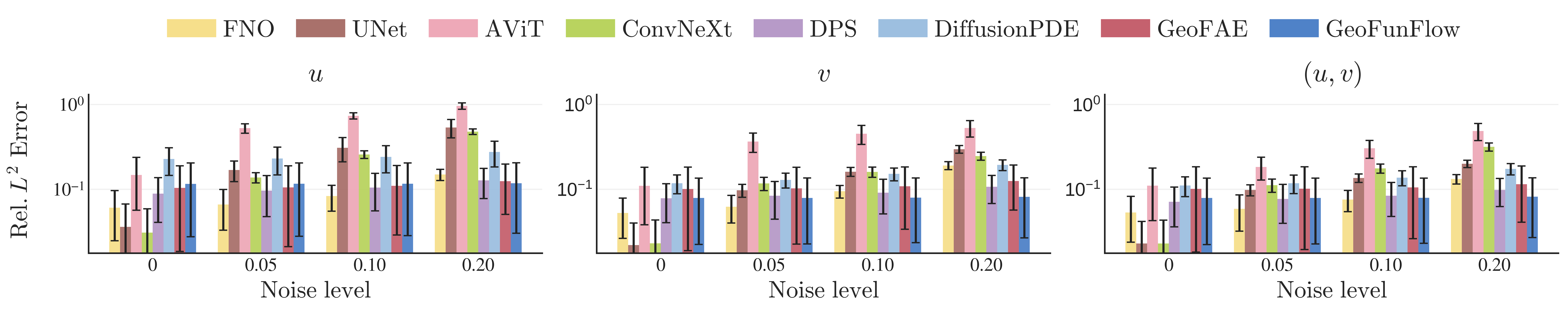}
    \caption{\textbf{Vertical-velocity reconstruction on FPO across observation settings and noise levels.} (\textbf{Top}) For one representative snapshot, rows condition on $u$, $v$, or $(u,v)$, and columns show the available inputs, reference $v$ field, and predictions from each method. (\textbf{Bottom}) Bar charts show the relative $L^2$ test error across noise levels for the same three observation settings. Bar heights represent test-set means, and error bars represent standard deviations.}
    \label{fig:FPO_vfield}
\end{figure}

\paragraph*{Flow past objects}
In FPO, the obstacle geometry is fixed within each trajectory while the observed wake varies across snapshots. Sec.~\ref{sec:joint_results_FPO} examines how aggregating geometry predictions over a temporal window uses these complementary observations. Here, Figs.~\ref{fig:FPO_ufield} and~\ref{fig:FPO_vfield} instead focus on snapshot-wise reconstruction of the two velocity components under different observed variables and noise levels. Table~\ref{tab:relative_l2_all_noise_case2} reports the complete numerical comparison for $u$, $v$, and the SDF. Together, these results distinguish the accuracy of each reconstructed output from the separate benefit of temporal aggregation for geometry inference.

\begin{table*}[!htb]
\centering
\caption{\textbf{Relative $L^2$ test error on FPO across observation settings and noise levels.} Results are grouped by conditioning on $u$, $v$, or $(u,v)$ at noise levels of $5\%$, $10\%$, and $20\%$. Each group reports reconstruction of $u$, $v$, and the SDF $\varphi$. Values are the test-set mean $\pm$ standard deviation, with the best and second-best mean in each column shown in bold and underline, respectively.}
\label{tab:relative_l2_all_noise_case2}

\newcommand{\err}[2]{#1 {\scriptsize$\pm$ #2}}
\newcommand{\best}[1]{\textbf{#1}}
\newcommand{\second}[1]{\underline{#1}}

\setlength{\tabcolsep}{4pt}
\renewcommand{\arraystretch}{1.08}

\resizebox{\textwidth}{!}{%
\begin{tabular}{lccccccccc}
\toprule
& \multicolumn{3}{c}{\textbf{$u$ observed}}
& \multicolumn{3}{c}{\textbf{$v$ observed}}
& \multicolumn{3}{c}{\textbf{$(u,v)$ observed}} \\
\cmidrule(lr){2-4} \cmidrule(lr){5-7} \cmidrule(lr){8-10}
& $u$ & $v$ & SDF & $u$ & $v$ & SDF & $u$ & $v$ & SDF \\
\midrule

\multicolumn{10}{c}{\textbf{Noise level $=0.05$}} \\
\midrule
FNO
& \best{\err{0.0203}{0.0096}} & \best{\err{0.0662}{0.0333}} & \err{0.0110}{0.0052}
& \best{\err{0.0267}{0.0117}} & \best{\err{0.0622}{0.0223}} & \err{0.0149}{0.0043}
& \best{\err{0.0195}{0.0093}} & \best{\err{0.0588}{0.0268}} & \err{0.0107}{0.0052} \\
UNet
& \err{0.0445}{0.0076} & \err{0.1691}{0.0460} & \err{0.0327}{0.0096}
& \err{0.0809}{0.0096} & \err{0.0973}{0.0169} & \err{0.0410}{0.0142}
& \err{0.0413}{0.0082} & \err{0.0980}{0.0145} & \err{0.0318}{0.0095} \\
AViT
& \err{0.0827}{0.0251} & \err{0.5260}{0.0668} & \err{0.0365}{0.0075}
& \err{0.1901}{0.0444} & \err{0.3659}{0.0935} & \err{0.0419}{0.0075}
& \err{0.0650}{0.0232} & \err{0.1837}{0.0554} & \err{0.0297}{0.0065} \\
ConvNeXt
& \err{0.0380}{0.0096} & \err{0.1377}{0.0191} & \err{0.0181}{0.0058}
& \err{0.0618}{0.0125} & \err{0.1173}{0.0211} & \err{0.0234}{0.0076}
& \err{0.0436}{0.0115} & \err{0.1118}{0.0200} & \err{0.0209}{0.0069} \\
DPS
& \err{0.0336}{0.0176} & \second{\err{0.0963}{0.0489}} & \err{0.0186}{0.0109}
& \err{0.0421}{0.0203} & \err{0.0838}{0.0395} & \err{0.0240}{0.0123}
& \err{0.0331}{0.0160} & \second{\err{0.0769}{0.0372}} & \err{0.0203}{0.0091} \\
Diffusion PDE
& \err{0.0493}{0.0149} & \err{0.2309}{0.0833} & \err{0.0711}{0.0119}
& \err{0.1593}{0.0231} & \err{0.1292}{0.0259} & \err{0.1541}{0.0200}
& \err{0.0543}{0.0138} & \err{0.1178}{0.0292} & \err{0.0718}{0.0119} \\
GeoFAE
& \err{0.0388}{0.0313} & \err{0.1052}{0.0841} & \second{\err{0.0095}{0.0068}}
& \second{\err{0.0404}{0.0335}} & \err{0.1023}{0.0798} & \second{\err{0.0096}{0.0068}}
& \err{0.0392}{0.0329} & \err{0.1018}{0.0824} & \second{\err{0.0095}{0.0068}} \\
GeoFunFlow
& \second{\err{0.0298}{0.0228}} & \err{0.1160}{0.0882} & \best{\err{0.0065}{0.0076}}
& \err{0.0418}{0.0332} & \second{\err{0.0791}{0.0566}} & \best{\err{0.0066}{0.0074}}
& \second{\err{0.0298}{0.0228}} & \err{0.0791}{0.0565} & \best{\err{0.0058}{0.0071}} \\

\midrule
\multicolumn{10}{c}{\textbf{Noise level $=0.10$}} \\
\midrule
FNO
& \best{\err{0.0269}{0.0079}} & \best{\err{0.0831}{0.0282}} & \err{0.0123}{0.0050}
& \err{0.0510}{0.0078} & \err{0.0947}{0.0160} & \err{0.0362}{0.0055}
& \best{\err{0.0261}{0.0075}} & \best{\err{0.0755}{0.0213}} & \err{0.0122}{0.0047} \\
UNet
& \err{0.0677}{0.0111} & \err{0.3092}{0.0983} & \err{0.0802}{0.0258}
& \err{0.1638}{0.0120} & \err{0.1613}{0.0193} & \err{0.1487}{0.0304}
& \err{0.0617}{0.0112} & \err{0.1356}{0.0155} & \err{0.0760}{0.0237} \\
AViT
& \err{0.1264}{0.0265} & \err{0.7366}{0.0619} & \err{0.0540}{0.0067}
& \err{0.2595}{0.0639} & \err{0.4528}{0.1147} & \err{0.0562}{0.0080}
& \err{0.1074}{0.0258} & \err{0.3043}{0.0727} & \err{0.0473}{0.0070} \\
ConvNeXt
& \err{0.0553}{0.0122} & \err{0.2571}{0.0267} & \err{0.0286}{0.0070}
& \err{0.1127}{0.0154} & \err{0.1605}{0.0224} & \err{0.0316}{0.0091}
& \err{0.0691}{0.0137} & \err{0.1766}{0.0216} & \err{0.0341}{0.0080} \\
DPS
& \err{0.0367}{0.0177} & \second{\err{0.1048}{0.0493}} & \err{0.0205}{0.0113}
& \err{0.0441}{0.0217} & \second{\err{0.0911}{0.0399}} & \err{0.0238}{0.0125}
& \err{0.0363}{0.0159} & \err{0.0840}{0.0364} & \err{0.0222}{0.0102} \\
Diffusion PDE
& \err{0.0573}{0.0140} & \err{0.2412}{0.0849} & \err{0.0783}{0.0125}
& \err{0.1652}{0.0224} & \err{0.1519}{0.0261} & \err{0.1645}{0.0194}
& \err{0.0652}{0.0130} & \err{0.1373}{0.0277} & \err{0.0836}{0.0124} \\
GeoFAE
& \err{0.0405}{0.0302} & \err{0.1098}{0.0810} & \second{\err{0.0101}{0.0064}}
& \second{\err{0.0427}{0.0319}} & \err{0.1086}{0.0749} & \second{\err{0.0102}{0.0067}}
& \err{0.0404}{0.0319} & \err{0.1051}{0.0790} & \second{\err{0.0098}{0.0066}} \\
GeoFunFlow
& \second{\err{0.0299}{0.0227}} & \err{0.1162}{0.0878} & \best{\err{0.0065}{0.0076}}
& \best{\err{0.0419}{0.0331}} & \best{\err{0.0797}{0.0562}} & \best{\err{0.0067}{0.0072}}
& \second{\err{0.0299}{0.0227}} & \second{\err{0.0792}{0.0562}} & \best{\err{0.0059}{0.0070}} \\

\midrule
\multicolumn{10}{c}{\textbf{Noise level $=0.20$}} \\
\midrule
FNO
& \err{0.0454}{0.0057} & \err{0.1494}{0.0224} & \err{0.0190}{0.0043}
& \err{0.1404}{0.0118} & \err{0.1910}{0.0208} & \err{0.1219}{0.0147}
& \err{0.0451}{0.0051} & \err{0.1321}{0.0169} & \err{0.0227}{0.0050} \\
UNet
& \err{0.0927}{0.0141} & \err{0.5352}{0.1304} & \err{0.1785}{0.0446}
& \err{0.2890}{0.0220} & \err{0.2966}{0.0308} & \err{0.3560}{0.0406}
& \err{0.0898}{0.0132} & \err{0.2004}{0.0200} & \err{0.1674}{0.0403} \\
AViT
& \err{0.1567}{0.0233} & \err{0.9637}{0.0834} & \err{0.0748}{0.0066}
& \err{0.3109}{0.0597} & \err{0.5299}{0.1173} & \err{0.0748}{0.0093}
& \err{0.1539}{0.0263} & \err{0.4865}{0.1112} & \err{0.0657}{0.0071} \\
ConvNeXt
& \err{0.0943}{0.0142} & \err{0.4788}{0.0364} & \err{0.2051}{0.0159}
& \err{0.2062}{0.0228} & \err{0.2464}{0.0265} & \err{0.1080}{0.0139}
& \err{0.1243}{0.0143} & \err{0.3168}{0.0346} & \err{0.2641}{0.0158} \\
DPS
& \second{\err{0.0449}{0.0177}} & \err{0.1271}{0.0494} & \err{0.0242}{0.0130}
& \err{0.0495}{0.0228} & \second{\err{0.1069}{0.0389}} & \err{0.0242}{0.0128}
& \second{\err{0.0421}{0.0155}} & \second{\err{0.0982}{0.0358}} & \err{0.0244}{0.0111} \\
Diffusion PDE
& \err{0.0792}{0.0126} & \err{0.2760}{0.0929} & \err{0.0959}{0.0127}
& \err{0.1703}{0.0229} & \err{0.1941}{0.0280} & \err{0.1767}{0.0195}
& \err{0.0879}{0.0129} & \err{0.1741}{0.0266} & \err{0.1042}{0.0130} \\
GeoFAE
& \err{0.0455}{0.0277} & \second{\err{0.1243}{0.0739}} & \second{\err{0.0115}{0.0060}}
& \second{\err{0.0493}{0.0297}} & \err{0.1249}{0.0680} & \second{\err{0.0122}{0.0066}}
& \err{0.0436}{0.0302} & \err{0.1147}{0.0739} & \second{\err{0.0108}{0.0063}} \\
GeoFunFlow
& \best{\err{0.0304}{0.0224}} & \best{\err{0.1176}{0.0873}} & \best{\err{0.0071}{0.0073}}
& \best{\err{0.0425}{0.0327}} & \best{\err{0.0815}{0.0548}} & \best{\err{0.0074}{0.0069}}
& \best{\err{0.0305}{0.0224}} & \best{\err{0.0816}{0.0547}} & \best{\err{0.0063}{0.0069}} \\

\bottomrule
\end{tabular}%
}
\end{table*}

\clearpage
\subsection{Uncertainty quantification and calibration}
Table~\ref{tab:uq_picp_by_noise_level} expands the SDF block of Table~\ref{tab:uq_combined_by_dataset} by reporting empirical coverage at nominal levels from 50\% to 95\%. The raw intervals are under-covering across the evaluated datasets and noise levels. Applying the scale fitted on the calibration subset moves holdout coverage closer to the nominal levels throughout this range, although the remaining deviations are reported explicitly.

\begin{table}[htbp]
\centering
\tiny
\setlength{\tabcolsep}{2.0pt}
\renewcommand{\arraystretch}{0.95}
\caption{\textbf{Empirical SDF interval coverage before and after calibration.} PICP is reported at nominal coverage levels of $50\%$, $80\%$, $90\%$, and $95\%$ for each dataset and noise level. For each of three repeated splits, the global SDF calibration factor is fitted on a calibration subset and evaluated on a disjoint holdout subset. Entries report the mean $\pm$ standard deviation across splits in the form raw $\rightarrow$ calibrated.}
\label{tab:uq_picp_by_noise_level}
\tiny
\begin{tabular}{@{}llcccc@{}}
\toprule
& Noise & PICP@50\% & PICP@80\% & PICP@90\% & PICP@95\% \\
\midrule
\multirow{3}{*}{LDC}
& 0.05 & $0.232 \pm 0.004 \rightarrow \mathbf{0.568 \pm 0.054}$ & $0.413 \pm 0.006 \rightarrow \mathbf{0.819 \pm 0.044}$ & $0.508 \pm 0.008 \rightarrow \mathbf{0.890 \pm 0.032}$ & $0.579 \pm 0.010 \rightarrow \mathbf{0.927 \pm 0.023}$ \\
& 0.10 & $0.171 \pm 0.003 \rightarrow \mathbf{0.581 \pm 0.070}$ & $0.302 \pm 0.008 \rightarrow \mathbf{0.824 \pm 0.052}$ & $0.377 \pm 0.009 \rightarrow \mathbf{0.888 \pm 0.037}$ & $0.439 \pm 0.010 \rightarrow \mathbf{0.922 \pm 0.026}$ \\
& 0.20 & $0.116 \pm 0.002 \rightarrow \mathbf{0.594 \pm 0.050}$ & $0.221 \pm 0.003 \rightarrow \mathbf{0.782 \pm 0.035}$ & $0.280 \pm 0.004 \rightarrow \mathbf{0.832 \pm 0.028}$ & $0.327 \pm 0.004 \rightarrow \mathbf{0.862 \pm 0.024}$ \\
\midrule
\multirow{3}{*}{FPO}
& 0.05 & $0.073 \pm 0.001 \rightarrow \mathbf{0.562 \pm 0.013}$ & $0.140 \pm 0.002 \rightarrow \mathbf{0.776 \pm 0.018}$ & $0.179 \pm 0.003 \rightarrow \mathbf{0.841 \pm 0.019}$ & $0.211 \pm 0.004 \rightarrow \mathbf{0.878 \pm 0.018}$ \\
& 0.10 & $0.085 \pm 0.001 \rightarrow \mathbf{0.577 \pm 0.001}$ & $0.160 \pm 0.002 \rightarrow \mathbf{0.779 \pm 0.009}$ & $0.203 \pm 0.003 \rightarrow \mathbf{0.842 \pm 0.011}$ & $0.240 \pm 0.003 \rightarrow \mathbf{0.878 \pm 0.012}$ \\
& 0.20 & $0.090 \pm 0.002 \rightarrow \mathbf{0.573 \pm 0.010}$ & $0.168 \pm 0.003 \rightarrow \mathbf{0.784 \pm 0.005}$ & $0.214 \pm 0.003 \rightarrow \mathbf{0.844 \pm 0.003}$ & $0.252 \pm 0.003 \rightarrow \mathbf{0.879 \pm 0.004}$ \\
\midrule
\multirow{3}{*}{OT}
& 0.05 & $0.099 \pm 0.002 \rightarrow \mathbf{0.589 \pm 0.043}$ & $0.188 \pm 0.004 \rightarrow \mathbf{0.819 \pm 0.034}$ & $0.237 \pm 0.005 \rightarrow \mathbf{0.889 \pm 0.026}$ & $0.280 \pm 0.006 \rightarrow \mathbf{0.925 \pm 0.020}$ \\
& 0.10 & $0.090 \pm 0.002 \rightarrow \mathbf{0.553 \pm 0.031}$ & $0.168 \pm 0.004 \rightarrow \mathbf{0.804 \pm 0.025}$ & $0.213 \pm 0.005 \rightarrow \mathbf{0.877 \pm 0.019}$ & $0.250 \pm 0.006 \rightarrow \mathbf{0.917 \pm 0.015}$ \\
& 0.20 & $0.063 \pm 0.001 \rightarrow \mathbf{0.534 \pm 0.021}$ & $0.118 \pm 0.001 \rightarrow \mathbf{0.782 \pm 0.016}$ & $0.151 \pm 0.002 \rightarrow \mathbf{0.857 \pm 0.015}$ & $0.179 \pm 0.002 \rightarrow \mathbf{0.900 \pm 0.013}$ \\
\bottomrule
\end{tabular}
\end{table}

\end{appendices}

\end{document}